\documentclass{article}

\usepackage{microtype}
\usepackage{graphicx}
\usepackage{subcaption}
\usepackage{booktabs} 

\usepackage{hyperref}
\usepackage[table,dvipsnames]{xcolor}
\usepackage{enumitem}

\usepackage[preprint]{icml2026}

\usepackage{amsmath}
\usepackage{amssymb}
\usepackage{mathtools}
\usepackage{amsthm}

\usepackage{pifont}
\usepackage{multirow}
\definecolor{darkblue}{HTML}{006a80} 
\definecolor{darkred}{HTML}{9f2e2e}
\definecolor{darkgreen}{HTML}{006400}
\newcommand{\bestr}[1]{\textcolor{darkred}{\textbf{#1}}}
\newcommand{\secr}[1]{\textcolor{darkred}{\textbf{\underline{#1}}}}
\newcommand{\bestb}[1]{\textcolor{darkblue}{\textbf{#1}}}
\newcommand{\secb}[1]{\textcolor{darkblue}{\textbf{\underline{#1}}}}

\usepackage[capitalize,noabbrev]{cleveref}

\theoremstyle{plain}

\theoremstyle{definition}

\theoremstyle{remark}

\hypersetup{
hidelinks,
colorlinks=true,
linkcolor=red,
urlcolor= RoyalBlue,
citecolor=cyan
}

\icmltitlerunning{CombinationTS: A Modular Framework for Understanding Time-Series Forecasting Models}

\begin{document}

\twocolumn[
  \icmltitle{CombinationTS: A Modular Framework for Understanding \\
  Time-Series Forecasting Models}

  \icmlsetsymbol{equal}{*}
  \icmlsetsymbol{corresponding}{$\dagger$}
  
  \begin{icmlauthorlist}
    \icmlauthor{Xiaorui Wang}{ict,ucas}
    \icmlauthor{Fanda Fan}{corresponding,ict}
    \icmlauthor{Chenxi Wang}{ict,ucas}
    \icmlauthor{Yuxuan Yang}{ict,ucas}
    \icmlauthor{Rui Tang}{bnu}
    \icmlauthor{Kuoyu Gao}{neu}
    \icmlauthor{Simiao Pang}{neu}
    \icmlauthor{Yuanfeng Shang}{ict}
    \icmlauthor{Zhipeng Liu}{neu}
    \icmlauthor{Wanling Gao}{ict}
    \icmlauthor{Lei Wang}{ict}
    \icmlauthor{Jianfeng Zhan}{ict,ucas}
  \end{icmlauthorlist}

  \icmlaffiliation{ict}{Institute of Computing Technology, Chinese Academy of Sciences, Beijing, China}
  \icmlaffiliation{ucas}{University of Chinese Academy of Sciences, Beijing, China}
  \icmlaffiliation{neu}{Northeastern University, Shenyang, China}
  \icmlaffiliation{bnu}{Beijing Normal University - Hong Kong Baptist University United International College, Zhuhai, China}

  \icmlcorrespondingauthor{Fanda Fan}{fanfanda@ict.ac.cn}

  \icmlkeywords{Time Series, Evaluation, Causality}

  \vskip 0.3in
]

\printAffiliationsAndNotice{}

\begin{abstract}
Recent progress in time-series forecasting has led to rapidly increasing architectural complexity, yet many reported State-of-the-Art gains are statistically fragile or misattributed.
We argue that progress requires a shift from model selection to modular attribution, identifying which components truly drive performance.
We propose CombinationTS, a self-contained probabilistic evaluation framework that decomposes forecasting models into orthogonal modules—Input Transformation, Embedding, Encoder, Decoder, and Output Transformation—and evaluates them under a shared evaluation condition space.
By quantifying each component via marginalized performance ($\mu$) and stability ($\sigma$), CombinationTS enables robust attribution beyond fragile point estimates.
Through large-scale paired evaluation, we uncover the Identity Paradox: once the data view (Embedding) is well-designed, a parameter-free Identity Encoder often matches or outperforms complex backbones.
We further show that explicit structural priors introduced via Input Transformations yield a more favorable performance–stability trade-off than increasing Encoder complexity, establishing a principled baseline for architectural necessity.
The code is available at \url{https://github.com/BenchCouncil/CombinationTS}.
\end{abstract}

\begin{figure*}[t]
    \begin{center}
        \includegraphics[width=\textwidth]{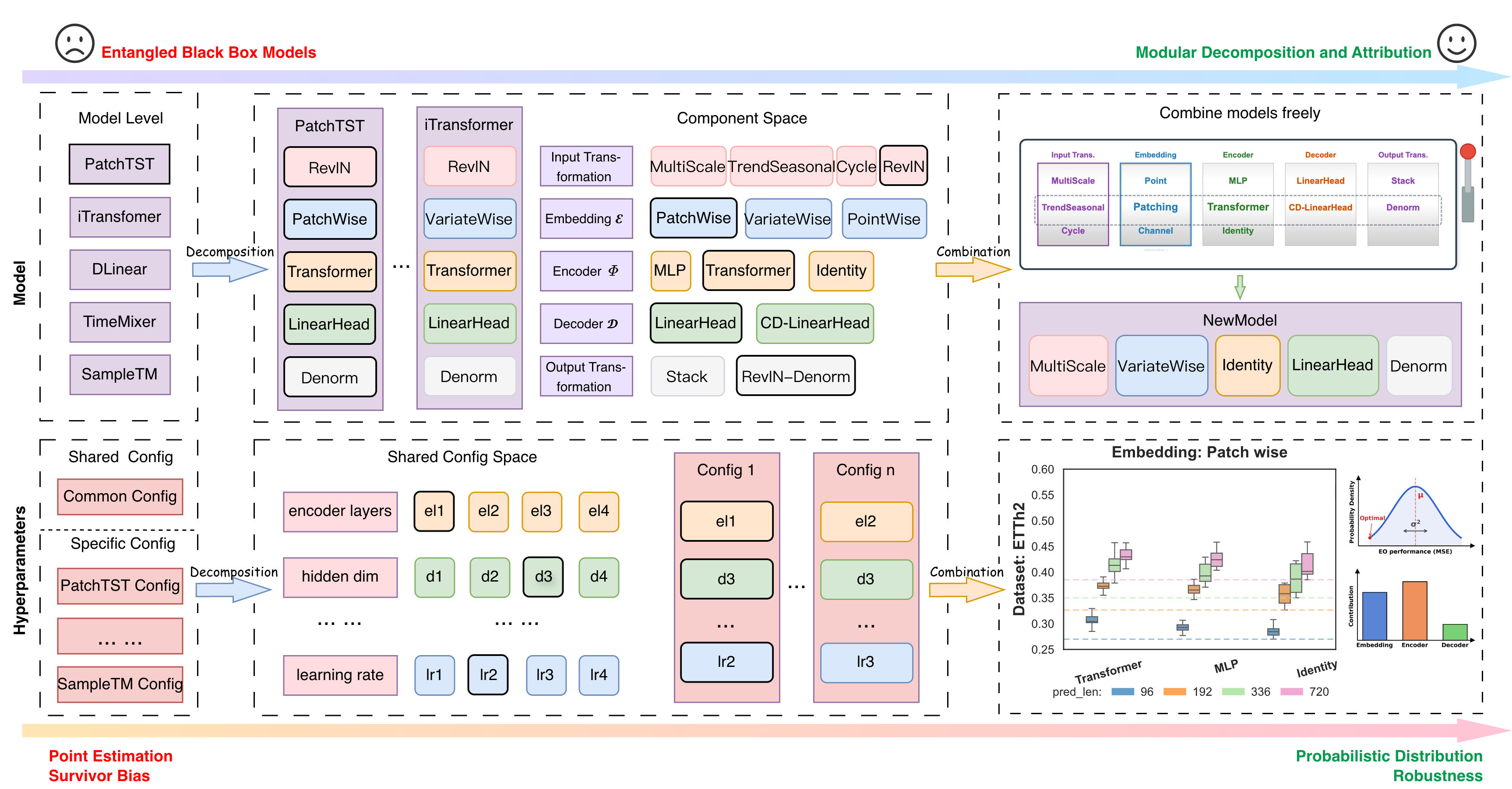}
    \end{center}
    \caption{
        The conceptual framework of CombinationTS. The diagram illustrates the paradigm shift proposed in this work across two dimensions. (Top) Model Architecture: We transition from entangled Black Box models to a modular decomposition framework. By dismantling models into five orthogonal components—Input Transformation, Embedding, Encoder, Decoder, and Output Transformation—we enable free recombination to identify the true sources of performance attribution. (Bottom) Evaluation Paradigm: We shift from point estimation, which is susceptible to survivor bias, to probabilistic distribution–based evaluation. Instead of fixing model-specific configurations or a shared configuration, we assess the robustness of the evaluated object over a shared configuration space, using distributional metrics to ensure fair and comparable evaluation.
    }
    \label{fig:fig1}
\end{figure*}

\section{Introduction}
\label{sec:intro}

Time-series forecasting drives high-stakes decisions in finance~\cite{sezer2020financial,liu2025multiplex,liu2025hypergraph}, energy~\cite{pinto2021ensemble-energy}, healthcare~\cite{kaushik2020ai-healthcare}, transportation~\cite{sun2023meta-transportation}, and climate science~\cite{du2019deep-climate}.
The landscape of Time-Series Forecasting (TSF) has undergone rapid architectural expansion, with models becoming increasingly complex in both structure and design.
Importantly, this complexity has not arisen solely from deeper or more powerful modeling modules, but from a progressive shift toward richer \emph{data view} (Embedding) and \emph{Input Transformation}.

Recent advances move from sparse-attention Transformers~\cite{zhou2021informer,zhou2022fedformer} to patch-based tokenization~\cite{nie2022time-patchtst,stitsyuk2025xpatch,hu2025timefilter}, channel-wise embeddings~\cite{liu2023itransformer,wang2024timexer,qiu2025duet}, multi-scale MLP-based designs~\cite{wang2024timemixer,tang2025unlocking}, and attention-modulation approaches~\cite{liu2026timeformer}, all of which substantially reshape how data is presented to the model rather than how it is reasoned over.
Such a trend of escalating architectural complexity is not confined to forecasting alone; analogous patterns of multi-scale redundancy and domain robustness challenges have been observed across broader time series analysis tasks, including classification~\cite{liu2025dismsTs,liu2026dualcd}.

While these developments report steady gains on public leaderboards, they also entangle architectural design with data preprocessing choices and evaluation configurations, making performance improvements increasingly difficult to attribute.
Recent studies show that the reported gap between State-of-the-Art models and strong baselines is often comparable to variability induced by random seeds or hyperparameter alignment~\cite{tan2024are,brigato2025position}, raising a fundamental question:
\emph{Do modern TSF gains reflect genuine architectural reasoning, or are they largely driven by improved data view and favorable evaluation conditions?}

This ambiguity stems from two fundamental methodological deficits in current benchmarking practices: (i). The Attribution Gap (Monolithic vs. Modular): Existing evaluations treat models as indivisible ``black boxes''~\cite{shao2024exploring-basicts,qiu2024tfb,qiu2025tab,wang2024tssurvey-tslib}, conflating the contribution of the \textit{data view} (Embedding) with that of the \textit{Encoder} (e.g., Attention, MLP). Consequently, it remains unclear whether a model's success is driven by \textit{how it sees data} or \textit{how it processes it}. (ii). The Benchmarking Crisis (Point vs. Distribution): Standard protocols~\cite{wu2023timesnet,wang2024timemixer} often fall into the ``Fairness Trap'' (using suboptimal fixed settings) or the ``Best Trap'' (cherry-picking peak results). Such point estimates fail to capture the \textit{expected performance} ($\mu$) and \textit{stability} ($\sigma$) of a design under the epistemic uncertainty of real-world hyperparameters.

To resolve these deficits, we advocate a shift from \textit{Model Selection} to \textit{Modular Attribution}, and construct a self-contained evaluation system for time-series forecasting.
Rather than proposing yet another architecture, \textbf{CombinationTS} reformulates model analysis as a component-level attribution problem.
Specifically, we decompose the forecasting pipeline into five orthogonal stages—\textit{Input Transformation, Embedding, Encoder, Decoder, and Output Transformation}—so that architectural effects can be isolated and recombined in a controlled manner.
Grounded in the principles of \textit{Evaluatology}~\cite{zhan2025evaluatology}, we characterize evaluation as a stochastic process by defining a shared Evaluation Condition Space $\Omega$, which captures variations in complementary modules and training configurations.
By performing stratified Monte Carlo sampling over $\Omega$, we estimate the expected effectiveness and stability of each component, enabling robust attribution that is resilient to configuration noise.

Through this rigorous probabilistic audit, we present three empirical findings that challenge prevailing intuitions:
\begin{itemize}[wide,itemsep=-0.1em]
    \item[\ding{182}] The ``Identity'' Paradox (data view (Embedding) $>$ modeling module (Encoder)):
    Within the explored condition space $\Omega$, we observe that once the \textit{data view} (Embedding) is well-designed, a parameter-free \textbf{Identity Encoder} frequently achieves comparable or even superior effectiveness and stability, outperforming complex Transformers and Mixers. This suggests that for many standard benchmarks, the primary performance bottleneck lies in data view formulation rather than modeling module.

    \item[\ding{183}] Conditional Utility of Structural Priors (Input Transformation):
    We find that Input Transformations are not universally superior. While explicit \textit{periodic inductive biases} (e.g., learnable cycles \cite{lin2024cyclenet}) provide consistent gains, decomposition strategies (e.g., multi-scale splitting) only outperform the RevIN baseline when coupled with sophisticated \textit{cross-component interaction}. Naive decomposition often degrades performance, highlighting the necessity of architectural synergy.

    \item[\ding{184}] Effectiveness Gain of Spectral Modeling:
    Our results reveal that frequency-domain modeling (e.g., Fourier/Wavelet) yields higher effectiveness ($\mu$) than temporal baselines, yet provides no significant reduction in instability ($\sigma$), suggesting that the spectral advantage stems from better signal representation rather than robustness to hyperparameter variation.
\end{itemize}

Our work offers a standardized protocol for architectural auditing: (i). \textbf{Framework}: We propose CombinationTS, a modular framework that decouples the TSF pipeline into five interchangeable stages, enabling the combinatorial analysis of $100+$ architectural variants.
(ii). \textbf{Protocol}: We establish a probabilistic evaluation protocol based on EC-sampling, shifting the metric from fragile point estimates (MSE) to robust distributional statistics ($\mu, \sigma$). (iii). \textbf{Systematic Evaluation}: We provide a systematic evaluation of modern components, identifying the ``Identity Paradox'' and establishing a new burden of proof for future architectural complexity.
(iv). \textbf{Open Source}: We release the modular library and the full log of experimental configurations to facilitate reproducible component-level research.

\section{Related Work}
\label{sec:related_work}

\paragraph{Evolution of TSF Architectures.}
The landscape of Time-Series Forecasting (TSF) has shifted from rigorous architectural complexity to a focus on data-centric mechanisms.
Early Transformer-based models, such as \textbf{Informer} \cite{zhou2021informer} and \textbf{Autoformer} \cite{wu2021autoformer}, focused on optimizing the Encoder efficiency via sparse attention or decomposition.
However, the emergence of \textbf{DLinear} \cite{zeng2023transformers-dlinear} challenged this direction, demonstrating that simple linear layers combined with trend-seasonal decomposition could achieve state-of-the-art performance.
Subsequently, the focus shifted to the \emph{data view} (Embedding): \textbf{PatchTST} \cite{nie2022time-patchtst} introduced patch-wise tokenization to capture local semantics, while \textbf{iTransformer} \cite{liu2023itransformer} proposed an inverted variate-wise embedding to explicitly model multivariate correlations; \textbf{DUET} \cite{qiu2025duet} further enhances this through dual clustering, and attention modulation~\cite{liu2026timeformer} improves Transformer expressiveness.
Most recently, architectures like \textbf{TimeMixer} \cite{wang2024timemixer} and \textbf{CycleNet} \cite{lin2024cyclenet} further emphasize the role of Input Transformations—incorporating multi-scale downsampling and explicit periodicity—over deep Encoder design.
Complementarily, decomposition-guided objectives~\cite{qiu2025dbloss} and simplicity-first baselines~\cite{liu2026apn,zeng2023transformers-dlinear} challenge the necessity of architectural complexity.

In this work, rather than proposing a new architecture, we select these representative models not as monolithic competitors, but as exemplars of distinct functional modules—Input Transformation, Embedding, Encoder, Decoder and Output Transformation—to rigorously dissect the true source of their effectiveness.

\paragraph{Benchmarking and Evaluation Paradigms.}
The ``reproducibility crisis'' in machine learning has prompted a critical re-examination of how forecasting models are evaluated.
While libraries like \textbf{TSLib} \cite{wang2024tssurvey-tslib}, \textbf{BasicTS} \cite{shao2024exploring-basicts}, \textbf{TFB}~\cite{qiu2024tfb}, and \textbf{TAB}~\cite{qiu2025tab} have standardized implementation interfaces for time series forecasting, recent studies highlight that code unification alone is insufficient to guarantee rigorous comparison.
Chen et al. \yrcite{chencloser} scrutinize the inner workings of Transformers, revealing that performance gains are frequently misattributed to complex reasoning mechanisms when they actually stem from simple \emph{data view} choices (e.g., intra-variate modeling).
Furthermore, Brigato et al. \yrcite{brigato2025position} explicitly argue that \textit{``There are no Champions''} in long-term forecasting. Through an exhaustive audit of over 3,500 models, they demonstrate that ``SOTA'' claims are statistically fragile, often flipping with minor changes in evaluation contexts.
These findings underscore a fundamental flaw in the current ``Leaderboard'' paradigm: ranking models based on point estimates (e.g., a single random seed) is scientifically untenable.

Addressing this, our work adopts the principles of \textbf{\textit{Evaluatology}} \cite{zhan2025evaluatology} to shift the focus from ranking models to \textbf{auditing mechanisms}.
Instead of relying on fragile point comparisons, CombinationTS employs a Probabilistic Evaluation Protocol. By treating hyperparameters and experimental settings as stochastic Evaluation Conditions (EC) sampled from a broad distribution, we quantify the \textit{marginalized performance} ($\mu$) and \textit{stability} ($\sigma$) of specific components, establishing a statistically defensible framework to measure the effectiveness-stability trade-off.

\section{The CombinationTS Framework}
\label{sec:method}

To transition from ``Model Selection'' to ``Modular Understanding,'' we must simultaneously address two fundamental methodological deficits in current research:

\textbf{The Attribution Gap.}
The monolithic treatment of forecasting models obscures the source of performance gains, making it difficult to determine whether success stems from a novel encoder or simply a superior data view.

\textbf{The Benchmarking Crisis.}
The reliance on point estimates leads to two fundamental traps, yielding conclusions that are statistically fragile and often unreproducible.
First, the \textit{``Fairness'' Trap}: constraining all models to a fixed setting may fail to activate the effective operating regime of specific architectures, systematically underestimating their potential.
Second, the \textit{``Best'' Trap}: reporting the best observed result is statistically fragile, as peak performance may arise from incidental hyperparameter alignment or randomness, thereby misrepresenting expected performance in practice.

Our framework, \textbf{CombinationTS}, mitigates these issues by intersecting \textbf{Modular Decomposition} (addressing the attribution gap) with a \textbf{Probabilistic Evaluation Protocol} (addressing the benchmarking crisis).

\subsection{Modular Decomposition for Structural Attribution}
\label{sec:modular_decomp}

We consider the multivariate time-series forecasting problem, mapping historical observations $\mathbf{X} \in \mathbb{R}^{T \times N}$ to future predictions $\mathbf{Y} \in \mathbb{R}^{P \times N}$ via a parameterized function $f$.
To systematically dissect the internal mechanisms of $f$ rather than treating it as a monolithic ``black box,'' we formalize the forecasting pipeline as a composite of five stages:
{
    \setlength{\abovedisplayskip}{3pt}
    \setlength{\belowdisplayskip}{3pt}
    \begin{equation}
    \label{eq:modular_func}
    f = \mathcal{T}^{-1}_{out} \circ \mathcal{D} \circ \varPhi \circ \mathcal{E} \circ \mathcal{T}_{in}.
    \end{equation}
}
Consequently, the forward pass is expressed as
{
    \setlength{\abovedisplayskip}{3pt}
    \setlength{\belowdisplayskip}{3pt}
    \begin{equation}
        \hat{\mathbf{Y}} = f(\mathbf{X})
        = (\mathcal{T}^{-1}_{out} \circ \mathcal{D} \circ \varPhi \circ \mathcal{E} \circ \mathcal{T}_{in})(\mathbf{X}).
    \end{equation}
}

This formulation allows us to isolate specific components for analysis:

\paragraph{Input Transformation ($\mathcal{T}_{in}$).}
Injects structural priors into the raw signal (e.g., normalization, detrending, or trend-seasonal decomposition).
\textit{Operationally}, $\mathcal{T}_{in}$ is restricted to transformations applied directly on $\mathbf{X}$ before tokenization and admits an inverse $\mathcal{T}^{-1}_{out}$ when applicable.

\paragraph{Embedding ($\mathcal{E}$).}
Defines the \textbf{data view} (tokenization strategy).
To unify diverse views (e.g., variate tokens vs. temporal patches), we standardize the latent representation into a unified tensor interface
$\mathcal{Z}=\mathcal{E}(\mathcal{T}_{in}(\mathbf{X})) \in \mathbb{R}^{B \times C \times L \times D}$,
where $B$ is the batch size, $C$ denotes the \textbf{spatial dimension} (number of variates/channels), $L$ denotes the \textbf{temporal dimension} (number of temporal tokens), and $D$ is the hidden channel size.
\textit{Crucially}, $\mathcal{E}$ only determines the view and local within-token encoding, while any cross-token dependency modeling is delegated to the encoder $\varPhi$.

\paragraph{Encoder ($\varPhi$).}
The modeling module operating on $\mathcal{Z}$ (e.g., self-attention, MLP-mixer), responsible for capturing interactions among tokens and/or within-token contexts.

\paragraph{Decoder ($\mathcal{D}$).}
Projects the processed latent features to the forecasting horizon, e.g., mapping encoder outputs to $\mathbb{R}^{P \times N}$ (or an intermediate horizon-aligned representation).

\paragraph{Output Transformation ($\mathcal{T}^{-1}_{out}$).}
The inverse operation of $\mathcal{T}_{in}$ (when applicable), e.g., adding back a removed trend component or applying inverse normalization to recover the original scale.

\subsection{Probabilistic Evaluation for Robust Benchmarking}
\label{sec:prob_eval}

Standard benchmarking practices in time-series forecasting typically fix a single configuration to report a single-value performance metric.
Such point-estimate–based evaluation suffers from two fundamental flaws.
First, the \textit{``Fairness'' Trap}: constraining all models to a fixed setting may fail to activate the effective operating regime of specific architectures, systematically underestimating their potential.
Second, the \textit{``Best'' Trap}: reporting the best observed result is statistically fragile, as peak performance may arise from incidental hyperparameter alignment or randomness, thereby misrepresenting expected performance in practice.

To overcome these limitations, we build a self-contained evaluation system that treats architectural analysis as a signal--noise disentanglement problem: the \emph{signal} is the component whose effect we aim to attribute, while the \emph{noise} arises from evaluation conditions.
Grounded in Evaluatology~\cite{zhan2025evaluatology}, this view naturally leads to a probabilistic protocol---we move beyond point estimates and instead compare performance distributions induced by evaluation environments under a shared condition space.

Concretely, we define the evaluation system through two complementary entities:

\paragraph{Evaluated Object.}\footnote{\textbf{EO}: We denote the evaluated object, i.e., the architectural component whose contribution is being analyzed, by $\theta$.}
The deterministic signal we aim to attribute, corresponding to the module variant under investigation at a specific stage of the forecasting pipeline (e.g., $\theta=\textit{Variate-wise Embedding}$).

\paragraph{Evaluation Condition.}\footnote{\textbf{EC}: Throughout this paper, we denote the evaluation condition by $\mathbf{c}$.}
The stochastic environment under which an EO operates.
It encapsulates \textit{all} factors that may influence the observed performance once $\theta$ is fixed, including (i) complementary module instantiations in the remaining stages, (ii) training hyperparameters, and (iii) sources of randomness.
Concretely, $\mathbf{c}$ includes (but is not limited to) random seeds, batch size, dropout rate, learning rate schedules, initialization, and other training or evaluation choices.
We treat $\mathbf{c}$ as a random variable sampled from a broad evaluation condition space $\Omega$, rather than as a fixed configuration or a quantity to be optimized away.

\paragraph{Statistical Attribution Target.}
Under this formulation, the performance of a module $\theta$ is modeled as a random variable $L(\theta,\mathbf{c})$ governed by $\mathbf{c}\sim\Omega$.
We quantify the contribution of $\theta$ using two distributional statistics:
{
    \setlength{\abovedisplayskip}{3pt}
    \setlength{\belowdisplayskip}{3pt}
    \begin{equation}
        \mu(\theta)=\mathbb{E}_{\mathbf{c}\sim\Omega}[L(\theta,\mathbf{c})],\quad
        \sigma(\theta)=\sqrt{\mathrm{Var}_{\mathbf{c}\sim\Omega}[L(\theta,\mathbf{c})]},
    \end{equation}
}
where $\mu$ denotes \textbf{Marginalized Performance}, capturing expected performance across evaluation conditions, and $\sigma$ denotes \textbf{Stability}, measuring sensitivity to such conditions.
A superior module is therefore expected to improve $\mu$ without incurring a disproportionate increase in $\sigma$.

\subsection{Stratified Monte Carlo Estimation}
\label{sec:mc_protocol}

To obtain robust and comparable estimates, we adopt a \textbf{stratified} and \textbf{paired} Monte Carlo protocol.
We construct a fixed evaluation condition set $\{\mathbf{c}_k\}_{k=1}^{K}$ sampled from $\Omega$, stratified across datasets and prediction horizons.
Crucially, to eliminate confounding factors arising from stochastic sampling, we employ a \textbf{paired design}: every Evaluated Object (EO) is assessed under the \textit{exact same} set of sampled conditions.

Based on these $K$ shared samples, we report three key statistics for each module $\theta$:

\paragraph{Marginalized Performance ($\hat{\mu}$).}
The sample mean of the loss over the sampled conditions:
{
\setlength{\abovedisplayskip}{3pt}
\setlength{\belowdisplayskip}{3pt}
\begin{equation}
    \hat{\mu}(\theta) = \frac{1}{K} \sum_{k=1}^{K} L(\theta, \mathbf{c}_k).
\end{equation}
}

\paragraph{Stability ($\hat{\sigma}$).}
The sample standard deviation of the loss over the sampled conditions:
{
\setlength{\abovedisplayskip}{3pt}
\setlength{\belowdisplayskip}{3pt}
\begin{equation}
    \hat{\sigma}(\theta) = \sqrt{ \frac{1}{K-1} \sum_{k=1}^{K} \left( L(\theta, \mathbf{c}_k) - \hat{\mu}(\theta) \right)^2 }.
\end{equation}
}

\paragraph{Peak Potential ($L_{best}$).}
To explicitly contrast with the ``Best Trap'', we also report the single best performance observed during sampling:
{
\setlength{\abovedisplayskip}{3pt}
\setlength{\belowdisplayskip}{3pt}
\begin{equation}
    L_{best}(\theta) = \min_{k} L(\theta, \mathbf{c}_k).
\end{equation}
}
By reporting both the expected ($\hat{\mu}$) and peak ($L_{best}$) performance, we can rigorously determine whether a model's SOTA claim is a generalized capability or a rare outlier.

\section{Experiments}
\label{sec:experiments}

In this section, we transition from theoretical formulation to empirical attribution.
Applying the \textbf{CombinationTS} framework (illustrated in Figure~\ref{fig:fig1}), we systematically deconstruct the ``black box'' of modern time-series forecasting models to isolate the true drivers of performance.
Our experimental suite is built upon a \textbf{highly customized version} of the Time Series Library (TSLib) \cite{wang2024tssurvey-tslib}, extensively modified to enable the \textbf{combinatorial assembly} of decoupled modules—serving as the implementation engine for \textbf{CombinationTS}.
We conduct extensive evaluations across \textbf{six} widely adopted benchmarks: \textbf{Weather}, \textbf{Electricity}, and the four subsets of the \textbf{ETT} dataset (ETTh1, ETTh2, ETTm1, ETTm2).

Rather than chasing State-of-the-Art (SOTA) on a fixed leaderboard, our experiments are designed to answer three fundamental questions that directly address the \textbf{Attribution Gap} and \textbf{Benchmarking Crisis} identified in the Introduction:

\textbf{RQ1: Dissecting the Backbone (\hyperref[exp:backbone]{Exp. 1}).}
Which drives forecasting performance—data view (Embedding $\mathcal{E}$) or modeling module (Encoder $\varPhi$)?
We evaluate all pairwise ($\mathcal{E}$, $\varPhi$) combinations under a unified probabilistic protocol, measuring $\hat{\mu}$, $\hat{\sigma}$, and $L_{best}$.

\textbf{RQ2: Auditing Input Transformations (\hyperref[exp:input_transformation]{Exp. 2}).}
Can Input Transformations ($\mathcal{T}_{in}$)—decomposition, downsampling, or cyclic embeddings—substitute for complex encoders?
We compare four representative strategies across an expanded EC space ($D \in \{16, \ldots, 512\}$) to test whether Input Transformations enable efficient forecasting without over-parameterization.

\textbf{RQ3: Revisiting Spectral Processing (\hyperref[exp:spectral]{Exp. 3}).}
Does frequency-domain modeling offer performance advantages over time-domain processing, and does it also improve stability?
We conduct a controlled pair (SimpleTM vs. iTransformer) under matched Variate-wise Embedding to determine whether spectral processing provides genuine gains in $\mu$ or $\sigma$ beyond what the Identity baseline already achieves.

\paragraph{Evaluation Protocol.}
We adopt MSE as the primary metric $L$ and report three statistics: \textbf{Marginalized Performance ($\hat{\mu}$)}, \textbf{Stability ($\hat{\sigma}$)}, and \textbf{Peak Potential ($L_{best}$)}.
To isolate configurational robustness from initialization noise, the random seed is fixed across all runs and dropout is set to $0.1$.
Multi-seed verification confirms that this choice does not affect our conclusions (\cref{sec:multiseed}).
Training is standardized at batch size $32$, $30$ epochs, and early-stopping patience $3$, ensuring any observed variance is attributable solely to architectural and hyperparameter choices.

\subsection{Dissecting the Backbone: The Hierarchy of View and Representation}
\label{exp:backbone}

Building upon the modular decomposition in Eq.~\ref{eq:modular_func}, we focus on the architectural ``backbone'' of modern forecasting models: the interplay between \textbf{data view (Embedding $\mathcal{E}$)} and \textbf{modeling module (Encoder $\varPhi$)}.

\paragraph{Controlled Factors.}
To isolate the backbone effect, we fix the Input and Output Transformations ($\mathcal{T}_{in}, \mathcal{T}_{out}^{-1}$) to \textbf{RevIN}~\cite{kim2021reversible-RevIN} for all models, and systematically evaluate combinations of \textbf{Embedding} and \textbf{Encoder}, together with a lightweight \textbf{Decoder}.
Importantly, \textbf{each encoder is evaluated on \emph{all} embedding strategies} under the same probabilistic protocol (paired EC sampling), avoiding selective reporting.

\subsubsection{Experiment Setup}
\label{sec:exp1_setup}

\paragraph{Evaluated Objects.}
We construct a modular search space for backbone analysis:
\[
\Theta_{\text{backbone}} \;=\; \mathcal{E} \times \varPhi \times \mathcal{D}.
\]
A specific EO is a concrete module combination $\theta \in \Theta_{\text{backbone}}$.
Table~\ref{tab:exp1_setup} summarizes all EO variants and their corresponding tensor transformations
(\emph{omitting batch dimension for simplicity}; the implementation follows a standard batch-first layout).

\begin{table*}[t]
    \centering
    \small
    \caption{\textbf{Search Space for Experiment 1 (Backbone Analysis).}
We deconstruct the forecasting model into three modular stages.
The \textbf{Tensor Transformation} column denotes the shape change from Input to Output (assuming input $\mathbf{X} \in \mathbb{R}^{B \times N \times T \times 1}$, output $\mathbf{Y} \in \mathbb{R}^{B \times N \times P \times 1}$), where $B$: batch size, $N$: variates, $T$: look-back length, $P$: prediction length, $S$: patch embedding stride, and $D$: latent dim. $N'$, $T'$, $D'$ depend on the embedding choice. For brevity, the $B$ dimension is omitted in the table.
}
    \label{tab:exp1_setup}
    \setlength{\tabcolsep}{8pt}
    \renewcommand{\arraystretch}{1.2}
    \resizebox{\textwidth}{!}{
            \begin{tabular}{c|l|l|l}
    \toprule
    \textbf{Module (EO)} & \textbf{Variant} & \textbf{Mechanism Description} & \textbf{Tensor Transformation} \\
    \midrule
    \multirow{5}{*}{\textbf{Embedding} ($\mathcal{E}$)}
        & Point-wise & Projects each time step independently & $\mathbb{R}^{N \times T \times 1} \to \mathbb{R}^{N \times T \times D}$ \\
        & Patch-wise & Aggregates local segments into tokens & $\mathbb{R}^{N \times T \times 1} \to \mathbb{R}^{N \times \lceil T/S \rceil \times D}$ \\
        & Variate-wise & Treats each variate history as one token & $\mathbb{R}^{N \times T \times 1} \to \mathbb{R}^{N \times 1 \times D}$ \\
        & Identity & Preserves raw inputs (No projection) & $\mathbb{R}^{N \times T \times 1} \to \mathbb{R}^{N \times T \times 1}$ \\
        & Time-as-Feature (\textit{T$\to$D} reshape) & Reshapes temporal/variate dim to feature (\textbf{parameter-free}) & $\mathbb{R}^{N \times T \times 1} \to \mathbb{R}^{N \times 1 \times T}\;\text{(or equivalent view)}$ \\
    \midrule
    \multirow{3}{*}{\textbf{Encoder} ($\varPhi$)}
        & Transformer & Self-Attention & \multirow{3}{*}{$\mathbb{R}^{N' \times T' \times D'} \to \mathbb{R}^{N' \times T' \times D'}$} \\
        & MLP & Global mixing across dimensions & \\
        & Identity & Pass-through baseline & \\
    \midrule
    \textbf{Decoder} ($\mathcal{D}$)
        & Shared Linear & Shared weights across variates & $\mathbb{R}^{N' \times T' \times D'} \to \mathbb{R}^{N \times P \times 1}$ \\
    \bottomrule
    \end{tabular}
    }
\end{table*}

\paragraph{Evaluation Conditions.}
To estimate $\mu(\theta)$ and $\sigma(\theta)$, we sample EC from a condition space $\Omega$.
Each condition $c\in\Omega$ specifies: dataset $d\in\{$Weather, Electricity, ETTh1, ETTh2, ETTm1, ETTm2$\}$, look-back $T\in\{96,192,336,512\}$, horizon $P\in\{96,192,336,720\}$, encoder layers $el\in\{1,2,3\}$, latent dim $D\in\{64,128,256,512\}$, and learning rate $\eta\in\{1\text{e-}3,1\text{e-}4\}$; RevIN~\cite{kim2021reversible-RevIN} and a fixed seed are applied uniformly.
We construct a fixed set $\Omega_{\text{eval}}=\{c_k\}_{k=1}^{K}$ with $K=600$ stratified ECs (100 per dataset); all EOs share the same paired EC set for fair comparison, with inactive dimensions ignored.

\subsubsection{Results and Analysis}
\label{sec:exp1_results}

\begin{table}[t]
    \centering
    \setlength{\tabcolsep}{1.5pt}
    \renewcommand{\arraystretch}{1.1}
    \caption{Performance comparison on embeddings.
    For each metric ($\hat{\mu}$ and $L_{best}$): \bestb{blue} = best, \secb{blue} = second best; \bestr{red} = best, \secr{red} = second best.
    Detailed results are reported in \cref{tab:exp1_emb_data}.
}
    \label{tab:exp1_emb_sub_data}
    \resizebox{\linewidth}{!}{
        \begin{tabular}{l c | ccc | ccc | ccc | ccc}
    \toprule
    \multirow{2}{*}{\textbf{Dataset}} & \multirow{2}{*}{\textbf{H}} & 
    \multicolumn{3}{c|}{\textbf{Point wise}} &
    \multicolumn{3}{c|}{\textbf{Patch wise}} &
    \multicolumn{3}{c|}{\textbf{Variate wise}} &
    \multicolumn{3}{c}{\textbf{L to dim}} \\
    \cmidrule(lr){3-5} \cmidrule(lr){6-8} \cmidrule(lr){9-11} \cmidrule(lr){12-14}
    & & $\hat{\mu}$ & $\hat{\sigma}$ & $min$ & $\hat{\mu}$ & $\hat{\sigma}$ & $min$ & $\hat{\mu}$ & $\hat{\sigma}$ & $min$ & $\hat{\mu}$ & $\hat{\sigma}$ & $min$ \\
    \midrule

    \multirow{5}{*}{\textbf{ETTh1}}

    & 96 & 0.3898 & 0.0143 & 0.3651 & \textcolor{darkblue}{\textbf{0.3772}} & 0.0058 & \textcolor{darkred}{\textbf{\underline{0.3645}}} & \textcolor{darkblue}{\textbf{\underline{0.3834}}} & 0.0080 & \textcolor{darkred}{\textbf{0.3641}} & 0.3937 & 0.0156 & 0.3721 \\

    & 192 & 0.4324 & 0.0166 & 0.3992 & \textcolor{darkblue}{\textbf{0.4240}} & 0.0133 & \textcolor{darkred}{\textbf{\underline{0.3984}}} & \textcolor{darkblue}{\textbf{\underline{0.4247}}} & 0.0128 & \textcolor{darkred}{\textbf{0.3983}} & 0.4343 & 0.0161 & 0.4028 \\

    & 336 & 0.4651 & 0.0193 & 0.4255 & \textcolor{darkblue}{\textbf{0.4575}} & 0.0170 & \textcolor{darkred}{\textbf{0.4196}} & \textcolor{darkblue}{\textbf{\underline{0.4589}}} & 0.0148 & \textcolor{darkred}{\textbf{\underline{0.4249}}} & 0.4693 & 0.0160 & 0.4294 \\

    & 720 & 0.4772 & 0.0415 & \textcolor{darkred}{\textbf{0.4248}} & 0.4812 & 0.0326 & 0.4366 & \textcolor{darkblue}{\textbf{0.4618}} & 0.0140 & \textcolor{darkred}{\textbf{\underline{0.4366}}} & \textcolor{darkblue}{\textbf{\underline{0.4751}}} & 0.0223 & 0.4409 \\

    & avg & 0.4367 & 0.0364 & - & \textcolor{darkblue}{\textbf{0.4277}} & 0.0370 & - & \textcolor{darkblue}{\textbf{\underline{0.4285}}} & 0.0327 & - & 0.4398 & 0.0342 & - \\
    \midrule
    
    \multirow{5}{*}{\textbf{ETTm2}}

    & 96 & 0.1800 & 0.0089 & \textcolor{darkred}{\textbf{\underline{0.1625}}} & \textcolor{darkblue}{\textbf{0.1728}} & 0.0051 & 0.1631 & \textcolor{darkblue}{\textbf{\underline{0.1758}}} & 0.0072 & \textcolor{darkred}{\textbf{0.1622}} & 0.1779 & 0.0073 & 0.1633 \\

    & 192 & 0.2414 & 0.0136 & \textcolor{darkred}{\textbf{0.2170}} & \textcolor{darkblue}{\textbf{0.2331}} & 0.0079 & \textcolor{darkred}{\textbf{\underline{0.2179}}} & \textcolor{darkblue}{\textbf{\underline{0.2343}}} & 0.0106 & 0.2184 & 0.2370 & 0.0115 & 0.2180 \\

    & 336 & 0.3019 & 0.0172 & \textcolor{darkred}{\textbf{0.2701}} & \textcolor{darkblue}{\textbf{0.2911}} & 0.0105 & 0.2732 & \textcolor{darkblue}{\textbf{\underline{0.2928}}} & 0.0135 & \textcolor{darkred}{\textbf{\underline{0.2707}}} & 0.2968 & 0.0134 & 0.2711 \\

    & 720 & 0.3981 & 0.0216 & \textcolor{darkred}{\textbf{0.3591}} & \textcolor{darkblue}{\textbf{0.3813}} & 0.0140 & 0.3608 & \textcolor{darkblue}{\textbf{\underline{0.3863}}} & 0.0170 & \textcolor{darkred}{\textbf{\underline{0.3594}}} & 0.3892 & 0.0174 & 0.3603 \\

    & avg & 0.2663 & 0.0762 & - & \textcolor{darkblue}{\textbf{0.2562}} & 0.0725 & - & \textcolor{darkblue}{\textbf{\underline{0.2596}}} & 0.0723 & - & 0.2612 & 0.0728 & - \\
    \midrule
    
    \multirow{5}{*}{\textbf{Electricity}}

    & 96 & 0.1609 & 0.0228 & 0.1431 & \textcolor{darkblue}{\textbf{0.1512}} & 0.0176 & \textcolor{darkred}{\textbf{0.1315}} & \textcolor{darkblue}{\textbf{\underline{0.1527}}} & 0.0135 & \textcolor{darkred}{\textbf{\underline{0.1335}}} & 0.1573 & 0.0190 & 0.1349 \\

    & 192 & 0.1716 & 0.0154 & 0.1574 & \textcolor{darkblue}{\textbf{0.1641}} & 0.0111 & \textcolor{darkred}{\textbf{0.1465}} & 0.1710 & 0.0111 & 0.1555 & \textcolor{darkblue}{\textbf{\underline{0.1710}}} & 0.0115 & \textcolor{darkred}{\textbf{\underline{0.1542}}} \\

    & 336 & 0.1917 & 0.0170 & 0.1735 & \textcolor{darkblue}{\textbf{0.1828}} & 0.0137 & \textcolor{darkred}{\textbf{0.1644}} & \textcolor{darkblue}{\textbf{\underline{0.1906}}} & 0.0153 & 0.1722 & 0.1928 & 0.0170 & \textcolor{darkred}{\textbf{\underline{0.1706}}} \\

    & 720 & 0.2342 & 0.0194 & 0.2132 & \textcolor{darkblue}{\textbf{\underline{0.2236}}} & 0.0160 & 0.2004 & 0.2293 & 0.0205 & \textcolor{darkred}{\textbf{0.1972}} & \textcolor{darkblue}{\textbf{0.2218}} & 0.0141 & \textcolor{darkred}{\textbf{\underline{0.1991}}} \\

    & avg & 0.1855 & 0.0252 & - & \textcolor{darkblue}{\textbf{0.1773}} & 0.0246 & - & \textcolor{darkblue}{\textbf{\underline{0.1827}}} & 0.0249 & - & 0.1851 & 0.0264 & - \\
    
    \bottomrule
\end{tabular}

    }
\end{table}

\begin{figure}[ht]
    \centering
    \includegraphics[width=\columnwidth]{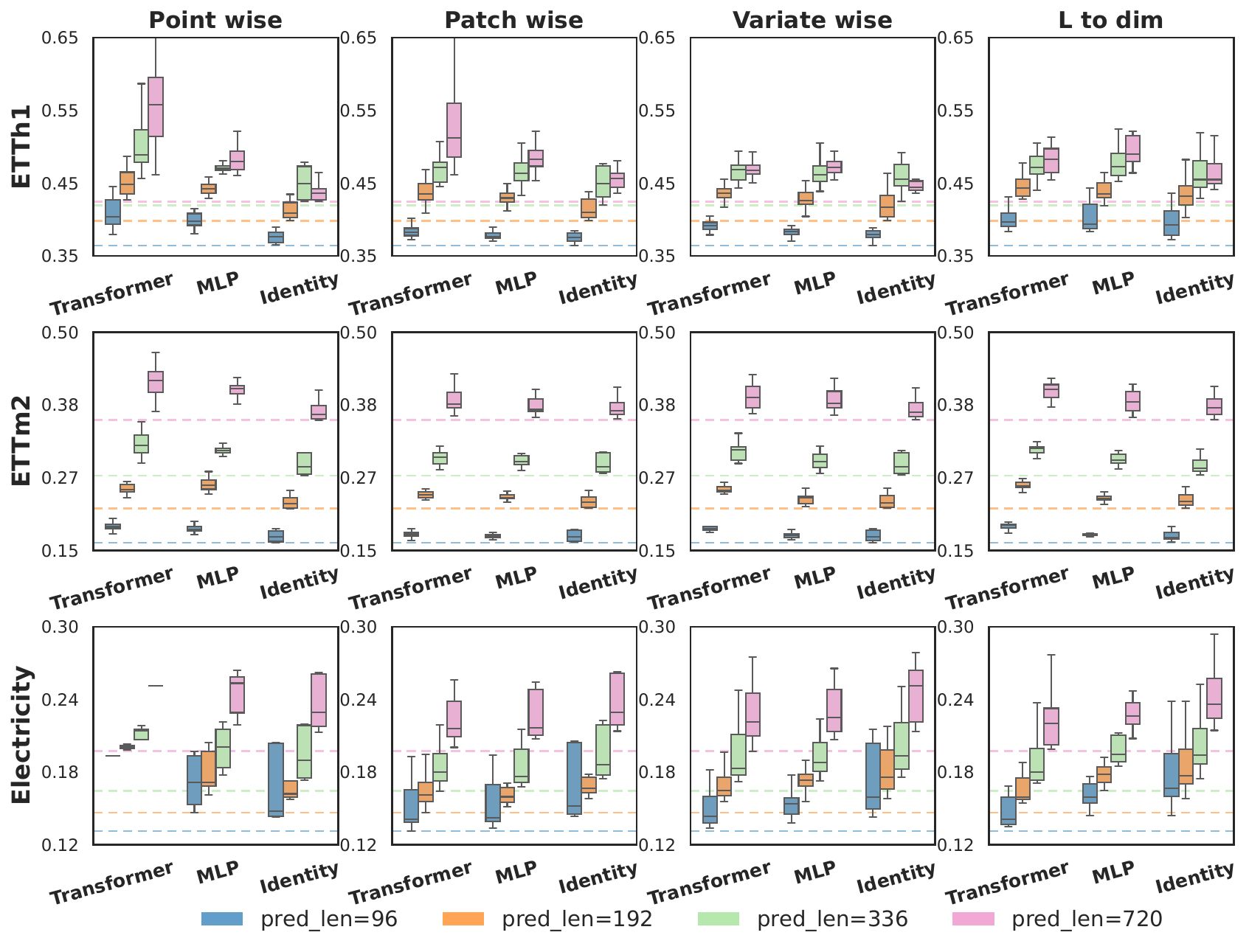}
    \caption{Distribution of encoder effectiveness under paired EC sampling (boxplot). The full figure is presented in \cref{fig:exp1_enc_effect}.}
    \label{fig:exp1_sub_enc_effect}
\end{figure}

\begin{table}[t]
    \centering
    \scriptsize
    \setlength{\tabcolsep}{1.5pt}
    \renewcommand{\arraystretch}{1.1}
    \caption{
        Performance of 6 datasets across 4 horizons on 3 encoders (paired EC evaluation).
        Highlighting follows the convention in \cref{tab:exp1_emb_sub_data}.
        Detailed results are reported in \cref{tab:exp1_enc_data}.
    }
    \label{tab:exp1_enc_sub_data}
    \resizebox{\columnwidth}{!}{
        \begin{tabular}{l c | ccc | ccc | ccc}
    \toprule
    \multirow{2}{*}{\textbf{Dataset}} & \multirow{2}{*}{\textbf{H}} &
    \multicolumn{3}{c|}{\textbf{MLP}} &
    \multicolumn{3}{c|}{\textbf{Transformer}} &
    \multicolumn{3}{c}{\textbf{Identity}} \\
    \cmidrule(lr){3-5} \cmidrule(lr){6-8} \cmidrule(lr){9-11}
    & & $\hat{\mu}$ & $\hat{\sigma}$ & $min$ & $\hat{\mu}$& $\hat{\sigma}$ & $min$ & $\hat{\mu}$ & $\hat{\sigma}$ & $min$ \\
    \midrule

    \multirow{5}{*}{\textbf{ETTh1}}
    & 96 & \textcolor{darkblue}{\textbf{\underline{0.4371}}} & 0.0816 & \textcolor{darkred}{\textbf{\underline{0.3700}}} & 0.4518 & 0.0930 & 0.3721 & \textcolor{darkblue}{\textbf{0.3907}} & 0.0231 & \textcolor{darkred}{\textbf{0.3641}} \\
    & 192 & \textcolor{darkblue}{\textbf{\underline{0.4870}}} & 0.0827 & \textcolor{darkred}{\textbf{\underline{0.4043}}} & 0.5044 & 0.0988 & 0.4091 & \textcolor{darkblue}{\textbf{0.4334}} & 0.0306 & \textcolor{darkred}{\textbf{0.3983}} \\
    & 336 & \textcolor{darkblue}{\textbf{\underline{0.5186}}} & 0.0787 & \textcolor{darkred}{\textbf{\underline{0.4335}}} & 0.5239 & 0.0795 & 0.4406 & \textcolor{darkblue}{\textbf{0.4684}} & 0.0268 & \textcolor{darkred}{\textbf{0.4196}} \\
    & 720 & \textcolor{darkblue}{\textbf{\underline{0.5538}}} & 0.1039 & 0.4533 & 0.5679 & 0.1014 & \textcolor{darkred}{\textbf{\underline{0.4504}}} & \textcolor{darkblue}{\textbf{0.4722}} & 0.0406 & \textcolor{darkred}{\textbf{0.4248}} \\
    & avg & \textcolor{darkblue}{\textbf{\underline{0.4960}}} & 0.0938 & - & 0.5095 & 0.1004 & - & \textcolor{darkblue}{\textbf{0.4392}} & 0.0424 & - \\
    \cmidrule{1-11}

    \multirow{5}{*}{\textbf{ETTm2}}
    & 96 & \textcolor{darkblue}{\textbf{\underline{0.1893}}} & 0.0197 & 0.1667 & 0.2027 & 0.0265 & \textcolor{darkred}{\textbf{\underline{0.1652}}} & \textcolor{darkblue}{\textbf{0.1755}} & 0.0087 & \textcolor{darkred}{\textbf{0.1622}} \\
    & 192 & \textcolor{darkblue}{\textbf{\underline{0.2525}}} & 0.0213 & \textcolor{darkred}{\textbf{\underline{0.2201}}} & 0.2610 & 0.0221 & 0.2313 & \textcolor{darkblue}{\textbf{0.2341}} & 0.0133 & \textcolor{darkred}{\textbf{0.2170}} \\
    & 336 & \textcolor{darkblue}{\textbf{\underline{0.3101}}} & 0.0198 & \textcolor{darkred}{\textbf{\underline{0.2738}}} & 0.3202 & 0.0199 & 0.2794 & \textcolor{darkblue}{\textbf{0.2956}} & 0.0194 & \textcolor{darkred}{\textbf{0.2701}} \\
    & 720 & \textcolor{darkblue}{\textbf{\underline{0.4057}}} & 0.0232 & \textcolor{darkred}{\textbf{\underline{0.3634}}} & 0.4163 & 0.0265 & 0.3653 & \textcolor{darkblue}{\textbf{0.3919}} & 0.0270 & \textcolor{darkred}{\textbf{0.3591}} \\
    & avg & \textcolor{darkblue}{\textbf{\underline{0.2784}}} & 0.0759 & - & 0.2909 & 0.0772 & - & \textcolor{darkblue}{\textbf{0.2621}} & 0.0756 & - \\
    \cmidrule{1-11}

    \multirow{5}{*}{\textbf{Electricity}}
    & 96 & \textcolor{darkblue}{\textbf{\underline{0.1599}}} & 0.0171 & \textcolor{darkred}{\textbf{\underline{0.1338}}} & \textcolor{darkblue}{\textbf{0.1535}} & 0.0194 & \textcolor{darkred}{\textbf{0.1315}} & 0.1679 & 0.0254 & 0.1428 \\
    & 192 & \textcolor{darkblue}{\textbf{\underline{0.1729}}} & 0.0130 & \textcolor{darkred}{\textbf{\underline{0.1515}}} & \textcolor{darkblue}{\textbf{0.1671}} & 0.0129 & \textcolor{darkred}{\textbf{0.1465}} & 0.1742 & 0.0161 & 0.1574 \\
    & 336 & \textcolor{darkblue}{\textbf{\underline{0.1925}}} & 0.0158 & \textcolor{darkred}{\textbf{\underline{0.1681}}} & \textcolor{darkblue}{\textbf{0.1858}} & 0.0153 & \textcolor{darkred}{\textbf{0.1644}} & 0.1939 & 0.0179 & 0.1735 \\
    & 720 & \textcolor{darkblue}{\textbf{\underline{0.2282}}} & 0.0171 & \textcolor{darkred}{\textbf{\underline{0.2070}}} & \textcolor{darkblue}{\textbf{0.2196}} & 0.0171 & \textcolor{darkred}{\textbf{0.1972}} & 0.2326 & 0.0191 & 0.2132 \\
    & avg & \textcolor{darkblue}{\textbf{\underline{0.1851}}} & 0.0253 & - & \textcolor{darkblue}{\textbf{0.1776}} & 0.0242 & - & 0.1888 & 0.0275 & - \\
    \bottomrule
\end{tabular}

    }
\end{table}

\begin{figure}[ht]
    \centering
    \includegraphics[width=\columnwidth]{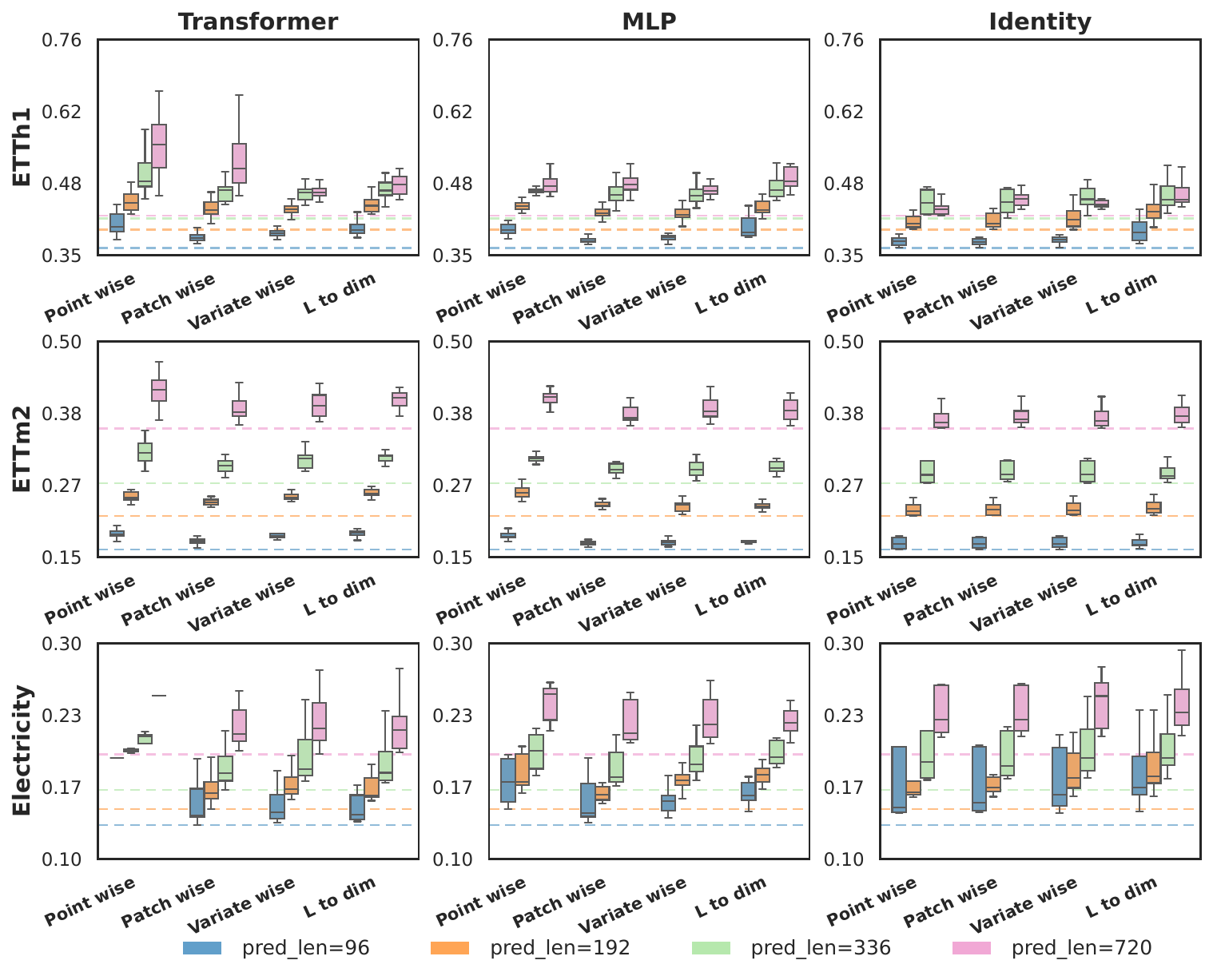}
    \caption{Distribution of embedding effectiveness under paired EC sampling (boxplot). The full figure is presented in \cref{fig:exp1_emb_effect}.}
    \label{fig:exp1_sub_emb_effect}
\end{figure}

\paragraph{Embedding (data view) Analysis.}
Based on Table~\ref{tab:exp1_emb_sub_data} and Figure~\ref{fig:exp1_sub_emb_effect}, we summarize three insights about \textbf{data view} (95\% confidence intervals reported in \cref{tab:overall_results}):

\begin{itemize}[wide,itemsep=-0.1em]
    \item[\ding{182}] \textbf{Robustness of structured tokenization.}
    Patch-wise and Variate-wise embeddings consistently achieve strong marginalized performance ($\hat{\mu}$) with comparatively low instability ($\hat{\sigma}$) across most datasets and horizons.
    Explicit inductive bias in $\mathcal{E}$---either local temporal aggregation (Patch) or tokenizing by variates---stabilizes optimization under hyperparameter variation.

    \item[\ding{183}] \textbf{The ``std trap'' of Point-wise embedding.}
    Point-wise embedding tends to exhibit worse average performance and higher standard deviation on several ETT subsets, yet can occasionally reach a very low $L_{best}$.
    This illustrates a key pitfall of ``best-result'' benchmarking: a volatile design may sporadically hit a favorable configuration despite being unreliable in expectation.

    \item[\ding{184}] \textbf{Efficacy of parameter-free reshaping.}
    The Time-as-Feature (T$\to$D reshape) heuristic (parameter-free) performs on par with more complex learnable views in many settings.
    This suggests that the \emph{structure of the tensor} can matter as much as---and sometimes more than---learnable projections of values.
\end{itemize}

\paragraph{Encoder (modeling module) Analysis.}
Shifting focus to the \textbf{Encoder} (Reasoning), Table~\ref{tab:exp1_enc_sub_data} and Figure~\ref{fig:exp1_sub_enc_effect} reveal the following (95\% confidence intervals reported in \cref{tab:exp1_enc_data_ci}):

\begin{itemize}[wide,itemsep=-0.1em]
    \item[\ding{182}] \textbf{The ``Identity’’ Paradox.}
    Across ETT datasets, \textbf{Transformer-based encoders fail to deliver any statistically significant gain over the parameter-free Identity baseline}, incurring substantially higher instability ($\hat{\sigma}$) without improving marginalized performance ($\hat{\mu}$).
    On standard benchmarks, widely adopted deep modeling modules act primarily as \textbf{over-parameterized noise generators} rather than essential feature extractors.

    \item[\ding{183}] \textbf{Transformer’s dependence on the data view.}
    Transformers exhibit pronounced sensitivity to the embedding stage: substantially more stable under structured tokenization (Patch/Variate-wise), but volatile under unstructured views.
    Self-attention is not a universal feature extractor; it benefits from semantically meaningful tokens.

    \item[\ding{184}] \textbf{MLP Robustness.}
    MLP encoders demonstrate \textbf{consistent stability} comparable to the Identity baseline, offering a reliable middle ground with learnable capacity but without the optimization volatility of attention mechanisms.
\end{itemize}

\paragraph{Robustness of the Identity Paradox.}
We verify that the Identity Paradox is not an artifact of statistical noise or limited evaluation scope through three complementary analyses.
\textbf{(i) Statistical significance.}
One-tailed Mann--Whitney U tests ($\alpha{=}0.05$) confirm that Identity achieves significantly lower MSE than both Transformer ($p{<}0.0001$) and MLP ($p{=}0.0115$) across all datasets combined (\cref{tab:mw_per_dataset,tab:mw_overall}).
\textbf{(ii) Non-stationary generalization.}
On the volatile Exchange-Rate dataset, Identity-based models (PatchLinear) still achieve the lowest average MSE (\cref{tab:exchange_rate}), ruling out benchmark periodicity as a confound.
\textbf{(iii) Hyperparameter and seed stability.}
Identity-based models maintain consistent optimal configurations across horizons, whereas Transformers require substantially different hyperparameters per horizon on volatile datasets (\cref{sec:hparam_tunability}).
Multi-seed verification further confirms that these findings are not seed-sensitive (\cref{sec:multiseed}).

\subsection{Auditing Input Transformations under a Unified EC Space}
\label{exp:input_transformation}

While \hyperref[exp:backbone]{Experiment 1} focuses on how data is tokenized and processed, this experiment investigates the impact of preprocessing the raw signal via Input Transformations.
We designate the Input Transformation ($\mathcal{T}_{in}$) as the Evaluated Object (EO), evaluating three distinct Input Transformations—Trend-Seasonal Decomposition, Multi-Scale Downsampling, and Residual Cycle Forecasting—with RevIN as the shared normalization baseline.

\subsubsection{Experiment Setup}

To rigorously audit the utility of Input Transformations, we construct a specialized Evaluation Condition space $\Omega_{\text{priors}}$.

\paragraph{Evaluated Objects.}
We select four representative Input Transformations ($\mathcal{T}_{in}$) to cover the spectrum of structural priors emerging in recent literature.
To address distribution shifts and non-stationarity, we include \textbf{RevIN}~\cite{kim2021reversible-RevIN} as the standard normalization baseline.
For temporal decomposition, we examine \textbf{Trend-Seasonal Decomposition}~\cite{zeng2023transformers-dlinear}, which explicitly isolates low-frequency trends.
To capture dependencies across varying granularities, we incorporate \textbf{Multi-Scale Downsampling}~\cite{wang2024timemixer}, while \textbf{Residual Cycle Forecasting}~\cite{lin2024cyclenet} is selected to represent explicit periodicity modeling via learnable cycle embeddings.

\paragraph{Evaluation Conditions.}

To rigorously analyze synergy and efficiency, we tailor the Evaluation Condition space around three key dimensions.
First, we restrict the backbone embedding to \textbf{Point-wise} and \textbf{Variate-wise} paradigms, contrasting step-level fragility with channel-level stability.
Second, to test the hypothesis that Input Transformations reduce the need for model depth, we expand the capacity search space to include ultra-lightweight configurations, sampling hidden dimensions $D \in \{16, \dots, 512\}$.
Finally, we broaden the learning rate spectrum to $\eta \in \{10^{-2}, 10^{-3}, 10^{-4}\}$, specifically assessing whether input simplifications enable stable convergence even under aggressive optimization steps.

\begin{table}[t]
    \centering
    \scriptsize
    \setlength{\tabcolsep}{1.5pt}
    \renewcommand{\arraystretch}{1.1}
    \caption{
        Performance of 6 Datasets across 4 Horizons on 4 input transformations.
        Highlighting follows the convention in \cref{tab:exp1_emb_sub_data}.
        \textsuperscript{$\dagger$}TimeMixer results are $L_{best}$ reported in~\cite{wang2024timemixer} under their own protocol and are not directly comparable to our paired EC estimates.
    }
    \label{tab:exp2_data}
    \resizebox{\columnwidth}{!}{
        \begin{tabular}{l | ccc | ccc | ccc | ccc | c}

    \toprule

    \multirow{2}{*}{\textbf{Dataset}} & \multicolumn{3}{c|}{\textbf{BaseLine}} & \multicolumn{3}{c|}{\textbf{Cycle}} & \multicolumn{3}{c|}{\textbf{MultiScale}} & \multicolumn{3}{c|}{\textbf{TrendSeasonal}} & \textbf{TimeMixer\textsuperscript{$\dagger$}} \\

\cmidrule(lr){2-4} \cmidrule(lr){5-7} \cmidrule(lr){8-10} \cmidrule(lr){11-13} \cmidrule(lr){14-14}

    & $\hat{\mu}$ & $\hat{\sigma}$ & $min$ & $\hat{\mu}$ & $\hat{\sigma}$ & $min$ & $\hat{\mu}$ & $\hat{\sigma}$ & $min$ & $\hat{\mu}$ & $\hat{\sigma}$ & $min$ & $min$ \\

    \midrule

    \textbf{ETTh1} & 0.3975 & 0.0315 & \textcolor{darkred}{\textbf{\underline{0.3749}}} & \textcolor{darkblue}{\textbf{0.3891}} & 0.0283 & \textcolor{darkred}{\textbf{0.3743}} & \textcolor{darkblue}{\textbf{\underline{0.3921}}} & 0.0183 & 0.3829 & 0.3947 & 0.0248 & 0.3811 & \textcolor{darkgreen}{\textbf{0.3610}} \\

    \textbf{ETTh2} & \textcolor{darkblue}{\textbf{\underline{0.3013}}} & 0.0116 & 0.2868 & \textcolor{darkblue}{\textbf{0.2969}} & 0.0115 & \textcolor{darkred}{\textbf{0.2818}} & 0.3071 & 0.0343 & 0.2874 & 0.3024 & 0.0135 & \textcolor{darkred}{\textbf{\underline{0.2849}}} & \textcolor{darkgreen}{\textbf{0.2710}} \\

    \textbf{ETTm1} & 0.3451 & 0.0206 & \textcolor{darkred}{\textbf{\underline{0.3139}}} & 0.3431 & 0.0193 & \textcolor{darkred}{\textbf{0.3137}} & \textcolor{darkblue}{\textbf{0.3380}} & 0.0171 & 0.3159 & \textcolor{darkblue}{\textbf{\underline{0.3384}}} & 0.0167 & 0.3166 & \textcolor{darkgreen}{\textbf{0.2910}} \\

    \textbf{ETTm2} & 0.1827 & 0.0037 & \textcolor{darkred}{\textbf{0.1738}} & \textcolor{darkblue}{\textbf{\underline{0.1825}}} & 0.0019 & \textcolor{darkred}{\textbf{\underline{0.1804}}} & 0.1830 & 0.0011 & 0.1818 & \textcolor{darkblue}{\textbf{0.1821}} & 0.0013 & 0.1805 & \textcolor{darkgreen}{\textbf{0.1640}} \\

    \textbf{Electricity} & 0.2061 & 0.0238 & 0.1679 & 0.1999 & 0.0225 & \textcolor{darkred}{\textbf{0.1580}} & \textcolor{darkblue}{\textbf{0.1952}} & 0.0202 & \textcolor{darkred}{\textbf{\underline{0.1580}}} & \textcolor{darkblue}{\textbf{\underline{0.1976}}} & 0.0197 & 0.1729 & \textcolor{darkgreen}{\textbf{0.1290}} \\

    \textbf{Weather} & \textcolor{darkblue}{\textbf{0.1882}} & 0.0094 & \textcolor{darkred}{\textbf{0.1599}} & \textcolor{darkblue}{\textbf{\underline{0.1917}}} & 0.0017 & \textcolor{darkred}{\textbf{\underline{0.1899}}} & 0.1950 & 0.0021 & 0.1915 & 0.1934 & 0.0008 & 0.1920 & \textcolor{darkgreen}{\textbf{0.1470}} \\

    \bottomrule

\end{tabular}

    }
\end{table}

\subsubsection{Results and Analysis}

We analyze the marginalized performance ($\mu$) and stability ($\sigma$) of various Input Transformations ($\mathcal{T}_{in}$) across the evaluated datasets. The comprehensive results are summarized in Table \ref{tab:exp2_data}; detailed per-embedding and per-encoder breakdowns are provided in \cref{tab:exp2_emb_data,tab:exp2_enc_data} in the Appendix. Our analysis yields three critical insights into the role of Input Transformations:

\begin{itemize}[wide,itemsep=-0.1em]
    \item[\ding{182}] \textbf{Universal Efficacy of Cyclic Priors.}
The \textbf{Cycle} transformation yields the most consistent gains in both effectiveness ($\mu$) and stability ($\sigma$) across datasets.
This confirms that explicit periodic inductive biases are far more efficient than implicit learning, significantly simplifying the optimization landscape even for deep backbones.

    \item[\ding{183}] \textbf{The Fallacy of Naive Decomposition.}
In contrast, naive \textbf{Trend-Seasonal} and \textbf{Multi-Scale} strategies frequently fail to outperform the \textbf{Identity} baseline.
A ``divide and conquer'' approach without subsequent interaction leads to information loss, as isolated branches cannot capture the entangled dynamics of the signal.

    \item[\ding{184}] \textbf{Interaction Defines Utility.}
The superior performance of \textbf{TimeMixer} over naive Multi-Scale—despite identical inputs—isolates \textit{interaction} as the decisive factor.
Input Transformations (like decomposition) are not standalone solutions; they only yield benefits when coupled with architectural mechanisms that enforce deep cross-component mixing.

\end{itemize}

\subsection{Revisiting Spectral Processing: Performance vs. Necessity}
\label{exp:spectral}

While Experiment 1 identified the dominance of data view (Embedding) and Experiment 2 validated the utility of Input Transformations, a critical question remains regarding the \textbf{Encoder ($\varPhi$)}: Does processing latent representations in the \textbf{frequency domain} offer superior effectiveness compared to the \textbf{time domain}, and does it also improve stability?
Recent works like \textbf{SimpleTM} \cite{chen2025simpletm} argue that spectral decomposition can effectively disentangle noise from signal.
To rigorously verify this, we designate the spectral encoder (from SimpleTM) as the Evaluated Object (EO), alongside the Transformer and Identity encoders from \hyperref[exp:backbone]{Experiment 1}.

\subsubsection{Experiment Setup}

\paragraph{Evaluated Objects.}
We adopt a strictly controlled pair to isolate the spectral mechanism.
The \textbf{Time-Domain} baseline uses \textbf{iTransformer} (Variate-wise Embedding + Self-Attention encoder), while the \textbf{Frequency-Domain} treatment uses \textbf{SimpleTM}~\cite{chen2025simpletm} with the same Variate-wise Embedding but replaces attention with learnable spectral decomposition (Wavelets/Fourier).

\paragraph{Evaluation Conditions.}
Both encoders are evaluated under the identical EC space from \hyperref[exp:backbone]{Experiment 1} ($K=600$ stratified samples across all 6 datasets), ensuring a controlled comparison.

\begin{table}[t]
    \centering
    \scriptsize
    \setlength{\tabcolsep}{1.5pt}
    \renewcommand{\arraystretch}{1.1}
    \caption{
        Paired EC evaluation of Identity, time-domain (iTransformer), and frequency-domain (SimpleTM) encoders under Variate-wise Embedding across 6 datasets and 4 prediction horizons.
        Highlighting follows the convention in \cref{tab:exp1_emb_sub_data}.
        Detailed results are reported in \cref{tab:exp3_enc_data}.
    }
    \label{tab:exp3_sub_data}
    \resizebox{\columnwidth}{!}{
        \begin{tabular}{l c | ccc | ccc | ccc}
    \toprule
    \multirow{2}{*}{\textbf{Dataset}} & \multirow{2}{*}{\textbf{H}} & 
    \multicolumn{3}{c|}{\textbf{iTransformer}} & 
    \multicolumn{3}{c|}{\textbf{SimpleTM}} & 
    \multicolumn{3}{c}{\textbf{Variate and Identity}} \\
    \cmidrule(lr){3-5} \cmidrule(lr){6-8} \cmidrule(lr){9-11}
    & & $\hat{\mu}$ & $\hat{\sigma}$ & $min$ & $\hat{\mu}$ & $\hat{\sigma}$ & $min$ & $\hat{\mu}$ & $\hat{\sigma}$ & $min$ \\
    \midrule

    \multirow{5}{*}{\textbf{ETTh1}}
    & 96 & 0.3934 & 0.0133 & 0.3789 & \textcolor{darkblue}{\textbf{\underline{0.3812}}} & 0.0076 & \textcolor{darkred}{\textbf{\underline{0.3730}}} & \textcolor{darkblue}{\textbf{0.3805}} & 0.0098 & \textcolor{darkred}{\textbf{0.3641}} \\
    & 192 & 0.4354 & 0.0099 & 0.4171 & \textcolor{darkblue}{\textbf{\underline{0.4287}}} & 0.0105 & \textcolor{darkred}{\textbf{\underline{0.4148}}} & \textcolor{darkblue}{\textbf{0.4199}} & 0.0179 & \textcolor{darkred}{\textbf{0.3983}} \\
    & 336 & 0.4656 & 0.0127 & 0.4434 & \textcolor{darkblue}{\textbf{\underline{0.4601}}} & 0.0132 & \textcolor{darkred}{\textbf{\underline{0.4314}}} & \textcolor{darkblue}{\textbf{0.4583}} & 0.0204 & \textcolor{darkred}{\textbf{0.4249}} \\
    & 720 & 0.4711 & 0.0155 & 0.4504 & \textcolor{darkblue}{\textbf{\underline{0.4646}}} & 0.0151 & \textcolor{darkred}{\textbf{\underline{0.4473}}} & \textcolor{darkblue}{\textbf{0.4520}} & 0.0191 & \textcolor{darkred}{\textbf{0.4366}} \\
    & avg & 0.4401 & 0.0335 & - & \textcolor{darkblue}{\textbf{\underline{0.4324}}} & 0.0354 & - & \textcolor{darkblue}{\textbf{0.4271}} & 0.0355 & - \\
    \cmidrule{1-11}

    \multirow{5}{*}{\textbf{ETTm2}}
    & 96 & 0.1980 & 0.0291 & 0.1782 & \textcolor{darkblue}{\textbf{\underline{0.1738}}} & 0.0061 & \textcolor{darkred}{\textbf{\underline{0.1646}}} & \textcolor{darkblue}{\textbf{0.1731}} & 0.0079 & \textcolor{darkred}{\textbf{0.1622}} \\
    & 192 & 0.2478 & 0.0057 & 0.2399 & \textcolor{darkblue}{\textbf{\underline{0.2332}}} & 0.0078 & \textcolor{darkred}{\textbf{\underline{0.2229}}} & \textcolor{darkblue}{\textbf{0.2296}} & 0.0119 & \textcolor{darkred}{\textbf{0.2184}} \\
    & 336 & 0.3080 & 0.0140 & 0.2894 & \textcolor{darkblue}{\textbf{\underline{0.2930}}} & 0.0133 & \textcolor{darkred}{\textbf{\underline{0.2735}}} & \textcolor{darkblue}{\textbf{0.2866}} & 0.0150 & \textcolor{darkred}{\textbf{0.2707}} \\
    & 720 & 0.3981 & 0.0195 & 0.3696 & \textcolor{darkblue}{\textbf{\underline{0.3846}}} & 0.0170 & \textcolor{darkred}{\textbf{\underline{0.3639}}} & \textcolor{darkblue}{\textbf{0.3793}} & 0.0184 & \textcolor{darkred}{\textbf{0.3594}} \\
    & avg & 0.2870 & 0.0803 & - & \textcolor{darkblue}{\textbf{0.2697}} & 0.0824 & - & \textcolor{darkblue}{\textbf{\underline{0.2721}}} & 0.0798 & - \\
    \cmidrule{1-11}

    \multirow{5}{*}{\textbf{Electricity}}
    & 96 & \textcolor{darkblue}{\textbf{0.1517}} & 0.0184 & \textcolor{darkred}{\textbf{\underline{0.1335}}} & \textcolor{darkblue}{\textbf{\underline{0.1537}}} & 0.0196 & \textcolor{darkred}{\textbf{0.1322}} & 0.1702 & 0.0270 & 0.1428 \\
    & 192 & \textcolor{darkblue}{\textbf{\underline{0.1705}}} & 0.0145 & \textcolor{darkred}{\textbf{\underline{0.1555}}} & \textcolor{darkblue}{\textbf{0.1701}} & 0.0160 & \textcolor{darkred}{\textbf{0.1493}} & 0.1810 & 0.0184 & 0.1584 \\
    & 336 & \textcolor{darkblue}{\textbf{\underline{0.1930}}} & 0.0213 & \textcolor{darkred}{\textbf{\underline{0.1722}}} & \textcolor{darkblue}{\textbf{0.1929}} & 0.0233 & \textcolor{darkred}{\textbf{0.1648}} & 0.2015 & 0.0230 & 0.1756 \\
    & 720 & \textcolor{darkblue}{\textbf{\underline{0.2287}}} & 0.0255 & \textcolor{darkred}{\textbf{\underline{0.1972}}} & \textcolor{darkblue}{\textbf{0.2244}} & 0.0280 & \textcolor{darkred}{\textbf{0.1964}} & 0.2448 & 0.0236 & 0.2137 \\
    & avg & \textcolor{darkblue}{\textbf{\underline{0.1858}}} & 0.0333 & - & \textcolor{darkblue}{\textbf{0.1851}} & 0.0329 & - & 0.1974 & 0.0351 & - \\
    
    \bottomrule
\end{tabular}

    }
\end{table}

\subsubsection{Results and Analysis}
The comprehensive comparison across all 6 datasets and 4 prediction horizons is provided in \cref{tab:exp3_enc_data} in the Appendix.
\begin{itemize}[wide,itemsep=-0.1em]
    \item[\ding{182}] \textbf{Spectral Superiority in Effectiveness.}
    Under Variate-wise embedding, the \textbf{Frequency-Domain} encoder consistently outperforms Time-Domain attention in effectiveness ($\mu$), yet exhibits comparable stability ($\sigma$). This indicates that spectral decomposition improves signal representation rather than acting as a stabilizing filter against hyperparameter sensitivity.

    \item[\ding{183}] \textbf{The ``Identity'' Ceiling.}
    Crucially, spectral processing fails to surpass the parameter-free \textbf{Identity} baseline on datasets such as ETTh1 and ETTm2, implying that its effectiveness gain over attention stems merely from introducing \textit{less signal distortion} rather than extracting meaningful patterns beyond what the optimized data view already provides.
\end{itemize}

\section{Conclusion}
\label{sec:conclusion}

We introduced \textbf{CombinationTS} to pivot TSF research from model selection to modular attribution via a rigorous probabilistic audit. 
Our findings expose the \textbf{``Identity'' Paradox}: optimized data view often dictates performance ceilings, rendering deep encoders statistically indistinguishable from parameter-free Identity mappings within standard benchmarks. 
Consequently, we propose a new \textbf{Evaluation Principle}: future innovations must shift the burden of proof, demonstrating significant gains in \textbf{Marginalized Performance ($\hat{\mu}$)} and \textbf{Stability ($\hat{\sigma}$)} over the robust \textbf{Identity} baseline, rather than relying on fragile leaderboard peaks. 
CombinationTS thus serves as a necessary lens to distinguish genuine reasoning capability from the illusion of progress.

\section*{Impact Statement}

This paper presents CombinationTS, a modular evaluation framework for time series forecasting, with the goal of advancing methodological rigor and interpretability in machine learning research. The work focuses on improving how forecasting models are analyzed and compared, rather than introducing new predictive capabilities.

The proposed framework is intended for benign applications such as scientific analysis, industrial forecasting, and infrastructure-related time series modeling, where reliable evaluation and reproducibility are essential. By emphasizing modular attribution and distributional evaluation, this work may help reduce misattributed performance gains and encourage more robust and efficient model design.

We do not anticipate negative societal impacts beyond those commonly associated with machine learning research, such as potential misuse or over-reliance on automated predictions. These risks are not specific to this work and can be mitigated through responsible use and appropriate human oversight. We do not identify ethical concerns requiring special consideration beyond standard practices in the field.

\bibliography{reference}
\bibliographystyle{icml2026}

\newpage
\appendix
\onecolumn

\makeatletter
\setlength{\@fptop}{0pt}        
\setlength{\@fpsep}{8pt plus 1fil}
\setlength{\@fpbot}{0pt}        
\setlength{\@dblfptop}{0pt}     
\setlength{\@dblfpsep}{8pt plus 1fil}
\setlength{\@dblfpbot}{0pt}
\makeatother

\section{Implementation Details}

\subsection{Dataset Descriptions}\label{sec:dataset}

We evaluate our approach on \textbf{six} widely used real-world multivariate time series datasets. These datasets originate from diverse application domains, including electricity systems, meteorology, traffic dynamics, and energy consumption, which enables a comprehensive assessment of the robustness and general applicability of the proposed method. The overall statistics of the datasets are summarized in Table~\ref{tab:datasets}. Specifically, the datasets are described as follows:

\begin{itemize}
    \item \textbf{Electricity Transformer Temperature (ETT)}~\cite{zhou2021informer}: This benchmark consists of four subsets. ETTh1 and ETTh2 are sampled at an hourly frequency, while ETTm1 and ETTm2 are collected at 15-minute intervals. All time series are measured from two electricity transformers.
    
    \item \textbf{Weather}~\cite{wu2021autoformer}: This dataset contains 21 meteorological variables collected at 10-minute intervals from a weather station in Germany during 2020, released by the Max Planck Institute for Biogeochemistry.
    
    \item \textbf{Electricity}~\cite{wu2021autoformer}: This dataset records hourly electricity consumption from 321 individual clients over the period from 2012 to 2014.

\end{itemize}

\begin{table}[th]
    \caption{Dataset statistics. ``Var'' denotes the number of variates; ``Dataset Size'' shows the number of time points in (Train/Validation/Test) splits; ``Frequency'' is the sampling interval.}
  \label{tab:datasets}
  \centering
  \setlength\tabcolsep{3pt}
  \begin{tabular}{c|c|c|c}
    \hline
    Dataset & Var & Dataset Size (Train/Val/Test) & Frequency \\
    \hline
    ETTh1, ETTh2 & 7 & (8545, 2881, 2881) & Hourly \\
    \hline
    ETTm1, ETTm2 & 7 & (34465, 11521, 11521) & 15 min \\
    \hline
    Weather & 21 & (36792, 5271, 10540) & 10 min \\
    \hline
    Electricity & 321 & (18317, 2633, 5261) & Hourly \\
    \hline
  \end{tabular}
\end{table}

\subsection{Evaluation Metrics}
\label{sec:metrics}

We follow standard evaluation protocols in time series forecasting and adopt commonly used metrics to assess predictive performance across all datasets.

For the considered benchmarks, we adopt Mean Squared Error (MSE) as the primary metric, defined as:
\begin{itemize}
    \item Mean Squared Error (MSE): $\tfrac{1}{T}\sum_{t=1}^T (\hat{y}_t - y_t)^2$
\end{itemize}
This metric is widely adopted in prior studies~\cite{nie2022time-patchtst} and penalizes large deviations more strongly, providing a rigorous measure of forecasting accuracy. All metrics are averaged over all variates and prediction horizons, with lower values indicating better performance.

\subsection{Experiment details}
\label{appendix:exp_details}
All models were implemented in PyTorch and trained on a Linux server with NVIDIA V100 GPUs.

\paragraph{Hyperparameter Search Space.}
\label{sec:hparam_search}
During the hyperparameter optimization process, all methods share the same predefined search space to guarantee fairness.
Specifically, the candidate ranges are defined as follows:
\begin{itemize}
    \item \texttt{Look-back windows} $\in \{96, 192, 336, 512\}$, 
    \item \texttt{Prediction horizons} $\in \{96, 192, 336, 720\}$,
    \item  \texttt {batch size} $\in \{32\}$,
    \item  \texttt {epochs} $\in \{30\}$,
    \item  \texttt {early stopping patience} $\in \{3\}$,
    \item  \texttt {learning rate}  $\in \{1\text{e-}3, 1\text{e-}4\}$,
    \item  \texttt {encoder layers} $\in \{1, 2, 3\}$,
    \item  \texttt {hidden dimensions}$ \in \{64, 128, 256, 512\}$ 
    
\end{itemize}

All baselines are evaluated within a common search space that encompasses architectural design choices and optimization-related settings, and this protocol is applied consistently across methods.
Consequently, the reported performance better reflects the intrinsic representational capability of each model, instead of discrepancies induced by unequal hyperparameter tuning.

\begin{table*}[t]
    \centering
    \scriptsize
    \setlength{\tabcolsep}{1.5pt}
    \renewcommand{\arraystretch}{1.1}
    \caption{
        Experiment 1 (Dissecting the Backbone): Performance of 6 Datasets across 4 Horizons on 6 Embeddings.
        Highlighting follows the convention in \cref{tab:exp1_emb_sub_data}.
    }
    \label{tab:exp1_emb_data}
    \resizebox{\textwidth}{!}{
        \begin{tabular}{l c | ccc | ccc | ccc | ccc | ccc | ccc}
    \toprule
    \multirow{2}{*}{\textbf{Dataset}} & \multirow{2}{*}{\textbf{H}} & 
    \multicolumn{3}{c|}{\textbf{Point wise}} &
    \multicolumn{3}{c|}{\textbf{Patch wise}} &
    \multicolumn{3}{c|}{\textbf{Variate wise}} &
    \multicolumn{3}{c|}{\textbf{Identity}} &
    \multicolumn{3}{c|}{\textbf{L to dim}} &
    \multicolumn{3}{c}{\textbf{C to dim}} \\
    \cmidrule(lr){3-5} \cmidrule(lr){6-8} \cmidrule(lr){9-11} \cmidrule(lr){12-14} \cmidrule(lr){15-17} \cmidrule(lr){18-20}
    & & $\hat{\mu}$ & $\hat{\sigma}$ & $min$ & $\hat{\mu}$ & $\hat{\sigma}$ & $min$ & $\hat{\mu}$ & $\hat{\sigma}$ & $min$ & $\hat{\mu}$ & $\hat{\sigma}$ & $min$ & $\hat{\mu}$ & $\hat{\sigma}$ & $min$ & $\hat{\mu}$ & $\hat{\sigma}$ & $min$ \\
    \midrule

    \multirow{5}{*}{\textbf{ETTh1}}

    & 96 & 0.3898 & 0.0143 & 0.3651 & \textcolor{darkblue}{\textbf{0.3772}} & 0.0058 & \textcolor{darkred}{\textbf{\underline{0.3645}}} & \textcolor{darkblue}{\textbf{\underline{0.3834}}} & 0.0080 & \textcolor{darkred}{\textbf{0.3641}} & 0.5890 & 0.1537 & 0.3721 & 0.3937 & 0.0156 & 0.3721 & 0.4602 & 0.0357 & 0.4013 \\

    & 192 & 0.4324 & 0.0166 & 0.3992 & \textcolor{darkblue}{\textbf{0.4240}} & 0.0133 & \textcolor{darkred}{\textbf{\underline{0.3984}}} & \textcolor{darkblue}{\textbf{\underline{0.4247}}} & 0.0128 & \textcolor{darkred}{\textbf{0.3983}} & 0.6092 & 0.1408 & 0.4028 & 0.4343 & 0.0161 & 0.4028 & 0.5096 & 0.0342 & 0.4484 \\

    & 336 & 0.4651 & 0.0193 & 0.4255 & \textcolor{darkblue}{\textbf{0.4575}} & 0.0170 & \textcolor{darkred}{\textbf{0.4196}} & \textcolor{darkblue}{\textbf{\underline{0.4589}}} & 0.0148 & \textcolor{darkred}{\textbf{\underline{0.4249}}} & 0.6262 & 0.1209 & 0.4294 & 0.4693 & 0.0160 & 0.4294 & 0.5219 & 0.0175 & 0.4859 \\

    & 720 & 0.4772 & 0.0415 & \textcolor{darkred}{\textbf{0.4248}} & 0.4812 & 0.0326 & 0.4366 & \textcolor{darkblue}{\textbf{0.4618}} & 0.0140 & \textcolor{darkred}{\textbf{\underline{0.4366}}} & 0.6317 & 0.1305 & 0.4409 & \textcolor{darkblue}{\textbf{\underline{0.4751}}} & 0.0223 & 0.4409 & 0.5652 & 0.0474 & 0.4934 \\

    & avg & 0.4367 & 0.0364 & - & \textcolor{darkblue}{\textbf{0.4277}} & 0.0370 & - & \textcolor{darkblue}{\textbf{\underline{0.4285}}} & 0.0327 & - & 0.6133 & 0.1362 & - & 0.4398 & 0.0342 & - & 0.5104 & 0.0451 & - \\

    \cmidrule{1-20}

    \multirow{5}{*}{\textbf{ETTh2}}

    & 96 & 0.3076 & 0.0264 & \textcolor{darkred}{\textbf{\underline{0.2684}}} & \textcolor{darkblue}{\textbf{0.2912}} & 0.0108 & 0.2699 & \textcolor{darkblue}{\textbf{\underline{0.2963}}} & 0.0163 & \textcolor{darkred}{\textbf{0.2656}} & 0.3438 & 0.0444 & 0.2751 & 0.3034 & 0.0171 & 0.2751 & 0.3874 & 0.0245 & 0.3552 \\

    & 192 & 0.3799 & 0.0267 & \textcolor{darkred}{\textbf{0.3258}} & \textcolor{darkblue}{\textbf{0.3626}} & 0.0153 & 0.3265 & \textcolor{darkblue}{\textbf{\underline{0.3709}}} & 0.0189 & \textcolor{darkred}{\textbf{\underline{0.3264}}} & 0.3969 & 0.0293 & 0.3363 & 0.3752 & 0.0184 & 0.3363 & 0.4380 & 0.0151 & 0.3946 \\

    & 336 & 0.4123 & 0.0286 & \textcolor{darkred}{\textbf{\underline{0.3502}}} & \textcolor{darkblue}{\textbf{\underline{0.3949}}} & 0.0202 & \textcolor{darkred}{\textbf{0.3502}} & \textcolor{darkblue}{\textbf{0.3937}} & 0.0191 & 0.3509 & 0.4102 & 0.0278 & 0.3553 & 0.3998 & 0.0204 & 0.3553 & 0.4478 & 0.0201 & 0.4147 \\

    & 720 & 0.4365 & 0.0363 & \textcolor{darkred}{\textbf{0.3799}} & 0.4205 & 0.0178 & 0.3852 & \textcolor{darkblue}{\textbf{0.4137}} & 0.0140 & \textcolor{darkred}{\textbf{\underline{0.3841}}} & 0.4408 & 0.0352 & 0.3897 & \textcolor{darkblue}{\textbf{\underline{0.4203}}} & 0.0174 & 0.3897 & 0.4731 & 0.0168 & 0.4481 \\

    & avg & 0.3838 & 0.0513 & - & \textcolor{darkblue}{\textbf{0.3657}} & 0.0459 & - & \textcolor{darkblue}{\textbf{\underline{0.3689}}} & 0.0436 & - & 0.3956 & 0.0416 & - & 0.3751 & 0.0436 & - & 0.4376 & 0.0330 & - \\

    \cmidrule{1-20}

    \multirow{5}{*}{\textbf{ETTm1}}

    & 96 & 0.3134 & 0.0136 & \textcolor{darkred}{\textbf{\underline{0.2866}}} & \textcolor{darkblue}{\textbf{0.3054}} & 0.0141 & \textcolor{darkred}{\textbf{0.2837}} & \textcolor{darkblue}{\textbf{\underline{0.3117}}} & 0.0142 & 0.2878 & 0.5506 & 0.1744 & 0.3030 & 0.3212 & 0.0171 & 0.3012 & 0.3437 & 0.0222 & 0.3085 \\

    & 192 & 0.3558 & 0.0172 & 0.3298 & \textcolor{darkblue}{\textbf{0.3483}} & 0.0188 & \textcolor{darkred}{\textbf{0.3235}} & \textcolor{darkblue}{\textbf{\underline{0.3515}}} & 0.0186 & \textcolor{darkred}{\textbf{\underline{0.3280}}} & 0.5708 & 0.1613 & 0.3368 & 0.3596 & 0.0194 & 0.3368 & 0.3883 & 0.0198 & 0.3471 \\

    & 336 & 0.3889 & 0.0172 & 0.3650 & \textcolor{darkblue}{\textbf{0.3776}} & 0.0145 & \textcolor{darkred}{\textbf{0.3581}} & \textcolor{darkblue}{\textbf{\underline{0.3866}}} & 0.0210 & \textcolor{darkred}{\textbf{\underline{0.3615}}} & 0.5908 & 0.1522 & 0.3714 & 0.3923 & 0.0180 & 0.3714 & 0.4242 & 0.0213 & 0.3860 \\

    & 720 & \textcolor{darkblue}{\textbf{\underline{0.4391}}} & 0.0146 & \textcolor{darkred}{\textbf{\underline{0.4129}}} & \textcolor{darkblue}{\textbf{0.4309}} & 0.0126 & \textcolor{darkred}{\textbf{0.4128}} & 0.4393 & 0.0160 & 0.4179 & 0.6190 & 0.1357 & 0.4228 & 0.4469 & 0.0180 & 0.4228 & 0.4758 & 0.0225 & 0.4417 \\

    & avg & 0.3686 & 0.0458 & - & \textcolor{darkblue}{\textbf{0.3603}} & 0.0459 & - & \textcolor{darkblue}{\textbf{\underline{0.3676}}} & 0.0478 & - & 0.5821 & 0.1569 & - & 0.3740 & 0.0458 & - & 0.4026 & 0.0507 & - \\

    \cmidrule{1-20}

    \multirow{5}{*}{\textbf{ETTm2}}

    & 96 & 0.1800 & 0.0089 & \textcolor{darkred}{\textbf{\underline{0.1625}}} & \textcolor{darkblue}{\textbf{0.1728}} & 0.0051 & 0.1631 & \textcolor{darkblue}{\textbf{\underline{0.1758}}} & 0.0072 & \textcolor{darkred}{\textbf{0.1622}} & 0.2235 & 0.0416 & 0.1633 & 0.1779 & 0.0073 & 0.1633 & 0.1993 & 0.0146 & 0.1779 \\

    & 192 & 0.2414 & 0.0136 & \textcolor{darkred}{\textbf{0.2170}} & \textcolor{darkblue}{\textbf{0.2331}} & 0.0079 & \textcolor{darkred}{\textbf{\underline{0.2179}}} & \textcolor{darkblue}{\textbf{\underline{0.2343}}} & 0.0106 & 0.2184 & 0.2739 & 0.0345 & 0.2180 & 0.2370 & 0.0115 & 0.2180 & 0.2639 & 0.0135 & 0.2406 \\

    & 336 & 0.3019 & 0.0172 & \textcolor{darkred}{\textbf{0.2701}} & \textcolor{darkblue}{\textbf{0.2911}} & 0.0105 & 0.2732 & \textcolor{darkblue}{\textbf{\underline{0.2928}}} & 0.0135 & \textcolor{darkred}{\textbf{\underline{0.2707}}} & 0.3218 & 0.0265 & 0.2711 & 0.2968 & 0.0134 & 0.2711 & 0.3280 & 0.0121 & 0.3099 \\

    & 720 & 0.3981 & 0.0216 & \textcolor{darkred}{\textbf{0.3591}} & \textcolor{darkblue}{\textbf{0.3813}} & 0.0140 & 0.3608 & \textcolor{darkblue}{\textbf{\underline{0.3863}}} & 0.0170 & \textcolor{darkred}{\textbf{\underline{0.3594}}} & 0.4095 & 0.0236 & 0.3603 & 0.3892 & 0.0174 & 0.3603 & 0.4368 & 0.0145 & 0.4138 \\

    & avg & 0.2663 & 0.0762 & - & \textcolor{darkblue}{\textbf{0.2562}} & 0.0725 & - & \textcolor{darkblue}{\textbf{\underline{0.2596}}} & 0.0723 & - & 0.2975 & 0.0704 & - & 0.2612 & 0.0728 & - & 0.2937 & 0.0827 & - \\

    \cmidrule{1-20}

    \multirow{5}{*}{\textbf{Electricity}}

    & 96 & 0.1609 & 0.0228 & 0.1431 & \textcolor{darkblue}{\textbf{0.1512}} & 0.0176 & \textcolor{darkred}{\textbf{0.1315}} & \textcolor{darkblue}{\textbf{\underline{0.1527}}} & 0.0135 & \textcolor{darkred}{\textbf{\underline{0.1335}}} & 0.1639 & 0.0193 & 0.1440 & 0.1573 & 0.0190 & 0.1349 & 0.1877 & 0.0106 & 0.1658 \\

    & 192 & 0.1716 & 0.0154 & 0.1574 & \textcolor{darkblue}{\textbf{0.1641}} & 0.0111 & \textcolor{darkred}{\textbf{0.1465}} & 0.1710 & 0.0111 & 0.1555 & 0.1777 & 0.0169 & 0.1583 & \textcolor{darkblue}{\textbf{\underline{0.1710}}} & 0.0115 & \textcolor{darkred}{\textbf{\underline{0.1542}}} & - & - & - \\

    & 336 & 0.1917 & 0.0170 & 0.1735 & \textcolor{darkblue}{\textbf{0.1828}} & 0.0137 & \textcolor{darkred}{\textbf{0.1644}} & \textcolor{darkblue}{\textbf{\underline{0.1906}}} & 0.0153 & 0.1722 & 0.2270 & 0.1270 & 0.1748 & 0.1928 & 0.0170 & \textcolor{darkred}{\textbf{\underline{0.1706}}} & - & - & - \\

    & 720 & 0.2342 & 0.0194 & 0.2132 & \textcolor{darkblue}{\textbf{\underline{0.2236}}} & 0.0160 & 0.2004 & 0.2293 & 0.0205 & \textcolor{darkred}{\textbf{0.1972}} & 0.2335 & 0.0195 & 0.2144 & \textcolor{darkblue}{\textbf{0.2218}} & 0.0141 & \textcolor{darkred}{\textbf{\underline{0.1991}}} & - & - & - \\

    & avg & 0.1855 & 0.0252 & - & \textcolor{darkblue}{\textbf{0.1773}} & 0.0246 & - & \textcolor{darkblue}{\textbf{\underline{0.1827}}} & 0.0249 & - & 0.2053 & 0.0730 & - & 0.1851 & 0.0264 & - & 0.1877 & 0.0106 & - \\

    \cmidrule{1-20}

    \multirow{5}{*}{\textbf{Weather}}

    & 96 & 0.1715 & 0.0114 & 0.1543 & \textcolor{darkblue}{\textbf{0.1673}} & 0.0135 & \textcolor{darkred}{\textbf{\underline{0.1474}}} & \textcolor{darkblue}{\textbf{\underline{0.1679}}} & 0.0126 & 0.1479 & 0.2120 & 0.0279 & 0.1718 & 0.1705 & 0.0106 & \textcolor{darkred}{\textbf{0.1468}} & 0.1741 & 0.0132 & 0.1587 \\

    & 192 & 0.2182 & 0.0120 & 0.1969 & \textcolor{darkblue}{\textbf{0.2142}} & 0.0127 & \textcolor{darkred}{\textbf{0.1912}} & \textcolor{darkblue}{\textbf{\underline{0.2144}}} & 0.0129 & \textcolor{darkred}{\textbf{\underline{0.1914}}} & 0.2519 & 0.0205 & 0.2162 & 0.2147 & 0.0096 & 0.1961 & 0.2260 & 0.0150 & 0.2090 \\

    & 336 & 0.2639 & 0.0107 & 0.2473 & \textcolor{darkblue}{\textbf{\underline{0.2596}}} & 0.0121 & \textcolor{darkred}{\textbf{\underline{0.2422}}} & \textcolor{darkblue}{\textbf{0.2593}} & 0.0109 & \textcolor{darkred}{\textbf{0.2415}} & 0.2950 & 0.0215 & 0.2615 & 0.2610 & 0.0105 & 0.2454 & 0.2817 & 0.0137 & 0.2644 \\

    & 720 & 0.3391 & 0.0118 & 0.3180 & \textcolor{darkblue}{\textbf{\underline{0.3357}}} & 0.0134 & \textcolor{darkred}{\textbf{\underline{0.3158}}} & \textcolor{darkblue}{\textbf{0.3345}} & 0.0135 & \textcolor{darkred}{\textbf{0.3131}} & 0.3594 & 0.0168 & 0.3267 & 0.3386 & 0.0124 & 0.3197 & 0.3652 & 0.0194 & 0.3421 \\

    & avg & 0.2299 & 0.0514 & - & \textcolor{darkblue}{\textbf{0.2271}} & 0.0526 & - & \textcolor{darkblue}{\textbf{\underline{0.2274}}} & 0.0521 & - & 0.2655 & 0.0502 & - & 0.2287 & 0.0520 & - & 0.2436 & 0.0610 & - \\

    \bottomrule

    \end{tabular}

    }
\end{table*}

\begin{table*}[t]
    \centering
    \scriptsize
    \setlength{\tabcolsep}{1.5pt}
    \renewcommand{\arraystretch}{1.1}
    \caption{
        Experiment 1 (Dissecting the Backbone): Performance of 6 Datasets across 4 Horizons on 3 Encoders.
        Highlighting follows the convention in \cref{tab:exp1_emb_sub_data}.
    }
    \label{tab:exp1_enc_data}
    \resizebox{0.8\textwidth}{!}{
        \begin{tabular}{l c | ccc | ccc | ccc}
    \toprule
    \multirow{2}{*}{\textbf{Dataset}} & \multirow{2}{*}{\textbf{H}} &
    \multicolumn{3}{c|}{\textbf{MLP}} &
    \multicolumn{3}{c|}{\textbf{Transformer}} &
    \multicolumn{3}{c}{\textbf{Identity}} \\
    \cmidrule(lr){3-5} \cmidrule(lr){6-8} \cmidrule(lr){9-11}
    & & $\hat{\mu}$ & $\hat{\sigma}$ & $min$ & $\hat{\mu}$& $\hat{\sigma}$ & $min$ & $\hat{\mu}$ & $\hat{\sigma}$ & $min$ \\
    \midrule

    \multirow{5}{*}{\textbf{ETTh1}}
    & 96 & \textcolor{darkblue}{\textbf{\underline{0.4371}}} & 0.0816 & \textcolor{darkred}{\textbf{\underline{0.3700}}} & 0.4518 & 0.0930 & 0.3721 & \textcolor{darkblue}{\textbf{0.3907}} & 0.0231 & \textcolor{darkred}{\textbf{0.3641}} \\
    & 192 & \textcolor{darkblue}{\textbf{\underline{0.4870}}} & 0.0827 & \textcolor{darkred}{\textbf{\underline{0.4043}}} & 0.5044 & 0.0988 & 0.4091 & \textcolor{darkblue}{\textbf{0.4334}} & 0.0306 & \textcolor{darkred}{\textbf{0.3983}} \\
    & 336 & \textcolor{darkblue}{\textbf{\underline{0.5186}}} & 0.0787 & \textcolor{darkred}{\textbf{\underline{0.4335}}} & 0.5239 & 0.0795 & 0.4406 & \textcolor{darkblue}{\textbf{0.4684}} & 0.0268 & \textcolor{darkred}{\textbf{0.4196}} \\
    & 720 & \textcolor{darkblue}{\textbf{\underline{0.5538}}} & 0.1039 & 0.4533 & 0.5679 & 0.1014 & \textcolor{darkred}{\textbf{\underline{0.4504}}} & \textcolor{darkblue}{\textbf{0.4722}} & 0.0406 & \textcolor{darkred}{\textbf{0.4248}} \\
    & avg & \textcolor{darkblue}{\textbf{\underline{0.4960}}} & 0.0938 & - & 0.5095 & 0.1004 & - & \textcolor{darkblue}{\textbf{0.4392}} & 0.0424 & - \\
    \cmidrule{1-11}

    \multirow{5}{*}{\textbf{ETTh2}}
    & 96 & \textcolor{darkblue}{\textbf{\underline{0.3373}}} & 0.0423 & \textcolor{darkred}{\textbf{\underline{0.2770}}} & 0.3498 & 0.0437 & 0.2835 & \textcolor{darkblue}{\textbf{0.3014}} & 0.0368 & \textcolor{darkred}{\textbf{0.2656}} \\
    & 192 & \textcolor{darkblue}{\textbf{\underline{0.3985}}} & 0.0293 & \textcolor{darkred}{\textbf{\underline{0.3468}}} & 0.4060 & 0.0310 & 0.3485 & \textcolor{darkblue}{\textbf{0.3753}} & 0.0373 & \textcolor{darkred}{\textbf{0.3258}} \\
    & 336 & \textcolor{darkblue}{\textbf{\underline{0.4197}}} & 0.0288 & \textcolor{darkred}{\textbf{\underline{0.3680}}} & 0.4317 & 0.0308 & 0.3788 & \textcolor{darkblue}{\textbf{0.3999}} & 0.0325 & \textcolor{darkred}{\textbf{0.3502}} \\
    & 720 & \textcolor{darkblue}{\textbf{\underline{0.4460}}} & 0.0339 & \textcolor{darkred}{\textbf{\underline{0.4034}}} & 0.4616 & 0.0378 & 0.4065 & \textcolor{darkblue}{\textbf{0.4231}} & 0.0374 & \textcolor{darkred}{\textbf{0.3799}} \\
    & avg & \textcolor{darkblue}{\textbf{\underline{0.4009}}} & 0.0473 & - & 0.4135 & 0.0512 & - & \textcolor{darkblue}{\textbf{0.3778}} & 0.0547 & - \\
    \cmidrule{1-11}

    \multirow{5}{*}{\textbf{ETTm1}}
    & 96 & \textcolor{darkblue}{\textbf{\underline{0.3414}}} & 0.0785 & \textcolor{darkred}{\textbf{0.2837}} & 0.3704 & 0.0956 & \textcolor{darkred}{\textbf{\underline{0.2866}}} & \textcolor{darkblue}{\textbf{0.3192}} & 0.0185 & 0.3026 \\
    & 192 & \textcolor{darkblue}{\textbf{\underline{0.3863}}} & 0.0695 & \textcolor{darkred}{\textbf{0.3235}} & 0.4103 & 0.0890 & \textcolor{darkred}{\textbf{\underline{0.3298}}} & \textcolor{darkblue}{\textbf{0.3593}} & 0.0221 & 0.3354 \\
    & 336 & \textcolor{darkblue}{\textbf{\underline{0.4235}}} & 0.0734 & \textcolor{darkred}{\textbf{0.3581}} & 0.4435 & 0.0876 & \textcolor{darkred}{\textbf{\underline{0.3650}}} & \textcolor{darkblue}{\textbf{0.3919}} & 0.0199 & 0.3658 \\
    & 720 & \textcolor{darkblue}{\textbf{\underline{0.4810}}} & 0.0697 & \textcolor{darkred}{\textbf{0.4128}} & 0.4930 & 0.0830 & \textcolor{darkred}{\textbf{\underline{0.4129}}} & \textcolor{darkblue}{\textbf{0.4408}} & 0.0158 & 0.4185 \\
    & avg & \textcolor{darkblue}{\textbf{\underline{0.4079}}} & 0.0912 & - & 0.4282 & 0.0997 & - & \textcolor{darkblue}{\textbf{0.3721}} & 0.0457 & - \\
    \cmidrule{1-11}

    \multirow{5}{*}{\textbf{ETTm2}}
    & 96 & \textcolor{darkblue}{\textbf{\underline{0.1893}}} & 0.0197 & 0.1667 & 0.2027 & 0.0265 & \textcolor{darkred}{\textbf{\underline{0.1652}}} & \textcolor{darkblue}{\textbf{0.1755}} & 0.0087 & \textcolor{darkred}{\textbf{0.1622}} \\
    & 192 & \textcolor{darkblue}{\textbf{\underline{0.2525}}} & 0.0213 & \textcolor{darkred}{\textbf{\underline{0.2201}}} & 0.2610 & 0.0221 & 0.2313 & \textcolor{darkblue}{\textbf{0.2341}} & 0.0133 & \textcolor{darkred}{\textbf{0.2170}} \\
    & 336 & \textcolor{darkblue}{\textbf{\underline{0.3101}}} & 0.0198 & \textcolor{darkred}{\textbf{\underline{0.2738}}} & 0.3202 & 0.0199 & 0.2794 & \textcolor{darkblue}{\textbf{0.2956}} & 0.0194 & \textcolor{darkred}{\textbf{0.2701}} \\
    & 720 & \textcolor{darkblue}{\textbf{\underline{0.4057}}} & 0.0232 & \textcolor{darkred}{\textbf{\underline{0.3634}}} & 0.4163 & 0.0265 & 0.3653 & \textcolor{darkblue}{\textbf{0.3919}} & 0.0270 & \textcolor{darkred}{\textbf{0.3591}} \\
    & avg & \textcolor{darkblue}{\textbf{\underline{0.2784}}} & 0.0759 & - & 0.2909 & 0.0772 & - & \textcolor{darkblue}{\textbf{0.2621}} & 0.0756 & - \\
    \cmidrule{1-11}

    \multirow{5}{*}{\textbf{Electricity}}
    & 96 & \textcolor{darkblue}{\textbf{\underline{0.1599}}} & 0.0171 & \textcolor{darkred}{\textbf{\underline{0.1338}}} & \textcolor{darkblue}{\textbf{0.1535}} & 0.0194 & \textcolor{darkred}{\textbf{0.1315}} & 0.1679 & 0.0254 & 0.1428 \\
    & 192 & \textcolor{darkblue}{\textbf{\underline{0.1729}}} & 0.0130 & \textcolor{darkred}{\textbf{\underline{0.1515}}} & \textcolor{darkblue}{\textbf{0.1671}} & 0.0129 & \textcolor{darkred}{\textbf{0.1465}} & 0.1742 & 0.0161 & 0.1574 \\
    & 336 & \textcolor{darkblue}{\textbf{\underline{0.1925}}} & 0.0158 & \textcolor{darkred}{\textbf{\underline{0.1681}}} & \textcolor{darkblue}{\textbf{0.1858}} & 0.0153 & \textcolor{darkred}{\textbf{0.1644}} & 0.1939 & 0.0179 & 0.1735 \\
    & 720 & \textcolor{darkblue}{\textbf{\underline{0.2282}}} & 0.0171 & \textcolor{darkred}{\textbf{\underline{0.2070}}} & \textcolor{darkblue}{\textbf{0.2196}} & 0.0171 & \textcolor{darkred}{\textbf{0.1972}} & 0.2326 & 0.0191 & 0.2132 \\
    & avg & \textcolor{darkblue}{\textbf{\underline{0.1851}}} & 0.0253 & - & \textcolor{darkblue}{\textbf{0.1776}} & 0.0242 & - & 0.1888 & 0.0275 & - \\
    \cmidrule{1-11}

    \multirow{5}{*}{\textbf{Weather}}
    & 96 & \textcolor{darkblue}{\textbf{0.1714}} & 0.0180 & \textcolor{darkred}{\textbf{0.1468}} & \textcolor{darkblue}{\textbf{\underline{0.1763}}} & 0.0198 & \textcolor{darkred}{\textbf{\underline{0.1474}}} & 0.1767 & 0.0101 & 0.1587 \\
    & 192 & \textcolor{darkblue}{\textbf{0.2191}} & 0.0192 & \textcolor{darkred}{\textbf{\underline{0.1914}}} & 0.2289 & 0.0235 & \textcolor{darkred}{\textbf{0.1912}} & \textcolor{darkblue}{\textbf{\underline{0.2238}}} & 0.0087 & 0.2090 \\
    & 336 & \textcolor{darkblue}{\textbf{\underline{0.2717}}} & 0.0224 & \textcolor{darkred}{\textbf{0.2415}} & 0.2727 & 0.0237 & \textcolor{darkred}{\textbf{\underline{0.2422}}} & \textcolor{darkblue}{\textbf{0.2714}} & 0.0084 & 0.2605 \\
    & 720 & \textcolor{darkblue}{\textbf{\underline{0.3461}}} & 0.0192 & \textcolor{darkred}{\textbf{0.3131}} & 0.3502 & 0.0209 & \textcolor{darkred}{\textbf{\underline{0.3179}}} & \textcolor{darkblue}{\textbf{0.3436}} & 0.0129 & 0.3257 \\
    & avg & \textcolor{darkblue}{\textbf{\underline{0.2360}}} & 0.0569 & - & 0.2424 & 0.0588 & - & \textcolor{darkblue}{\textbf{0.2348}} & 0.0514 & - \\
    \bottomrule
\end{tabular}

    }
\end{table*}

\begin{table*}[t]
    \centering
    \scriptsize
    \setlength{\tabcolsep}{1.5pt}
    \renewcommand{\arraystretch}{1.1}
    \caption{
        Experiment 2 (Auditing Input Transformations): Ablation study on embedding strategies: Performance comparison between channel-wise and point-wise embeddings across 6 datasets.
        Highlighting follows the convention in \cref{tab:exp1_emb_sub_data}.
    }
    \label{tab:exp2_emb_data}
    \resizebox{0.75\textwidth}{!}{
        \begin{tabular}{l | ccc | ccc}
    \toprule
    \multirow{2}{*}{\textbf{Dataset}} &
    \multicolumn{3}{c|}{\textbf{Channel}} &
    \multicolumn{3}{c}{\textbf{Point}} \\
    \cmidrule(lr){2-4} \cmidrule(lr){5-7}
    & $\hat{\mu}$ & $\hat{\sigma}$ & $min$ & $\hat{\mu}$ & $\hat{\sigma}$ & $min$ \\
    \midrule

    \textbf{ETTh1} & 0.4028 & 0.0369 & \textcolor{darkred}{\textbf{0.3743}} & \textcolor{darkblue}{\textbf{0.3866}} & 0.0090 & 0.3747 \\
    \textbf{ETTh2} & \textcolor{darkblue}{\textbf{0.3010}} & 0.0139 & \textcolor{darkred}{\textbf{0.2818}} & 0.3028 & 0.0253 & 0.2829 \\
    \textbf{ETTm1} & 0.3462 & 0.0186 & 0.3160 & \textcolor{darkblue}{\textbf{0.3362}} & 0.0173 & \textcolor{darkred}{\textbf{0.3137}} \\
    \textbf{ETTm2} & 0.1830 & 0.0035 & \textcolor{darkred}{\textbf{0.1738}} & \textcolor{darkblue}{\textbf{0.1822}} & 0.0018 & 0.1778 \\
    \textbf{Electricity} & 0.2018 & 0.0291 & \textcolor{darkred}{\textbf{0.1580}} & \textcolor{darkblue}{\textbf{0.1976}} & 0.0102 & 0.1840 \\
    \textbf{Weather} & \textcolor{darkblue}{\textbf{0.1903}} & 0.0096 & \textcolor{darkred}{\textbf{0.1599}} & 0.1913 & 0.0038 & 0.1816 \\
    \bottomrule
\end{tabular}

    }
\end{table*}

\begin{table*}[t]
    \centering
    \scriptsize
    \setlength{\tabcolsep}{1.5pt}
    \renewcommand{\arraystretch}{1.1}
    \caption{
        Experiment 2 (Auditing Input Transformations): Ablation study on encoder architectures: Comparing Identity, MLP, and Transformer-based encoders across six datasets.
        Highlighting follows the convention in \cref{tab:exp1_emb_sub_data}.
    }
    \label{tab:exp2_enc_data}
    \resizebox{\textwidth}{!}{
        \begin{tabular}{l | ccc | ccc | ccc}
    \toprule
    \multirow{2}{*}{\textbf{Dataset}} & \multicolumn{3}{c|}{\textbf{Identity}} & \multicolumn{3}{c|}{\textbf{MLP}} & \multicolumn{3}{c}{\textbf{Transformer}} \\
\cmidrule(lr){2-4} \cmidrule(lr){5-7} \cmidrule(lr){8-10}
    & $\hat{\mu}$ & $\hat{\sigma}$ & $min$ & $\hat{\mu}$ & $\hat{\sigma}$ & $min$ & $\hat{\mu}$ & $\hat{\sigma}$ & $min$ \\
    \midrule

    \textbf{ETTh1} & \textcolor{darkblue}{\textbf{0.3932}} & 0.0253 & \textcolor{darkred}{\textbf{0.3743}} & \textcolor{darkblue}{\textbf{\underline{0.3971}}} & 0.0334 & \textcolor{darkred}{\textbf{\underline{0.3749}}} & 0.3984 & 0.0330 & 0.3773 \\
    \textbf{ETTh2} & \textcolor{darkblue}{\textbf{0.2908}} & 0.0091 & \textcolor{darkred}{\textbf{\underline{0.2829}}} & \textcolor{darkblue}{\textbf{\underline{0.3052}}} & 0.0123 & \textcolor{darkred}{\textbf{0.2818}} & 0.3097 & 0.0287 & 0.2873 \\
    \textbf{ETTm1} & 0.3613 & 0.0072 & 0.3522 & \textcolor{darkblue}{\textbf{\underline{0.3314}}} & 0.0123 & \textcolor{darkred}{\textbf{\underline{0.3159}}} & \textcolor{darkblue}{\textbf{0.3308}} & 0.0152 & \textcolor{darkred}{\textbf{0.3137}} \\
    \textbf{ETTm2} & \textcolor{darkblue}{\textbf{\underline{0.1827}}} & 0.0017 & 0.1804 & \textcolor{darkblue}{\textbf{0.1805}} & 0.0042 & \textcolor{darkred}{\textbf{0.1738}} & 0.1843 & 0.0034 & \textcolor{darkred}{\textbf{\underline{0.1770}}} \\
    \textbf{Electricity} & 0.2078 & 0.0152 & 0.1981 & \textcolor{darkblue}{\textbf{0.1930}} & 0.0217 & \textcolor{darkred}{\textbf{\underline{0.1588}}} & \textcolor{darkblue}{\textbf{\underline{0.1983}}} & 0.0251 & \textcolor{darkred}{\textbf{0.1580}} \\
    \textbf{Weather} & 0.1935 & 0.0019 & 0.1899 & \textcolor{darkblue}{\textbf{0.1807}} & 0.0114 & \textcolor{darkred}{\textbf{0.1599}} & \textcolor{darkblue}{\textbf{\underline{0.1901}}} & 0.0069 & \textcolor{darkred}{\textbf{\underline{0.1785}}} \\
    \bottomrule
\end{tabular}

    }
\end{table*}

\begin{figure}
    \centering
    \includegraphics[width=\textwidth]{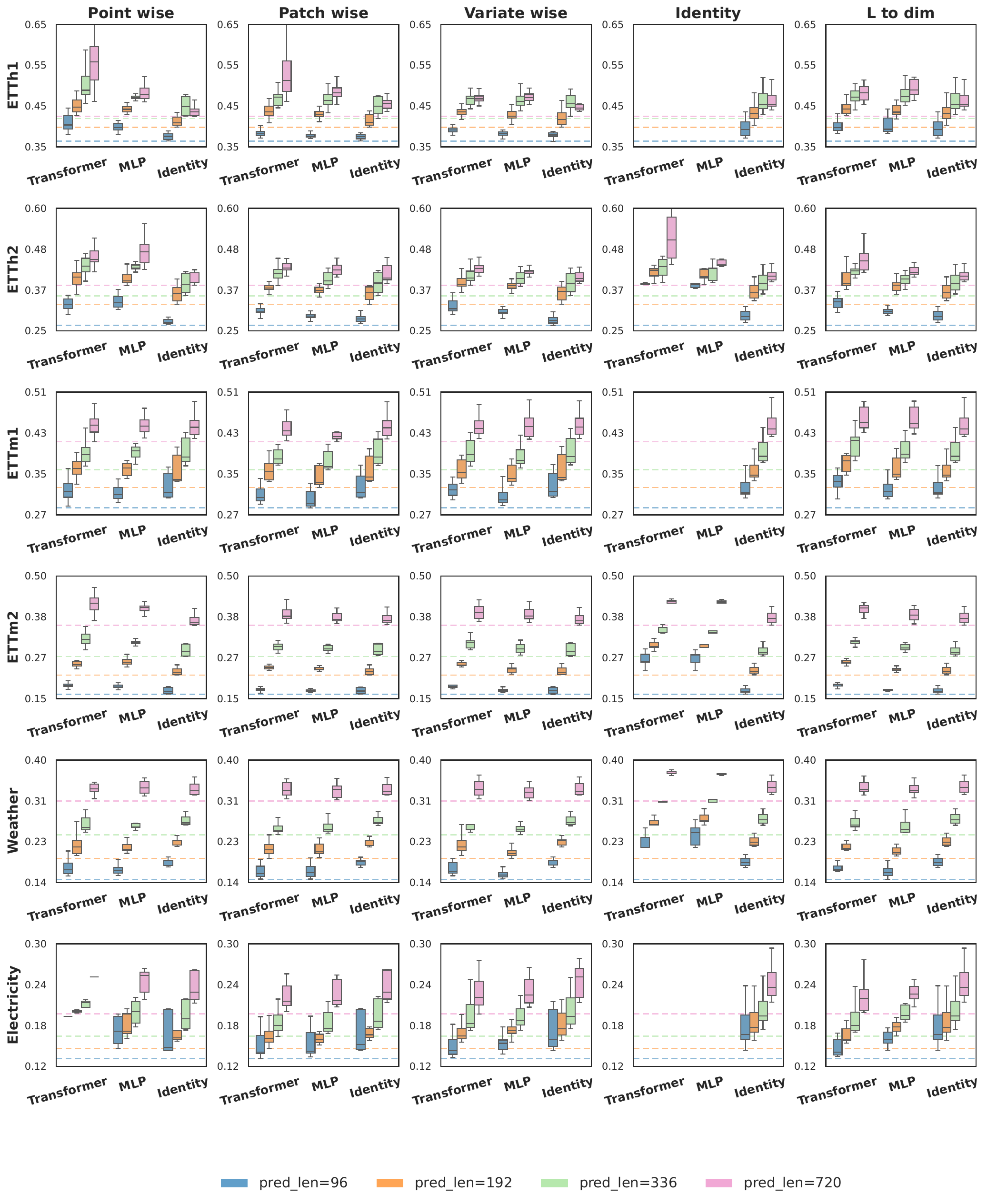}
    \caption{Distribution of encoder effectiveness under paired EC sampling (full figure).}
    \label{fig:exp1_enc_effect}
\end{figure}

\begin{table*}[t]
    \centering
    \scriptsize
    \setlength{\tabcolsep}{1.5pt}
    \renewcommand{\arraystretch}{1.1}
    \caption{
        Experiment 3 (Revisiting Spectral Processing): Comprehensive performance evaluation across various prediction horizons: A comparison between iTransformer and SimpleTM.
        Highlighting follows the convention in \cref{tab:exp1_emb_sub_data}.
    }
    \label{tab:exp3_enc_data}
    \resizebox{0.65\textwidth}{!}{
        \begin{tabular}{l c | ccc | ccc}
    \toprule
    \multirow{2}{*}{\textbf{Dataset}} & \multirow{2}{*}{\textbf{H}} & 
    \multicolumn{3}{c|}{\textbf{iTransformer}} & 
    \multicolumn{3}{c}{\textbf{SimpleTM}} \\
    \cmidrule(lr){3-5} \cmidrule(lr){6-8}
    & & $\hat{\mu}$ & $\hat{\sigma}$ & $min$ & $\hat{\mu}$ & $\hat{\sigma}$ & $min$ \\
    \midrule

    \multirow{5}{*}{\textbf{ETTh1}}
    & 96 & 0.3934 & 0.0133 & 0.3789 & \textcolor{darkblue}{\textbf{0.3812}} & 0.0076 & \textcolor{darkred}{\textbf{0.3730}} \\
    & 192 & 0.4354 & 0.0099 & 0.4171 & \textcolor{darkblue}{\textbf{0.4287}} & 0.0105 & \textcolor{darkred}{\textbf{0.4148}} \\
    & 336 & 0.4656 & 0.0127 & 0.4434 & \textcolor{darkblue}{\textbf{0.4601}} & 0.0132 & \textcolor{darkred}{\textbf{0.4314}} \\
    & 720 & 0.4711 & 0.0155 & 0.4504 & \textcolor{darkblue}{\textbf{0.4646}} & 0.0151 & \textcolor{darkred}{\textbf{0.4473}} \\
    & avg & 0.4401 & 0.0335 & - & \textcolor{darkblue}{\textbf{0.4324}} & 0.0354 & - \\
    \cmidrule{1-8}
    \multirow{5}{*}{\textbf{ETTh2}}
    & 96 & 0.3214 & 0.0228 & 0.2964 & \textcolor{darkblue}{\textbf{0.2945}} & 0.0065 & \textcolor{darkred}{\textbf{0.2840}} \\
    & 192 & 0.3887 & 0.0167 & 0.3599 & \textcolor{darkblue}{\textbf{0.3709}} & 0.0110 & \textcolor{darkred}{\textbf{0.3520}} \\
    & 336 & 0.4076 & 0.0209 & 0.3809 & \textcolor{darkblue}{\textbf{0.4030}} & 0.0204 & \textcolor{darkred}{\textbf{0.3727}} \\
    & 720 & 0.4309 & 0.0227 & 0.4065 & \textcolor{darkblue}{\textbf{0.4148}} & 0.0120 & \textcolor{darkred}{\textbf{0.3921}} \\
    & avg & 0.3894 & 0.0434 & - & \textcolor{darkblue}{\textbf{0.3736}} & 0.0463 & - \\
    \cmidrule{1-8}
    \multirow{5}{*}{\textbf{ETTm1}}
    & 96 & 0.3507 & 0.1022 & 0.2987 & \textcolor{darkblue}{\textbf{0.2974}} & 0.0164 & \textcolor{darkred}{\textbf{0.2825}} \\
    & 192 & 0.3670 & 0.0519 & 0.3319 & \textcolor{darkblue}{\textbf{0.3454}} & 0.0186 & \textcolor{darkred}{\textbf{0.3232}} \\
    & 336 & 0.4044 & 0.0608 & 0.3655 & \textcolor{darkblue}{\textbf{0.3792}} & 0.0179 & \textcolor{darkred}{\textbf{0.3590}} \\
    & 720 & 0.4447 & 0.0194 & 0.4194 & \textcolor{darkblue}{\textbf{0.4334}} & 0.0171 & \textcolor{darkred}{\textbf{0.4183}} \\
    & avg & 0.3911 & 0.0764 & - & \textcolor{darkblue}{\textbf{0.3624}} & 0.0551 & - \\
    \cmidrule{1-8}
    \multirow{5}{*}{\textbf{ETTm2}}
    & 96 & 0.1980 & 0.0291 & 0.1782 & \textcolor{darkblue}{\textbf{0.1738}} & 0.0061 & \textcolor{darkred}{\textbf{0.1646}} \\
    & 192 & 0.2478 & 0.0057 & 0.2399 & \textcolor{darkblue}{\textbf{0.2332}} & 0.0078 & \textcolor{darkred}{\textbf{0.2229}} \\
    & 336 & 0.3080 & 0.0140 & 0.2894 & \textcolor{darkblue}{\textbf{0.2930}} & 0.0133 & \textcolor{darkred}{\textbf{0.2735}} \\
    & 720 & 0.3981 & 0.0195 & 0.3696 & \textcolor{darkblue}{\textbf{0.3846}} & 0.0170 & \textcolor{darkred}{\textbf{0.3639}} \\
    & avg & 0.2870 & 0.0803 & - & \textcolor{darkblue}{\textbf{0.2697}} & 0.0824 & - \\
    \cmidrule{1-8}
    \multirow{5}{*}{\textbf{Electricity}}
    & 96 & \textcolor{darkblue}{\textbf{0.1517}} & 0.0184 & 0.1335 & 0.1537 & 0.0196 & \textcolor{darkred}{\textbf{0.1322}} \\
    & 192 & 0.1705 & 0.0145 & 0.1555 & \textcolor{darkblue}{\textbf{0.1701}} & 0.0160 & \textcolor{darkred}{\textbf{0.1493}} \\
    & 336 & 0.1930 & 0.0213 & 0.1722 & \textcolor{darkblue}{\textbf{0.1929}} & 0.0233 & \textcolor{darkred}{\textbf{0.1648}} \\
    & 720 & 0.2287 & 0.0255 & 0.1972 & \textcolor{darkblue}{\textbf{0.2244}} & 0.0280 & \textcolor{darkred}{\textbf{0.1964}} \\
    & avg & 0.1858 & 0.0333 & - & \textcolor{darkblue}{\textbf{0.1851}} & 0.0329 & - \\
    \cmidrule{1-8}
    \multirow{5}{*}{\textbf{Weather}}
    & 96 & 0.1738 & 0.0231 & 0.1544 & \textcolor{darkblue}{\textbf{0.1564}} & 0.0118 & \textcolor{darkred}{\textbf{0.1440}} \\
    & 192 & 0.2230 & 0.0226 & 0.1976 & \textcolor{darkblue}{\textbf{0.2030}} & 0.0121 & \textcolor{darkred}{\textbf{0.1873}} \\
    & 336 & 0.2610 & 0.0156 & 0.2465 & \textcolor{darkblue}{\textbf{0.2524}} & 0.0133 & \textcolor{darkred}{\textbf{0.2376}} \\
    & 720 & 0.3384 & 0.0157 & 0.3179 & \textcolor{darkblue}{\textbf{0.3305}} & 0.0145 & \textcolor{darkred}{\textbf{0.3137}} \\
    & avg & 0.2412 & 0.0621 & - & \textcolor{darkblue}{\textbf{0.2274}} & 0.0644 & - \\
    \bottomrule
\end{tabular}

    }
\end{table*}

\begin{table*}[t]
    \centering
    \scriptsize
    \setlength{\tabcolsep}{1.5pt}
    \renewcommand{\arraystretch}{1.1}
    \caption{
        Experiment 2 (Auditing Input Transformations): Performance of 6 Datasets across 4 Horizons on 4 Input Transformations.
        Highlighting follows the convention in \cref{tab:exp1_emb_sub_data}, where red denotes MSE and blue denotes MAE.
    }
    \label{tab:proposed_baseline_new}
    \resizebox{\textwidth}{!}{
        \begin{tabular}{l c | cc | cc | cc | cc | cc | cc | cc | cc}
    \toprule
    \multicolumn{2}{c|}{\multirow{3}{*}{\textbf{Methods}}} & 
    \multicolumn{8}{c|}{\textbf{Comparison Baseline}} & 
    \multicolumn{8}{c}{\textbf{Compared Model}} \\
    
    \cmidrule(lr){3-10} \cmidrule(lr){11-18}
    
     & & 
    \multicolumn{2}{c|}{\textbf{PatchLinear}} & 
    \multicolumn{2}{c|}{\textbf{iLinear}} & 
    \multicolumn{2}{c|}{\textbf{PointLinear}} & 
    \multicolumn{2}{c|}{\textbf{RLinear}} &
    \multicolumn{2}{c|}{\textbf{TimeMixer}} &
    \multicolumn{2}{c|}{\textbf{SimpleTM}} &
    \multicolumn{2}{c|}{\textbf{PatchTST}} &
    \multicolumn{2}{c}{\textbf{iTransformer}} \\
    
     & & 
    \multicolumn{2}{c|}{(\textcolor[RGB]{170,88,0}{Ours})} &
    \multicolumn{2}{c|}{(\textcolor[RGB]{170,88,0}{Ours})} &
    \multicolumn{2}{c|}{(\textcolor[RGB]{170,88,0}{Ours})} &
    \multicolumn{2}{c|}{(\citeyear{kim2021reversible-RevIN})} &
    \multicolumn{2}{c|}{(\citeyear{wang2024timemixer})} &
    \multicolumn{2}{c|}{(\citeyear{chen2025simpletm})} &
    \multicolumn{2}{c|}{(\citeyear{nie2022time-patchtst})} &
    \multicolumn{2}{c}{(\citeyear{liu2023itransformer})} \\

    \cmidrule(lr){1-2} \cmidrule(lr){3-4} \cmidrule(lr){5-6} \cmidrule(lr){7-8} \cmidrule(lr){9-10} \cmidrule(lr){11-12} \cmidrule(lr){13-14} \cmidrule(lr){15-16} \cmidrule(lr){17-18}
    
    \multicolumn{2}{c|}{\textbf{Metric}} & MSE & MAE & MSE & MAE & MSE & MAE & MSE & MAE & MSE & MAE & MSE & MAE & MSE & MAE & MSE & MAE \\
    
    \midrule
    
    \multirow{5}{*}{\textbf{ETTh1}} 
    & 96 & 0.365 & 0.392 & \secr{0.364} & 0.392 & 0.365 & 0.392 & 0.366 & \secb{0.391} & \bestr{0.361} & \bestb{0.390} & 0.366 & 0.392 & 0.370 & 0.400 & 0.386 & 0.405 \\
    & 192 & \bestr{0.398} & 0.414 & \bestr{0.398} & 0.414 & \secr{0.399} & \secb{0.413} & 0.404 & \bestb{0.412} & 0.409 & 0.414 & 0.422 & 0.421 & 0.413 & 0.429 & 0.441 & 0.436 \\
    & 336 & \bestr{0.420} & \secb{0.429} & 0.425 & 0.435 & 0.426 & 0.432 & \bestr{0.420} & \bestb{0.423} & 0.430 & \secb{0.429} & 0.440 & 0.438 & \secr{0.422} & 0.440 & 0.487 & 0.458 \\
    & 720 & \secr{0.437} & 0.459 & \secr{0.437} & 0.460 & \bestr{0.425} & \bestb{0.451} & 0.442 & \secb{0.456} & 0.445 & 0.460 & 0.463 & 0.462 & 0.447 & 0.468 & 0.503 & 0.491 \\
    \cmidrule(lr){2-18}
    & avg & \secr{0.405} & 0.424 & 0.406 & 0.425 & \bestr{0.404} & \secb{0.422} & 0.408 & \bestb{0.421} & 0.411 & 0.423 & 0.423 & 0.428 & 0.413 & 0.434 & 0.454 & 0.448 \\
    \midrule

    \multirow{5}{*}{\textbf{ETTh2}} 
    & 96 & 0.270 & 0.337 & \secr{0.266} & 0.334 & 0.268 & 0.335 & \bestr{0.262} & \secb{0.331} & 0.271 & \bestb{0.330} & 0.281 & 0.338 & 0.274 & 0.337 & 0.297 & 0.349 \\
    & 192 & 0.327 & 0.376 & 0.326 & \secb{0.373} & 0.326 & \bestb{0.372} & \secr{0.319} & 0.374 & \bestr{0.317} & 0.402 & 0.355 & 0.387 & 0.339 & 0.379 & 0.380 & 0.400 \\
    & 336 & 0.350 & 0.396 & 0.351 & 0.396 & 0.350 & 0.395 & \bestr{0.325} & \secb{0.386} & 0.332 & 0.396 & 0.365 & 0.401 & \secr{0.329} & \bestb{0.384} & 0.428 & 0.432 \\
    & 720 & 0.385 & 0.429 & 0.384 & 0.427 & 0.380 & 0.425 & \secr{0.372} & \secb{0.421} & \bestr{0.342} & \bestb{0.408} & 0.413 & 0.436 & 0.379 & 0.422 & 0.427 & 0.445 \\
    \cmidrule(lr){2-18}
    & avg & 0.333 & 0.385 & 0.332 & 0.383 & 0.331 & 0.382 & \secr{0.320} & \bestb{0.378} & \bestr{0.316} & 0.384 & 0.354 & 0.391 & 0.330 & \secb{0.381} & 0.383 & 0.407 \\
    \midrule

    \multirow{5}{*}{\textbf{ETTm1}} 
    & 96 & 0.303 & 0.347 & 0.304 & 0.348 & 0.303 & 0.347 & 0.301 & \secb{0.342} & \secr{0.291} & \bestb{0.340} & 0.321 & 0.361 & \bestr{0.290} & \secb{0.342} & 0.334 & 0.368 \\
    & 192 & 0.335 & 0.367 & 0.337 & 0.369 & 0.335 & 0.367 & 0.335 & \bestb{0.363} & \bestr{0.327} & \secb{0.365} & 0.360 & 0.380 & \secr{0.332} & 0.369 & 0.377 & 0.391 \\
    & 336 & 0.367 & 0.386 & 0.368 & 0.387 & \secr{0.366} & 0.385 & 0.370 & \secb{0.383} & \bestr{0.360} & \bestb{0.381} & 0.390 & 0.404 & \secr{0.366} & 0.392 & 0.426 & 0.420 \\
    & 720 & 0.419 & \secb{0.415} & 0.419 & 0.417 & 0.419 & 0.416 & 0.425 & \bestb{0.414} & \bestr{0.415} & 0.417 & 0.454 & 0.438 & \secr{0.416} & 0.420 & 0.491 & 0.459 \\
    \cmidrule(lr){2-18}
    & avg & 0.356 & \secb{0.379} & 0.357 & 0.380 & 0.356 & \secb{0.379} & 0.358 & \bestb{0.376} & \bestr{0.348} & \bestb{0.376} & 0.381 & 0.396 & \secr{0.351} & 0.381 & 0.407 & 0.410 \\
    \midrule

    \multirow{5}{*}{\textbf{ETTm2}} 
    & 96 & \secr{0.163} & \secb{0.254} & \bestr{0.162} & \bestb{0.253} & \secr{0.163} & \bestb{0.253} & 0.164 & \bestb{0.253} & 0.164 & \secb{0.254} & 0.173 & 0.257 & 0.165 & 0.255 & 0.180 & 0.264 \\
    & 192 & \secr{0.218} & \secb{0.291} & \secr{0.218} & \secb{0.291} & \bestr{0.217} & \bestb{0.290} & 0.219 & \bestb{0.290} & 0.223 & 0.295 & 0.238 & 0.299 & 0.220 & 0.292 & 0.250 & 0.309 \\
    & 336 & 0.273 & 0.329 & \secr{0.271} & \secb{0.326} & \bestr{0.270} & \bestb{0.325} & 0.273 & \secb{0.326} & 0.279 & 0.330 & 0.296 & 0.338 & 0.274 & 0.329 & 0.311 & 0.348 \\
    & 720 & \secr{0.361} & \secb{0.383} & \bestr{0.359} & \bestb{0.381} & \bestr{0.359} & \bestb{0.381} & 0.366 & 0.385 & \bestr{0.359} & \secb{0.383} & 0.393 & 0.395 & 0.362 & 0.385 & 0.412 & 0.407 \\
    \cmidrule(lr){2-18}
    & avg & 0.254 & 0.314 & \secr{0.253} & \secb{0.313} & \bestr{0.252} & \bestb{0.312} & 0.256 & 0.314 & 0.256 & 0.316 & 0.275 & 0.322 & 0.255 & 0.315 & 0.288 & 0.332 \\
    \midrule

    \multirow{5}{*}{\textbf{Electricity}} 
    & 96 & 0.144 & 0.248 & 0.143 & 0.246 & 0.143 & 0.248 & \secr{0.140} & 0.235 & \bestr{0.129} & \secb{0.224} & 0.141 & 0.235 & \bestr{0.129} & \bestb{0.222} & 0.148 & 0.240 \\
    & 192 & 0.158 & 0.261 & 0.158 & 0.261 & 0.157 & 0.260 & 0.154 & 0.248 & \bestr{0.140} & \bestb{0.220} & 0.151 & 0.247 & \secr{0.147} & \secb{0.240} & 0.162 & 0.253 \\
    & 336 & 0.175 & 0.276 & 0.176 & 0.277 & 0.174 & 0.275 & 0.171 & 0.264 & \bestr{0.161} & \bestb{0.255} & 0.173 & 0.267 & \secr{0.163} & \secb{0.259} & 0.178 & 0.269 \\
    & 720 & 0.214 & 0.308 & 0.214 & 0.308 & 0.213 & 0.307 & 0.209 & 0.297 & \bestr{0.194} & \bestb{0.287} & 0.201 & 0.293 & \secr{0.197} & \secb{0.290} & 0.225 & 0.317 \\
    \cmidrule(lr){2-18}
    & avg & 0.172 & 0.273 & 0.173 & 0.273 & 0.172 & 0.272 & 0.169 & 0.261 & \bestr{0.156} & \bestb{0.247} & 0.167 & 0.261 & \secr{0.159} & \secb{0.253} & 0.178 & 0.270 \\
    \midrule

    \multirow{5}{*}{\textbf{Weather}} 
    & 96 & 0.172 & 0.226 & 0.172 & 0.227 & 0.173 & 0.227 & 0.175 & 0.225 & \bestr{0.147} & \bestb{0.197} & 0.162 & 0.207 & \secr{0.149} & \secb{0.198} & 0.174 & 0.214 \\
    & 192 & 0.216 & 0.262 & 0.216 & 0.262 & 0.216 & 0.263 & 0.218 & 0.260 & \bestr{0.189} & \bestb{0.239} & 0.208 & 0.248 & \secr{0.194} & \secb{0.241} & 0.221 & 0.254 \\
    & 336 & 0.261 & 0.295 & 0.261 & 0.295 & 0.262 & 0.296 & 0.265 & 0.294 & \bestr{0.241} & \bestb{0.280} & 0.263 & 0.290 & \secr{0.245} & \secb{0.282} & 0.278 & 0.296 \\
    & 720 & 0.326 & 0.341 & 0.326 & 0.342 & 0.326 & 0.342 & 0.329 & 0.339 & \bestr{0.310} & \bestb{0.330} & 0.340 & 0.341 & \secr{0.314} & \secb{0.334} & 0.358 & 0.347 \\
    \cmidrule(lr){2-18}
    & avg & 0.244 & 0.281 & 0.244 & 0.281 & 0.244 & 0.282 & 0.247 & 0.280 & \bestr{0.222} & \bestb{0.262} & 0.243 & 0.272 & \secr{0.226} & \secb{0.264} & 0.258 & 0.278 \\

    \bottomrule
\end{tabular}
    }
\end{table*}

\begin{figure}
    \centering
    \includegraphics[width=\textwidth]{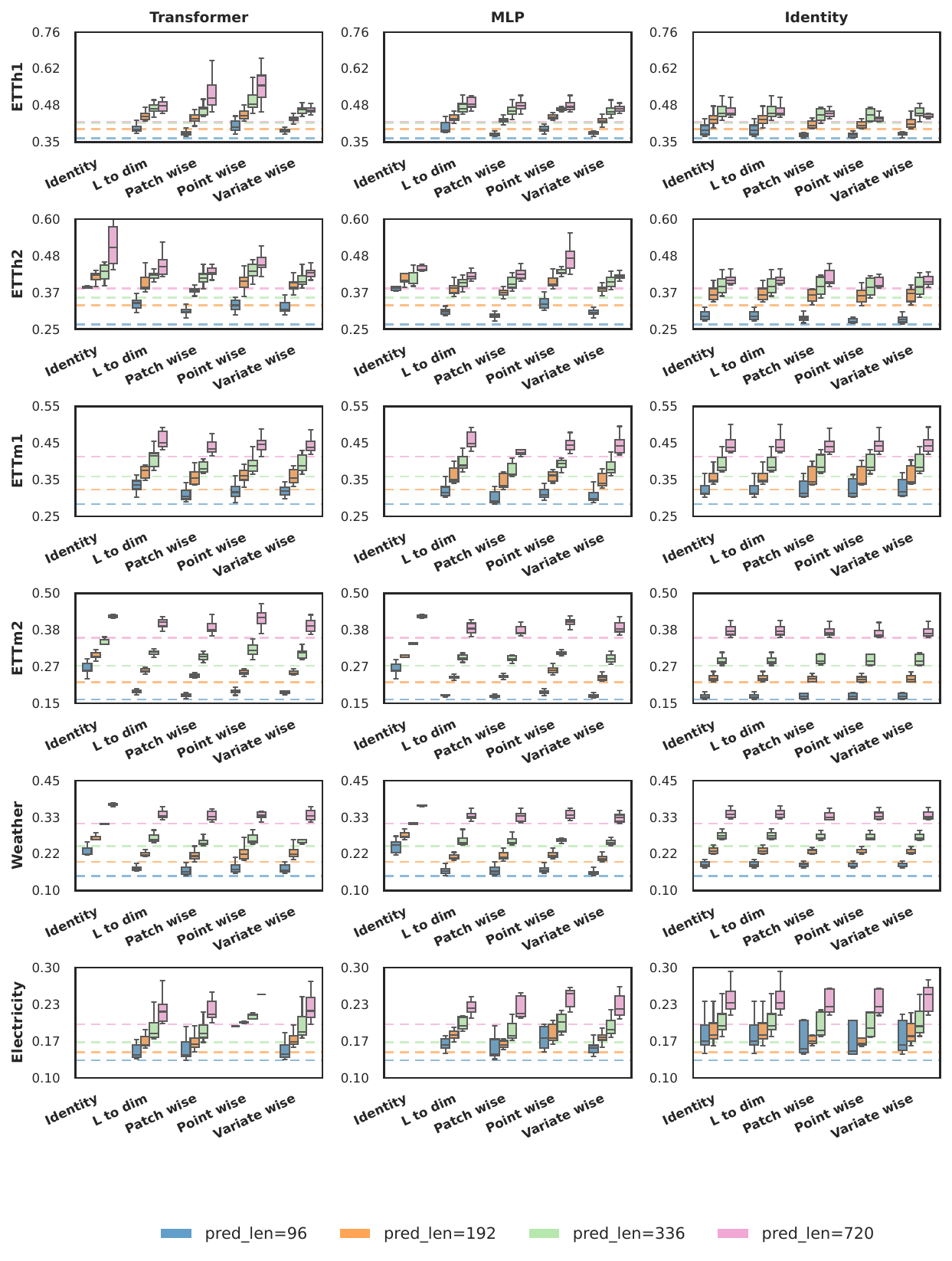}
    \caption{Distribution of embedding effectiveness under paired EC sampling (full figure).}
    \label{fig:exp1_emb_effect}

\end{figure}

\begin{table*}[t]
    \centering
    \scriptsize
    \setlength{\tabcolsep}{1.5pt}
    \renewcommand{\arraystretch}{1.1}
    \caption{Experiment 1 (Dissecting the Backbone): Marginalized performance and stability with 95\% Confidence Intervals.}
    \label{tab:overall_results}

    \begin{subtable}{\textwidth}
        \centering
        \caption{Experiment 1 (Dissecting the Backbone): Performance evaluation and statistical reliability analysis: Comparison of five configurations with 95\% Confidence Intervals}
        \resizebox{\textwidth}{!}{
            \begin{tabular}{l c | ccc | ccc | ccc | ccc | ccc}
    \toprule
    \multirow{2}{*}{\textbf{Dataset}} & \multirow{2}{*}{\textbf{H}} & \multicolumn{3}{c|}{\textbf{Point wise}} & \multicolumn{3}{c|}{\textbf{Patch wise}} & \multicolumn{3}{c|}{\textbf{Variate wise}} & \multicolumn{3}{c|}{\textbf{Identity}} & \multicolumn{3}{c}{\textbf{L to dim}} \\
    \cmidrule(lr){3-5} \cmidrule(lr){6-8} \cmidrule(lr){9-11} \cmidrule(lr){12-14} \cmidrule(lr){15-17}
    & & $\hat{\mu}$ & $ci_{low}$ & $ci_{upp}$ & $\hat{\mu}$ & $ci_{low}$ & $ci_{upp}$ & $\hat{\mu}$ & $ci_{low}$ & $ci_{upp}$ & $\hat{\mu}$ & $ci_{low}$ & $ci_{upp}$ & $\hat{\mu}$ & $ci_{low}$ & $ci_{upp}$ \\
    \midrule
    \multirow{5}{*}{\textbf{ETTh1}}
    & 96 & 0.3898 & 0.3865 & 0.3932 & \textcolor{darkblue}{\textbf{0.3772}} & 0.3758 & 0.3786 & \textcolor{darkblue}{\textbf{\underline{0.3834}}} & 0.3816 & 0.3853 & 0.5890 & 0.5530 & 0.6250 & 0.3937 & 0.3900 & 0.3974 \\
    & 192 & 0.4324 & 0.4286 & 0.4363 & \textcolor{darkblue}{\textbf{0.4240}} & 0.4208 & 0.4271 & \textcolor{darkblue}{\textbf{\underline{0.4247}}} & 0.4217 & 0.4277 & 0.6092 & 0.5760 & 0.6424 & 0.4343 & 0.4305 & 0.4381 \\
    & 336 & 0.4651 & 0.4606 & 0.4696 & \textcolor{darkblue}{\textbf{0.4575}} & 0.4536 & 0.4614 & \textcolor{darkblue}{\textbf{\underline{0.4589}}} & 0.4555 & 0.4624 & 0.6262 & 0.5983 & 0.6541 & 0.4693 & 0.4656 & 0.4730 \\
    & 720 & 0.4772 & 0.4663 & 0.4881 & 0.4812 & 0.4727 & 0.4898 & \textcolor{darkblue}{\textbf{0.4618}} & 0.4581 & 0.4654 & 0.6317 & 0.5975 & 0.6659 & \textcolor{darkblue}{\textbf{\underline{0.4751}}} & 0.4692 & 0.4810 \\
    & avg & 0.4394 & 0.4345 & 0.4444 & \textcolor{darkblue}{\textbf{\underline{0.4327}}} & 0.4276 & 0.4379 & \textcolor{darkblue}{\textbf{0.4308}} & 0.4267 & 0.4349 & 0.6132 & 0.5967 & 0.6297 & 0.4417 & 0.4373 & 0.4461 \\
    \cmidrule{1-17}
    \multirow{5}{*}{\textbf{ETTh2}}
    & 96 & 0.3076 & 0.3007 & 0.3145 & \textcolor{darkblue}{\textbf{0.2912}} & 0.2884 & 0.2940 & \textcolor{darkblue}{\textbf{\underline{0.2963}}} & 0.2921 & 0.3006 & 0.3438 & 0.3321 & 0.3556 & 0.3034 & 0.2989 & 0.3080 \\
    & 192 & 0.3799 & 0.3740 & 0.3858 & \textcolor{darkblue}{\textbf{0.3626}} & 0.3592 & 0.3660 & \textcolor{darkblue}{\textbf{\underline{0.3709}}} & 0.3667 & 0.3751 & 0.3969 & 0.3903 & 0.4034 & 0.3752 & 0.3711 & 0.3794 \\
    & 336 & 0.4123 & 0.4057 & 0.4189 & \textcolor{darkblue}{\textbf{\underline{0.3949}}} & 0.3903 & 0.3996 & \textcolor{darkblue}{\textbf{0.3937}} & 0.3892 & 0.3981 & 0.4102 & 0.4038 & 0.4166 & 0.3998 & 0.3951 & 0.4046 \\
    & 720 & 0.4365 & 0.4275 & 0.4456 & 0.4205 & 0.4161 & 0.4249 & \textcolor{darkblue}{\textbf{0.4137}} & 0.4102 & 0.4172 & 0.4408 & 0.4320 & 0.4495 & \textcolor{darkblue}{\textbf{\underline{0.4203}}} & 0.4160 & 0.4247 \\
    & avg & 0.3866 & 0.3801 & 0.3931 & \textcolor{darkblue}{\textbf{0.3698}} & 0.3640 & 0.3756 & \textcolor{darkblue}{\textbf{\underline{0.3713}}} & 0.3659 & 0.3767 & 0.3997 & 0.3941 & 0.4054 & 0.3774 & 0.3719 & 0.3829 \\
    \cmidrule{1-17}
    \multirow{5}{*}{\textbf{ETTm1}}
    & 96 & 0.3134 & 0.3103 & 0.3164 & \textcolor{darkblue}{\textbf{0.3054}} & 0.3023 & 0.3085 & \textcolor{darkblue}{\textbf{\underline{0.3117}}} & 0.3085 & 0.3149 & 0.5506 & 0.5119 & 0.5893 & 0.3212 & 0.3174 & 0.3250 \\
    & 192 & 0.3558 & 0.3515 & 0.3602 & \textcolor{darkblue}{\textbf{0.3483}} & 0.3436 & 0.3530 & \textcolor{darkblue}{\textbf{\underline{0.3515}}} & 0.3468 & 0.3561 & 0.5708 & 0.5303 & 0.6113 & 0.3596 & 0.3547 & 0.3644 \\
    & 336 & 0.3889 & 0.3844 & 0.3935 & \textcolor{darkblue}{\textbf{0.3776}} & 0.3738 & 0.3814 & \textcolor{darkblue}{\textbf{\underline{0.3866}}} & 0.3811 & 0.3921 & 0.5908 & 0.5509 & 0.6307 & 0.3923 & 0.3875 & 0.3972 \\
    & 720 & \textcolor{darkblue}{\textbf{\underline{0.4391}}} & 0.4357 & 0.4424 & \textcolor{darkblue}{\textbf{0.4309}} & 0.4280 & 0.4338 & 0.4393 & 0.4356 & 0.4430 & 0.6190 & 0.5876 & 0.6503 & 0.4469 & 0.4428 & 0.4511 \\
    & avg & 0.3730 & 0.3669 & 0.3791 & \textcolor{darkblue}{\textbf{0.3641}} & 0.3581 & 0.3702 & \textcolor{darkblue}{\textbf{\underline{0.3711}}} & 0.3648 & 0.3773 & 0.5821 & 0.5631 & 0.6011 & 0.3786 & 0.3724 & 0.3849 \\
    \cmidrule{1-17}
    \multirow{5}{*}{\textbf{ETTm2}}
    & 96 & 0.1800 & 0.1781 & 0.1820 & \textcolor{darkblue}{\textbf{0.1728}} & 0.1717 & 0.1740 & \textcolor{darkblue}{\textbf{\underline{0.1758}}} & 0.1742 & 0.1773 & 0.2235 & 0.2145 & 0.2326 & 0.1779 & 0.1763 & 0.1795 \\
    & 192 & 0.2414 & 0.2377 & 0.2452 & \textcolor{darkblue}{\textbf{0.2331}} & 0.2309 & 0.2352 & \textcolor{darkblue}{\textbf{\underline{0.2343}}} & 0.2314 & 0.2372 & 0.2739 & 0.2645 & 0.2834 & 0.2370 & 0.2337 & 0.2402 \\
    & 336 & 0.3019 & 0.2977 & 0.3060 & \textcolor{darkblue}{\textbf{0.2911}} & 0.2886 & 0.2936 & \textcolor{darkblue}{\textbf{\underline{0.2928}}} & 0.2896 & 0.2961 & 0.3218 & 0.3154 & 0.3281 & 0.2968 & 0.2936 & 0.3000 \\
    & 720 & 0.3981 & 0.3930 & 0.4032 & \textcolor{darkblue}{\textbf{0.3813}} & 0.3780 & 0.3846 & \textcolor{darkblue}{\textbf{\underline{0.3863}}} & 0.3823 & 0.3903 & 0.4095 & 0.4040 & 0.4151 & 0.3892 & 0.3851 & 0.3933 \\
    & avg & 0.2782 & 0.2680 & 0.2885 & \textcolor{darkblue}{\textbf{0.2680}} & 0.2583 & 0.2776 & \textcolor{darkblue}{\textbf{\underline{0.2708}}} & 0.2610 & 0.2806 & 0.3056 & 0.2962 & 0.3150 & 0.2735 & 0.2636 & 0.2835 \\
    \cmidrule{1-17}
    \multirow{5}{*}{\textbf{Electricity}}
    & 96 & 0.1609 & 0.1516 & 0.1702 & \textcolor{darkblue}{\textbf{0.1512}} & 0.1464 & 0.1560 & \textcolor{darkblue}{\textbf{\underline{0.1527}}} & 0.1489 & 0.1564 & 0.1639 & 0.1545 & 0.1734 & 0.1573 & 0.1521 & 0.1626 \\
    & 192 & 0.1716 & 0.1671 & 0.1762 & \textcolor{darkblue}{\textbf{0.1641}} & 0.1618 & 0.1665 & \textcolor{darkblue}{\textbf{\underline{0.1710}}} & 0.1687 & 0.1733 & 0.1777 & 0.1716 & 0.1837 & \textcolor{darkblue}{\textbf{\underline{0.1710}}} & 0.1685 & 0.1734 \\
    & 336 & 0.1917 & 0.1862 & 0.1971 & \textcolor{darkblue}{\textbf{0.1828}} & 0.1797 & 0.1860 & \textcolor{darkblue}{\textbf{\underline{0.1906}}} & 0.1871 & 0.1942 & 0.2270 & 0.1800 & 0.2740 & 0.1928 & 0.1888 & 0.1967 \\
    & 720 & 0.2342 & 0.2271 & 0.2412 & \textcolor{darkblue}{\textbf{\underline{0.2236}}} & 0.2195 & 0.2277 & 0.2293 & 0.2240 & 0.2345 & 0.2335 & 0.2250 & 0.2420 & \textcolor{darkblue}{\textbf{0.2218}} & 0.2180 & 0.2255 \\
    & avg & 0.1890 & 0.1836 & 0.1944 & \textcolor{darkblue}{\textbf{0.1798}} & 0.1763 & 0.1833 & 0.1856 & 0.1820 & 0.1893 & 0.2019 & 0.1866 & 0.2172 & \textcolor{darkblue}{\textbf{\underline{0.1849}}} & 0.1816 & 0.1882 \\
    \cmidrule{1-17}
    \multirow{5}{*}{\textbf{Weather}}
    & 96 & 0.1715 & 0.1689 & 0.1741 & \textcolor{darkblue}{\textbf{0.1673}} & 0.1644 & 0.1703 & \textcolor{darkblue}{\textbf{\underline{0.1679}}} & 0.1652 & 0.1707 & 0.2120 & 0.2054 & 0.2186 & 0.1705 & 0.1681 & 0.1728 \\
    & 192 & 0.2182 & 0.2152 & 0.2213 & \textcolor{darkblue}{\textbf{0.2142}} & 0.2111 & 0.2173 & \textcolor{darkblue}{\textbf{\underline{0.2144}}} & 0.2113 & 0.2176 & 0.2519 & 0.2466 & 0.2572 & 0.2147 & 0.2122 & 0.2171 \\
    & 336 & 0.2639 & 0.2613 & 0.2665 & \textcolor{darkblue}{\textbf{\underline{0.2596}}} & 0.2568 & 0.2625 & \textcolor{darkblue}{\textbf{0.2593}} & 0.2568 & 0.2618 & 0.2950 & 0.2894 & 0.3007 & 0.2610 & 0.2585 & 0.2635 \\
    & 720 & 0.3391 & 0.3357 & 0.3425 & \textcolor{darkblue}{\textbf{\underline{0.3357}}} & 0.3322 & 0.3393 & \textcolor{darkblue}{\textbf{0.3345}} & 0.3309 & 0.3381 & 0.3594 & 0.3545 & 0.3642 & 0.3386 & 0.3353 & 0.3419 \\
    & avg & 0.2391 & 0.2315 & 0.2466 & \textcolor{darkblue}{\textbf{\underline{0.2363}}} & 0.2289 & 0.2437 & \textcolor{darkblue}{\textbf{0.2362}} & 0.2289 & 0.2435 & 0.2721 & 0.2645 & 0.2796 & 0.2384 & 0.2309 & 0.2459 \\
    \bottomrule
\end{tabular}
        }
    \end{subtable}

    \vspace{1em}

\end{table*}

\begin{table*}[t]
    \centering
    \scriptsize
    \setlength{\tabcolsep}{1.5pt}
    \renewcommand{\arraystretch}{1.1}
    \caption{Experiment 1 (Dissecting the Backbone): Performance of 6 Datasets across 4 Horizons on 3 Encoders with 95\% Confidence Interval}
    \label{tab:exp1_enc_data_ci}
    \resizebox{\textwidth}{!}{
        \begin{tabular}{l c | ccccc | ccccc | ccccc}
    \toprule
    \multirow{2}{*}{\textbf{Dataset}} & \multirow{2}{*}{\textbf{H}} & \multicolumn{5}{c|}{\textbf{MLP}} & \multicolumn{5}{c|}{\textbf{Transformer}} & \multicolumn{5}{c}{\textbf{Identity}} \\
    \cmidrule(lr){3-7} \cmidrule(lr){8-12} \cmidrule(lr){13-17}
    & & $\hat{\mu}$ & $ci_{low}$ & $ci_{upp}$ & $\hat{\sigma}$ & $min$ & $\hat{\mu}$ & $ci_{low}$ & $ci_{upp}$ & $\hat{\sigma}$ & $min$ & $\hat{\mu}$ & $ci_{low}$ & $ci_{upp}$ & $\hat{\sigma}$ & $min$ \\
    \midrule
    \multirow{5}{*}{\textbf{ETTh1}}
    & 96 & \textcolor{darkblue}{\textbf{\underline{0.4267}}} & 0.4130 & 0.4404 & 0.0816 & \textcolor{darkred}{\textbf{\underline{0.3700}}} & 0.4406 & 0.4255 & 0.4557 & 0.0930 & 0.3721 & \textcolor{darkblue}{\textbf{0.3837}} & 0.3810 & 0.3863 & 0.0231 & \textcolor{darkred}{\textbf{0.3641}} \\
    & 192 & \textcolor{darkblue}{\textbf{\underline{0.4745}}} & 0.4610 & 0.4879 & 0.0827 & \textcolor{darkred}{\textbf{\underline{0.4043}}} & 0.4852 & 0.4709 & 0.4994 & 0.0988 & 0.4091 & \textcolor{darkblue}{\textbf{0.4242}} & 0.4206 & 0.4278 & 0.0306 & \textcolor{darkred}{\textbf{0.3983}} \\
    & 336 & \textcolor{darkblue}{\textbf{\underline{0.5002}}} & 0.4892 & 0.5112 & 0.0787 & \textcolor{darkred}{\textbf{\underline{0.4335}}} & 0.5069 & 0.4957 & 0.5181 & 0.0795 & 0.4406 & \textcolor{darkblue}{\textbf{0.4623}} & 0.4585 & 0.4661 & 0.0268 & \textcolor{darkred}{\textbf{0.4196}} \\
    & 720 & \textcolor{darkblue}{\textbf{\underline{0.5251}}} & 0.5103 & 0.5399 & 0.1039 & 0.4533 & 0.5412 & 0.5259 & 0.5565 & 0.1014 & \textcolor{darkred}{\textbf{\underline{0.4504}}} & \textcolor{darkblue}{\textbf{0.4600}} & 0.4552 & 0.4648 & 0.0406 & \textcolor{darkred}{\textbf{0.4248}} \\
    & avg & \textcolor{darkblue}{\textbf{\underline{0.4798}}} & 0.4726 & 0.4870 & 0.0938 & - & 0.4911 & 0.4835 & 0.4987 & 0.1004 & - & \textcolor{darkblue}{\textbf{0.4316}} & 0.4283 & 0.4349 & 0.0424 & - \\
    \cmidrule{1-17}
    \multirow{5}{*}{\textbf{ETTh2}}
    & 96 & \textcolor{darkblue}{\textbf{\underline{0.3276}}} & 0.3210 & 0.3342 & 0.0423 & \textcolor{darkred}{\textbf{\underline{0.2770}}} & 0.3370 & 0.3312 & 0.3427 & 0.0437 & 0.2835 & \textcolor{darkblue}{\textbf{0.2890}} & 0.2849 & 0.2931 & 0.0368 & \textcolor{darkred}{\textbf{0.2656}} \\
    & 192 & \textcolor{darkblue}{\textbf{\underline{0.3916}}} & 0.3876 & 0.3955 & 0.0293 & \textcolor{darkred}{\textbf{\underline{0.3468}}} & 0.3978 & 0.3938 & 0.4017 & 0.0310 & 0.3485 & \textcolor{darkblue}{\textbf{0.3645}} & 0.3605 & 0.3686 & 0.0373 & \textcolor{darkred}{\textbf{0.3258}} \\
    & 336 & \textcolor{darkblue}{\textbf{\underline{0.4117}}} & 0.4079 & 0.4155 & 0.0288 & \textcolor{darkred}{\textbf{\underline{0.3680}}} & 0.4224 & 0.4189 & 0.4260 & 0.0308 & 0.3788 & \textcolor{darkblue}{\textbf{0.3914}} & 0.3871 & 0.3957 & 0.0325 & \textcolor{darkred}{\textbf{0.3502}} \\
    & 720 & \textcolor{darkblue}{\textbf{\underline{0.4356}}} & 0.4320 & 0.4393 & 0.0339 & \textcolor{darkred}{\textbf{\underline{0.4034}}} & 0.4517 & 0.4465 & 0.4569 & 0.0378 & 0.4065 & \textcolor{darkblue}{\textbf{0.4109}} & 0.4068 & 0.4149 & 0.0374 & \textcolor{darkred}{\textbf{0.3799}} \\
    & avg & \textcolor{darkblue}{\textbf{\underline{0.3937}}} & 0.3898 & 0.3976 & 0.0473 & - & 0.4042 & 0.4002 & 0.4083 & 0.0512 & - & \textcolor{darkblue}{\textbf{0.3666}} & 0.3624 & 0.3709 & 0.0547 & - \\
    \cmidrule{1-17}
    \multirow{5}{*}{\textbf{ETTm1}}
    & 96 & \textcolor{darkblue}{\textbf{\underline{0.3446}}} & 0.3295 & 0.3597 & 0.0785 & \textcolor{darkred}{\textbf{0.2837}} & 0.3607 & 0.3450 & 0.3764 & 0.0956 & \textcolor{darkred}{\textbf{\underline{0.2866}}} & \textcolor{darkblue}{\textbf{0.3191}} & 0.3160 & 0.3221 & 0.0185 & 0.3026 \\
    & 192 & \textcolor{darkblue}{\textbf{\underline{0.3866}}} & 0.3710 & 0.4022 & 0.0695 & \textcolor{darkred}{\textbf{0.3235}} & 0.3984 & 0.3822 & 0.4146 & 0.0890 & \textcolor{darkred}{\textbf{\underline{0.3298}}} & \textcolor{darkblue}{\textbf{0.3592}} & 0.3550 & 0.3633 & 0.0221 & 0.3354 \\
    & 336 & \textcolor{darkblue}{\textbf{\underline{0.4187}}} & 0.4034 & 0.4340 & 0.0734 & \textcolor{darkred}{\textbf{0.3581}} & 0.4284 & 0.4126 & 0.4441 & 0.0876 & \textcolor{darkred}{\textbf{\underline{0.3650}}} & \textcolor{darkblue}{\textbf{0.3922}} & 0.3881 & 0.3963 & 0.0199 & 0.3658 \\
    & 720 & \textcolor{darkblue}{\textbf{\underline{0.4701}}} & 0.4586 & 0.4815 & 0.0697 & \textcolor{darkred}{\textbf{0.4128}} & 0.4759 & 0.4640 & 0.4878 & 0.0830 & \textcolor{darkred}{\textbf{\underline{0.4129}}} & \textcolor{darkblue}{\textbf{0.4397}} & 0.4369 & 0.4425 & 0.0158 & 0.4185 \\
    & avg & \textcolor{darkblue}{\textbf{\underline{0.4035}}} & 0.3952 & 0.4118 & 0.0912 & - & 0.4148 & 0.4064 & 0.4231 & 0.0997 & - & \textcolor{darkblue}{\textbf{0.3762}} & 0.3719 & 0.3805 & 0.0457 & - \\
    \cmidrule{1-17}
    \multirow{5}{*}{\textbf{ETTm2}}
    & 96 & \textcolor{darkblue}{\textbf{\underline{0.1879}}} & 0.1846 & 0.1912 & 0.0197 & 0.1667 & 0.1975 & 0.1939 & 0.2011 & 0.0265 & \textcolor{darkred}{\textbf{\underline{0.1652}}} & \textcolor{darkblue}{\textbf{0.1732}} & 0.1721 & 0.1744 & 0.0087 & \textcolor{darkred}{\textbf{0.1622}} \\
    & 192 & \textcolor{darkblue}{\textbf{\underline{0.2492}}} & 0.2451 & 0.2532 & 0.0213 & \textcolor{darkred}{\textbf{\underline{0.2201}}} & 0.2559 & 0.2524 & 0.2594 & 0.0221 & 0.2313 & \textcolor{darkblue}{\textbf{0.2308}} & 0.2287 & 0.2330 & 0.0133 & \textcolor{darkred}{\textbf{0.2170}} \\
    & 336 & \textcolor{darkblue}{\textbf{\underline{0.3067}}} & 0.3036 & 0.3099 & 0.0198 & \textcolor{darkred}{\textbf{\underline{0.2738}}} & 0.3154 & 0.3125 & 0.3182 & 0.0199 & 0.2794 & \textcolor{darkblue}{\textbf{0.2901}} & 0.2875 & 0.2926 & 0.0194 & \textcolor{darkred}{\textbf{0.2701}} \\
    & 720 & \textcolor{darkblue}{\textbf{\underline{0.4009}}} & 0.3974 & 0.4044 & 0.0232 & \textcolor{darkred}{\textbf{\underline{0.3634}}} & 0.4094 & 0.4060 & 0.4129 & 0.0265 & 0.3653 & \textcolor{darkblue}{\textbf{0.3837}} & 0.3804 & 0.3871 & 0.0270 & \textcolor{darkred}{\textbf{0.3591}} \\
    & avg & \textcolor{darkblue}{\textbf{\underline{0.2846}}} & 0.2774 & 0.2917 & 0.0759 & - & 0.2929 & 0.2858 & 0.3000 & 0.0772 & - & \textcolor{darkblue}{\textbf{0.2679}} & 0.2609 & 0.2749 & 0.0756 & - \\
    \cmidrule{1-17}
    \multirow{5}{*}{\textbf{Electricity}}
    & 96 & \textcolor{darkblue}{\textbf{\underline{0.1580}}} & 0.1537 & 0.1622 & 0.0171 & \textcolor{darkred}{\textbf{\underline{0.1338}}} & \textcolor{darkblue}{\textbf{0.1498}} & 0.1452 & 0.1545 & 0.0194 & \textcolor{darkred}{\textbf{0.1315}} & 0.1645 & 0.1598 & 0.1693 & 0.0254 & 0.1428 \\
    & 192 & \textcolor{darkblue}{\textbf{\underline{0.1713}}} & 0.1690 & 0.1735 & 0.0130 & \textcolor{darkred}{\textbf{\underline{0.1515}}} & \textcolor{darkblue}{\textbf{0.1644}} & 0.1623 & 0.1665 & 0.0129 & \textcolor{darkred}{\textbf{0.1465}} & 0.1731 & 0.1706 & 0.1756 & 0.0161 & 0.1574 \\
    & 336 & \textcolor{darkblue}{\textbf{\underline{0.1918}}} & 0.1886 & 0.1951 & 0.0158 & \textcolor{darkred}{\textbf{\underline{0.1681}}} & \textcolor{darkblue}{\textbf{0.1851}} & 0.1817 & 0.1885 & 0.0153 & \textcolor{darkred}{\textbf{0.1644}} & 0.1928 & 0.1897 & 0.1959 & 0.0179 & 0.1735 \\
    & 720 & \textcolor{darkblue}{\textbf{\underline{0.2282}}} & 0.2242 & 0.2322 & 0.0171 & \textcolor{darkred}{\textbf{\underline{0.2070}}} & \textcolor{darkblue}{\textbf{0.2196}} & 0.2153 & 0.2239 & 0.0171 & \textcolor{darkred}{\textbf{0.1972}} & 0.2326 & 0.2288 & 0.2364 & 0.0191 & 0.2132 \\
    & avg & \textcolor{darkblue}{\textbf{\underline{0.1867}}} & 0.1836 & 0.1899 & 0.0253 & - & \textcolor{darkblue}{\textbf{0.1789}} & 0.1756 & 0.1823 & 0.0242 & avg & 0.1895 & 0.1866 & 0.1924 & 0.0275 & - \\
    \cmidrule{1-17}
    \multirow{5}{*}{\textbf{Weather}}
    & 96 & \textcolor{darkblue}{\textbf{0.1713}} & 0.1681 & 0.1744 & 0.0180 & \textcolor{darkred}{\textbf{0.1468}} & \textcolor{darkblue}{\textbf{\underline{0.1731}}} & 0.1702 & 0.1760 & 0.0198 & \textcolor{darkred}{\textbf{\underline{0.1474}}} & 0.1780 & 0.1766 & 0.1795 & 0.0101 & 0.1587 \\
    & 192 & \textcolor{darkblue}{\textbf{0.2179}} & 0.2145 & 0.2213 & 0.0192 & \textcolor{darkred}{\textbf{\underline{0.1914}}} & 0.2248 & 0.2208 & 0.2288 & 0.0235 & \textcolor{darkred}{\textbf{0.1912}} & \textcolor{darkblue}{\textbf{\underline{0.2245}}} & 0.2230 & 0.2260 & 0.0087 & 0.2090 \\
    & 336 & \textcolor{darkblue}{\textbf{\underline{0.2673}}} & 0.2639 & 0.2707 & 0.0224 & \textcolor{darkred}{\textbf{0.2415}} & \textcolor{darkblue}{\textbf{0.2663}} & 0.2631 & 0.2696 & 0.0237 & \textcolor{darkred}{\textbf{\underline{0.2422}}} & 0.2707 & 0.2693 & 0.2722 & 0.0084 & 0.2605 \\
    & 720 & \textcolor{darkblue}{\textbf{\underline{0.3435}}} & 0.3400 & 0.3470 & 0.0192 & \textcolor{darkred}{\textbf{0.3131}} & 0.3457 & 0.3421 & 0.3492 & 0.0209 & \textcolor{darkred}{\textbf{\underline{0.3179}}} & \textcolor{darkblue}{\textbf{0.3424}} & 0.3399 & 0.3448 & 0.0129 & 0.3257 \\
    & avg & \textcolor{darkblue}{\textbf{0.2415}} & 0.2359 & 0.2470 & 0.0569 & - & \textcolor{darkblue}{\textbf{\underline{0.2439}}} & 0.2382 & 0.2497 & 0.0588 & - & 0.2465 & 0.2414 & 0.2516 & 0.0514 & - \\
    \bottomrule
\end{tabular}
    }
\end{table*}

\begin{table*}[t]
    \centering
    \scriptsize
    \setlength{\tabcolsep}{1.5pt}
    \renewcommand{\arraystretch}{1.1}
    \caption{
        Experiment 2 (Auditing Input Transformations): Marginalized performance and stability with 95\% Confidence Intervals.
        Highlighting follows the convention in \cref{tab:exp1_emb_sub_data}.
    }
    \label{tab:overall_results_exp2}

    {\small (a) Performance across 2 Embeddings.} \\
    \resizebox{0.7\textwidth}{!}{
        \begin{tabular}{l | ccccc | ccccc}
    \toprule
    \multirow{2}{*}{\textbf{Dataset}} & 
    \multicolumn{5}{c|}{\textbf{Channel}} & 
    \multicolumn{5}{c}{\textbf{Point}} \\
    \cmidrule(lr){2-6} \cmidrule(lr){7-11}
    & $\hat{\mu}$ & $ci_{low}$ & $ci_{upp}$ & $\hat{\sigma}$ & $B$ & $\hat{\mu}$ & $ci_{low}$ & $ci_{upp}$ & $\hat{\sigma}$ & $B$ \\
    \midrule
    \textbf{ETTh1} & 0.4028 & 0.3943 & 0.4114 & 0.0014 & \textcolor{darkred}{\textbf{0.3743}} & \textcolor{darkblue}{\textbf{0.3866}} & 0.3845 & 0.3887 & 0.0001 & 0.3747 \\
    \textbf{ETTh2} & \textcolor{darkblue}{\textbf{0.3010}} & 0.2987 & 0.3033 & 0.0002 & \textcolor{darkred}{\textbf{0.2818}} & 0.3028 & 0.2987 & 0.3069 & 0.0006 & 0.2829 \\
    \textbf{ETTm1} & 0.3462 & 0.3431 & 0.3492 & 0.0003 & 0.3160 & \textcolor{darkblue}{\textbf{0.3362}} & 0.3334 & 0.3390 & 0.0003 & \textcolor{darkred}{\textbf{0.3137}} \\
    \textbf{ETTm2} & 0.1830 & 0.1822 & 0.1838 & 0.0000 & \textcolor{darkred}{\textbf{0.1738}} & \textcolor{darkblue}{\textbf{0.1822}} & 0.1818 & 0.1826 & 0.0000 & 0.1778 \\
    \textbf{Electricity} & 0.2018 & 0.1971 & 0.2066 & 0.0008 & \textcolor{darkred}{\textbf{0.1580}} & \textcolor{darkblue}{\textbf{0.1976}} & 0.1959 & 0.1993 & 0.0001 & 0.1840 \\
    \textbf{Weather} & \textcolor{darkblue}{\textbf{0.1903}} & 0.1881 & 0.1925 & 0.0001 & \textcolor{darkred}{\textbf{0.1599}} & 0.1913 & 0.1904 & 0.1922 & 0.0000 & 0.1816 \\
    \bottomrule
\end{tabular}
    }

    \vspace{1.5em}

    {\small (b) Performance across different Encoders.} \\
    \resizebox{\textwidth}{!}{
        \begin{tabular}{l | ccccc | ccccc | ccccc}
    \toprule
    \multirow{2}{*}{\textbf{Dataset}} & \multicolumn{5}{c|}{\textbf{Identity}} & \multicolumn{5}{c|}{\textbf{MLP}} & \multicolumn{5}{c}{\textbf{Transformer}} \\
\cmidrule(lr){2-6} \cmidrule(lr){7-11} \cmidrule(lr){12-16}
    & $\hat{\mu}$ &$ci_{low}$ &$ci_{upp}$ & $\hat{\sigma}$ & $min$ & $\hat{\mu}$ &$ci_{low}$ &$ci_{upp}$ & $\hat{\sigma}$ & $min$ & $\hat{\mu}$ &$ci_{low}$ &$ci_{upp}$ & $\hat{\sigma}$ & $min$ \\
    \midrule
    \textbf{ETTh1} & \textcolor{darkblue}{\textbf{0.3932}} & 0.3881 & 0.3982 & 0.0006 & \textcolor{darkred}{\textbf{0.3743}} & \textcolor{darkblue}{\underline{\textbf{0.3971}}} & 0.3837 & 0.4104 & 0.0011 & \textcolor{darkred}{\underline{\textbf{0.3749}}} & 0.3984 & 0.3852 & 0.4116 & 0.0011 & 0.3773 \\
    \textbf{ETTh2} & \textcolor{darkblue}{\textbf{0.2908}} & 0.2890 & 0.2926 & 0.0001 & \textcolor{darkred}{\underline{\textbf{0.2829}}} & \textcolor{darkblue}{\underline{\textbf{0.3052}}} & 0.3027 & 0.3077 & 0.0002 & \textcolor{darkred}{\textbf{0.2818}} & 0.3097 & 0.3040 & 0.3155 & 0.0008 & 0.2873 \\
    \textbf{ETTm1} & 0.3613 & 0.3599 & 0.3627 & 0.0001 & 0.3522 & \textcolor{darkblue}{\underline{\textbf{0.3314}}} & 0.3290 & 0.3339 & 0.0002 & \textcolor{darkred}{\underline{\textbf{0.3159}}} & \textcolor{darkblue}{\textbf{0.3308}} & 0.3277 & 0.3338 & 0.0002 & \textcolor{darkred}{\textbf{0.3137}} \\
    \textbf{ETTm2} & \textcolor{darkblue}{\underline{\textbf{0.1827}}} & 0.1823 & 0.1830 & 0.0000 & 0.1804 & \textcolor{darkblue}{\textbf{0.1805}} & 0.1788 & 0.1822 & 0.0000 & \textcolor{darkred}{\textbf{0.1738}} & 0.1843 & 0.1830 & 0.1857 & 0.0000 & \textcolor{darkred}{\underline{\textbf{0.1770}}} \\
    \textbf{Electricity} & 0.2078 & 0.2048 & 0.2109 & 0.0002 & 0.1981 & \textcolor{darkblue}{\textbf{0.1930}} & 0.1886 & 0.1973 & 0.0005 & \textcolor{darkred}{\underline{\textbf{0.1588}}} & \textcolor{darkblue}{\underline{\textbf{0.1983}}} & 0.1933 & 0.2033 & 0.0006 & \textcolor{darkred}{\textbf{0.1580}} \\
    \textbf{Weather} & 0.1935 & 0.1931 & 0.1939 & 0.0000 & 0.1899 & \textcolor{darkblue}{\textbf{0.1807}} & 0.1761 & 0.1853 & 0.0001 & \textcolor{darkred}{\textbf{0.1599}} & \textcolor{darkblue}{\underline{\textbf{0.1901}}} & 0.1873 & 0.1929 & 0.0000 & \textcolor{darkred}{\underline{\textbf{0.1785}}} \\
    \bottomrule
\end{tabular}
    }
    {\small (c) Performance across different Input Transformations.} \\
    \resizebox{\textwidth}{!}{
        \begin{tabular}{l | ccccc | ccccc | ccccc | ccccc}
    \toprule
    \multirow{2}{*}{\textbf{Dataset}} & \multicolumn{5}{c|}{\textbf{BaseLine}} & \multicolumn{5}{c|}{\textbf{Cycle}} & \multicolumn{5}{c|}{\textbf{MultiScale}} & \multicolumn{5}{c}{\textbf{TrendSeasonal}} \\
\cmidrule(lr){2-6} \cmidrule(lr){7-11} \cmidrule(lr){12-16} \cmidrule(lr){17-21}
    & $\hat{\mu}$ & $ci_{low}$ &$ci_{upp}$ & $\hat{\sigma}$ & $min$ & $\hat{\mu}$ & $ci_{low}$ &$ci_{upp}$ & $\hat{\sigma}$ & $min$ & $\hat{\mu}$ & $ci_{low}$ &$ci_{upp}$ & $\hat{\sigma}$ & $min$ & $\hat{\mu}$ & $ci_{low}$ &$ci_{upp}$ & $\hat{\sigma}$ & $min$ \\
    \midrule
    \textbf{ETTh1} & 0.3975 & 0.3902 & 0.4048 & 0.0010 & \textcolor{darkred}{\underline{\textbf{0.3749}}} & \textcolor{darkblue}{\textbf{0.3891}} & 0.3777 & 0.4004 & 0.0008 & \textcolor{darkred}{\textbf{0.3743}} & \textcolor{darkblue}{\underline{\textbf{0.3921}}} & 0.3848 & 0.3994 & 0.0003 & 0.3829 & 0.3947 & 0.3848 & 0.4046 & 0.0006 & 0.3811 \\
    \textbf{ETTh2} & \textcolor{darkblue}{\underline{\textbf{0.3013}}} & 0.2986 & 0.3040 & 0.0001 & 0.2868 & \textcolor{darkblue}{\textbf{0.2969}} & 0.2943 & 0.2996 & 0.0001 & \textcolor{darkred}{\textbf{0.2818}} & 0.3071 & 0.2991 & 0.3150 & 0.0012 & 0.2874 & 0.3024 & 0.2993 & 0.3055 & 0.0002 & \textcolor{darkred}{\underline{\textbf{0.2849}}} \\
    \textbf{ETTm1} & 0.3451 & 0.3404 & 0.3499 & 0.0004 & \textcolor{darkred}{\underline{\textbf{0.3139}}} & 0.3431 & 0.3386 & 0.3475 & 0.0004 & \textcolor{darkred}{\textbf{0.3137}} & \textcolor{darkblue}{\textbf{0.3380}} & 0.3341 & 0.3420 & 0.0003 & 0.3159 & \textcolor{darkblue}{\underline{\textbf{0.3384}}} & 0.3346 & 0.3423 & 0.0003 & 0.3166 \\
    \textbf{ETTm2} & 0.1827 & 0.1818 & 0.1835 & 0.0000 & \textcolor{darkred}{\textbf{0.1738}} & \textcolor{darkblue}{\underline{\textbf{0.1825}}} & 0.1818 & 0.1833 & 0.0000 & \textcolor{darkred}{\underline{\textbf{0.1804}}} & 0.1830 & 0.1825 & 0.1834 & 0.0000 & 0.1818 & \textcolor{darkblue}{\textbf{0.1821}} & 0.1815 & 0.1826 & 0.0000 & 0.1805 \\
    \textbf{Electricity} & 0.2061 & 0.2006 & 0.2116 & 0.0006 & \textcolor{darkred}{\underline{\textbf{0.1679}}} & 0.1999 & 0.1947 & 0.2051 & 0.0005 & \textcolor{darkred}{\textbf{0.1580}} & \textcolor{darkblue}{\textbf{0.1952}} & 0.1905 & 0.1999 & 0.0004 & \textcolor{darkred}{\textbf{0.1580}} & \textcolor{darkblue}{\underline{\textbf{0.1976}}} & 0.1931 & 0.2022 & 0.0004 & 0.1729 \\
    \textbf{Weather} & \textcolor{darkblue}{\textbf{0.1882}} & 0.1860 & 0.1904 & 0.0001 & \textcolor{darkred}{\textbf{0.1599}} & \textcolor{darkblue}{\underline{\textbf{0.1917}}} & 0.1910 & 0.1924 & 0.0000 & \textcolor{darkred}{\underline{\textbf{0.1899}}} & 0.1950 & 0.1942 & 0.1959 & 0.0000 & 0.1915 & 0.1934 & 0.1931 & 0.1937 & 0.0000 & 0.1920 \\
    \bottomrule
\end{tabular}
    }
\end{table*}

\begin{table*}[t]
    \centering
    \scriptsize
    \setlength{\tabcolsep}{1.5pt}
    \renewcommand{\arraystretch}{1.1}
    \caption{Experiment 3 (Revisiting Spectral Processing): Performance of 6 Datasets across 4 Horizons on 3 Attentions.}
    \label{tab:exp3_enc_ci}
    \resizebox{\textwidth}{!}{
        \begin{tabular}{l c | ccccc | ccccc | ccccc}
    \toprule
    \multirow{2}{*}{\textbf{Dataset}} & \multirow{2}{*}{\textbf{H}} & \multicolumn{5}{c|}{\textbf{Transformer}} & \multicolumn{5}{c|}{\textbf{GeomAttention}} & \multicolumn{5}{c}{\textbf{Variate wise and Identity}} \\
\cmidrule(lr){3-7} \cmidrule(lr){8-12} \cmidrule(lr){13-17}
    & & $\mu$ & $ci_{low}$ & $ci_{upp}$ & $\hat{\sigma}$ & $min$ & $\mu$ & $ci_{low}$ & $ci_{upp}$ & $\hat{\sigma}$ & $min$ & $\mu$ & $ci_{low}$ & $ci_{upp}$ & $\hat{\sigma}$ & $min$ \\
    \midrule
    \multirow{5}{*}{\textbf{ETTh1}}
    & 96 & 0.3934 & 0.3882 & 0.3985 & 0.0141 & 0.3789 & \textcolor{darkblue}{\underline{\textbf{0.3812}}} & 0.3783 & 0.3841 & 0.0100 & \textcolor{darkred}{\underline{\textbf{0.3730}}} & \textcolor{darkblue}{\textbf{0.3805}} & 0.3761 & 0.3848 & 0.0098 & \textcolor{darkred}{\textbf{0.3641}} \\
    & 192 & 0.4354 & 0.4316 & 0.4392 & 0.0100 & 0.4171 & \textcolor{darkblue}{\underline{\textbf{0.4287}}} & 0.4246 & 0.4327 & 0.0100 & \textcolor{darkred}{\underline{\textbf{0.4148}}} & \textcolor{darkblue}{\textbf{0.4199}} & 0.4121 & 0.4277 & 0.0179 & \textcolor{darkred}{\textbf{0.3983}} \\
    & 336 & 0.4656 & 0.4608 & 0.4704 & 0.0141 & 0.4434 & \textcolor{darkblue}{\underline{\textbf{0.4601}}} & 0.4552 & 0.4651 & 0.0141 & \textcolor{darkred}{\underline{\textbf{0.4314}}} & \textcolor{darkblue}{\textbf{0.4583}} & 0.4496 & 0.4671 & 0.0204 & \textcolor{darkred}{\textbf{0.4249}} \\
    & 720 & 0.4711 & 0.4645 & 0.4778 & 0.0141 & 0.4504 & \textcolor{darkblue}{\underline{\textbf{0.4646}}} & 0.4581 & 0.4710 & 0.0141 & \textcolor{darkred}{\underline{\textbf{0.4473}}} & \textcolor{darkblue}{\textbf{0.4520}} & 0.4424 & 0.4617 & 0.0191 & \textcolor{darkred}{\textbf{0.4366}} \\
    & avg & 0.4401 & 0.4336 & 0.4467 & 0.0332 & - & \textcolor{darkblue}{\underline{\textbf{0.4324}}} & 0.4254 & 0.4393 & 0.0361 & - & \textcolor{darkblue}{\textbf{0.4271}} & 0.4191 & 0.4351 & 0.0355 & - \\
    \cmidrule{1-17}
    \multirow{5}{*}{\textbf{ETTh2}}
    & 96 & 0.3214 & 0.3117 & 0.3312 & 0.0224 & 0.2964 & \textcolor{darkblue}{\underline{\textbf{0.2945}}} & 0.2918 & 0.2973 & 0.0000 & \textcolor{darkred}{\underline{\textbf{0.2840}}} & \textcolor{darkblue}{\textbf{0.2809}} & 0.2755 & 0.2864 & 0.0100 & \textcolor{darkred}{\textbf{0.2656}} \\
    & 192 & 0.3887 & 0.3826 & 0.3948 & 0.0173 & 0.3599 & \textcolor{darkblue}{\underline{\textbf{0.3709}}} & 0.3669 & 0.3749 & 0.0100 & \textcolor{darkred}{\underline{\textbf{0.3520}}} & \textcolor{darkblue}{\textbf{0.3584}} & 0.3493 & 0.3674 & 0.0200 & \textcolor{darkred}{\textbf{0.3264}} \\
    & 336 & 0.4076 & 0.3997 & 0.4155 & 0.0200 & 0.3809 & \textcolor{darkblue}{\underline{\textbf{0.4030}}} & 0.3953 & 0.4107 & 0.0200 & \textcolor{darkred}{\underline{\textbf{0.3727}}} & \textcolor{darkblue}{\textbf{0.3852}} & 0.3736 & 0.3967 & 0.0265 & \textcolor{darkred}{\textbf{0.3509}} \\
    & 720 & 0.4309 & 0.4216 & 0.4401 & 0.0224 & 0.4065 & \textcolor{darkblue}{\underline{\textbf{0.4148}}} & 0.4099 & 0.4197 & 0.0100 & \textcolor{darkred}{\underline{\textbf{0.3921}}} & \textcolor{darkblue}{\textbf{0.4030}} & 0.3960 & 0.4099 & 0.0141 & \textcolor{darkred}{\textbf{0.3841}} \\
    & avg & 0.3894 & 0.3809 & 0.3979 & 0.0436 & - & \textcolor{darkblue}{\underline{\textbf{0.3736}}} & 0.3646 & 0.3827 & 0.0458 & - & \textcolor{darkblue}{\textbf{0.3597}} & 0.3486 & 0.3708 & 0.0490 & - \\
    \cmidrule{1-17}
    \multirow{5}{*}{\textbf{ETTm1}}
    & 96 & 0.3507 & 0.3135 & 0.3879 & 0.1020 & \textcolor{darkred}{\underline{\textbf{0.2987}}} & \textcolor{darkblue}{\textbf{0.2974}} & 0.2915 & 0.3034 & 0.0173 & \textcolor{darkred}{\textbf{0.2825}} & \textcolor{darkblue}{\underline{\textbf{0.3260}}} & 0.3157 & 0.3364 & 0.0245 & 0.3040 \\
    & 192 & 0.3670 & 0.3458 & 0.3883 & 0.0520 & \textcolor{darkred}{\underline{\textbf{0.3319}}} & \textcolor{darkblue}{\textbf{0.3454}} & 0.3378 & 0.3530 & 0.0173 & \textcolor{darkred}{\textbf{0.3232}} & \textcolor{darkblue}{\underline{\textbf{0.3584}}} & 0.3457 & 0.3710 & 0.0265 & 0.3367 \\
    & 336 & 0.4044 & 0.3784 & 0.4305 & 0.0608 & \textcolor{darkred}{\underline{\textbf{0.3655}}} & \textcolor{darkblue}{\textbf{0.3792}} & 0.3716 & 0.3869 & 0.0173 & \textcolor{darkred}{\textbf{0.3590}} & \textcolor{darkblue}{\underline{\textbf{0.3932}}} & 0.3792 & 0.4073 & 0.0265 & 0.3676 \\
    & 720 & \textcolor{darkblue}{\underline{\textbf{0.4447}}} & 0.4374 & 0.4520 & 0.0200 & 0.4194 & \textcolor{darkblue}{\textbf{0.4334}} & 0.4269 & 0.4399 & 0.0173 & \textcolor{darkred}{\textbf{0.4183}} & 0.4478 & 0.4359 & 0.4598 & 0.0265 & \textcolor{darkred}{\underline{\textbf{0.4192}}} \\
    & avg & 0.3911 & 0.3761 & 0.4061 & 0.0762 & - & \textcolor{darkblue}{\textbf{0.3624}} & 0.3516 & 0.3732 & 0.0548 & - & \textcolor{darkblue}{\underline{\textbf{0.3795}}} & 0.3671 & 0.3919 & 0.0548 & - \\
    \cmidrule{1-17}
    \multirow{5}{*}{\textbf{ETTm2}}
    & 96 & 0.1980 & 0.1876 & 0.2084 & 0.0283 & 0.1782 & \textcolor{darkblue}{\underline{\textbf{0.1738}}} & 0.1717 & 0.1760 & 0.0000 & \textcolor{darkred}{\underline{\textbf{0.1646}}} & \textcolor{darkblue}{\textbf{0.1731}} & 0.1697 & 0.1766 & 0.0079 & \textcolor{darkred}{\textbf{0.1622}} \\
    & 192 & 0.2478 & 0.2453 & 0.2504 & 0.0000 & 0.2399 & \textcolor{darkblue}{\underline{\textbf{0.2332}}} & 0.2297 & 0.2367 & 0.0100 & \textcolor{darkred}{\underline{\textbf{0.2229}}} & \textcolor{darkblue}{\textbf{0.2296}} & 0.2238 & 0.2354 & 0.0119 & \textcolor{darkred}{\textbf{0.2184}} \\
    & 336 & 0.3080 & 0.3025 & 0.3135 & 0.0141 & 0.2894 & \textcolor{darkblue}{\underline{\textbf{0.2930}}} & 0.2878 & 0.2982 & 0.0141 & \textcolor{darkred}{\underline{\textbf{0.2735}}} & \textcolor{darkblue}{\textbf{0.2866}} & 0.2802 & 0.2930 & 0.0150 & \textcolor{darkred}{\textbf{0.2707}} \\
    & 720 & 0.3981 & 0.3906 & 0.4056 & 0.0200 & 0.3696 & \textcolor{darkblue}{\underline{\textbf{0.3846}}} & 0.3781 & 0.3911 & 0.0173 & \textcolor{darkred}{\underline{\textbf{0.3639}}} & \textcolor{darkblue}{\textbf{0.3793}} & 0.3716 & 0.3870 & 0.0184 & \textcolor{darkred}{\textbf{0.3594}} \\
    & avg & 0.2870 & 0.2713 & 0.3027 & 0.0800 & - & \textcolor{darkblue}{\textbf{0.2697}} & 0.2536 & 0.2859 & 0.0825 & - & \textcolor{darkblue}{\underline{\textbf{0.2721}}} & 0.2545 & 0.2897 & 0.0798 & - \\
    \cmidrule{1-17}
    \multirow{5}{*}{\textbf{Electricity}}
    & 96 & \textcolor{darkblue}{\textbf{0.1517}} & 0.1434 & 0.1600 & 0.0173 & \textcolor{darkred}{\underline{\textbf{0.1335}}} & \textcolor{darkblue}{\underline{\textbf{0.1537}}} & 0.1449 & 0.1625 & 0.0200 & \textcolor{darkred}{\textbf{0.1322}} & 0.1702 & 0.1574 & 0.1830 & 0.0270 & 0.1428 \\
    & 192 & \textcolor{darkblue}{\underline{\textbf{0.1705}}} & 0.1655 & 0.1755 & 0.0141 & \textcolor{darkred}{\underline{\textbf{0.1555}}} & \textcolor{darkblue}{\textbf{0.1701}} & 0.1646 & 0.1757 & 0.0173 & \textcolor{darkred}{\textbf{0.1493}} & 0.1810 & 0.1735 & 0.1885 & 0.0184 & 0.1584 \\
    & 336 & \textcolor{darkblue}{\underline{\textbf{0.1930}}} & 0.1850 & 0.2011 & 0.0224 & \textcolor{darkred}{\underline{\textbf{0.1722}}} & \textcolor{darkblue}{\textbf{0.1929}} & 0.1841 & 0.2017 & 0.0224 & \textcolor{darkred}{\textbf{0.1648}} & 0.2015 & 0.1915 & 0.2116 & 0.0230 & 0.1756 \\
    & 720 & \textcolor{darkblue}{\underline{\textbf{0.2287}}} & 0.2180 & 0.2393 & 0.0245 & \textcolor{darkred}{\underline{\textbf{0.1972}}} & \textcolor{darkblue}{\textbf{0.2244}} & 0.2127 & 0.2361 & 0.0283 & \textcolor{darkred}{\textbf{0.1964}} & 0.2448 & 0.2333 & 0.2564 & 0.0236 & 0.2137 \\
    & avg & \textcolor{darkblue}{\underline{\textbf{0.1858}}} & 0.1793 & 0.1923 & 0.0332 & - & \textcolor{darkblue}{\textbf{0.1851}} & 0.1787 & 0.1915 & 0.0332 & - & 0.1974 & 0.1895 & 0.2053 & 0.0351 & - \\
    \cmidrule{1-17}
    \multirow{5}{*}{\textbf{Weather}}
    & 96 & \textcolor{darkblue}{\underline{\textbf{0.1738}}} & 0.1656 & 0.1821 & 0.0224 & \textcolor{darkred}{\underline{\textbf{0.1544}}} & \textcolor{darkblue}{\textbf{0.1564}} & 0.1522 & 0.1606 & 0.0100 & \textcolor{darkred}{\textbf{0.1440}} & 0.1830 & 0.1799 & 0.1861 & 0.0100 & 0.1724 \\
    & 192 & \textcolor{darkblue}{\underline{\textbf{0.2230}}} & 0.2140 & 0.2321 & 0.0224 & \textcolor{darkred}{\underline{\textbf{0.1976}}} & \textcolor{darkblue}{\textbf{0.2030}} & 0.1982 & 0.2079 & 0.0100 & \textcolor{darkred}{\textbf{0.1873}} & 0.2267 & 0.2228 & 0.2305 & 0.0100 & 0.2155 \\
    & 336 & \textcolor{darkblue}{\underline{\textbf{0.2610}}} & 0.2550 & 0.2670 & 0.0141 & \textcolor{darkred}{\underline{\textbf{0.2465}}} & \textcolor{darkblue}{\textbf{0.2524}} & 0.2472 & 0.2575 & 0.0141 & \textcolor{darkred}{\textbf{0.2376}} & 0.2724 & 0.2668 & 0.2780 & 0.0100 & 0.2605 \\
    & 720 & \textcolor{darkblue}{\underline{\textbf{0.3384}}} & 0.3315 & 0.3453 & 0.0141 & \textcolor{darkred}{\underline{\textbf{0.3179}}} & \textcolor{darkblue}{\textbf{0.3305}} & 0.3242 & 0.3369 & 0.0141 & \textcolor{darkred}{\textbf{0.3137}} & 0.3412 & 0.3339 & 0.3486 & 0.0141 & 0.3259 \\
    & avg & \textcolor{darkblue}{\underline{\textbf{0.2412}}} & 0.2290 & 0.2534 & 0.0624 & - & \textcolor{darkblue}{\textbf{0.2274}} & 0.2147 & 0.2400 & 0.0640 & - & 0.2499 & 0.2360 & 0.2638 & 0.0608 & - \\
    \bottomrule
\end{tabular}
    }
\end{table*}

\subsection{Statistical Significance Analysis}
\label{sec:stat_test}

To rigorously validate the ``Identity'' Paradox, we perform one-tailed Mann--Whitney U tests ($\alpha=0.05$) on the paired EC samples to assess whether the Identity encoder achieves significantly lower MSE than alternative encoders.

\begin{table}[th]
    \centering
    \caption{Mann--Whitney U test p-values (one-tailed, $\alpha=0.05$) for Identity Encoder vs.\ other encoders per dataset. Significant results ($p < 0.05$) are shown in \textbf{bold}.}
    \label{tab:mw_per_dataset}
    \setlength{\tabcolsep}{6pt}
    \begin{tabular}{l|cc}
        \toprule
        Dataset & vs.\ Transformer & vs.\ MLP \\
        \midrule
        ECL     & 0.9903 & 0.8863 \\
        ETTh1   & \textbf{$<$0.0001} & \textbf{$<$0.0001} \\
        ETTh2   & \textbf{$<$0.0001} & \textbf{$<$0.0001} \\
        ETTm1   & 0.4845 & 0.9647 \\
        ETTm2   & \textbf{0.0041} & 0.1111 \\
        Weather & 0.9859 & 0.9995 \\
        \bottomrule
    \end{tabular}
\end{table}

\begin{table}[th]
    \centering
    \caption{Mann--Whitney U test p-values (one-tailed, $\alpha=0.05$) for Identity Encoder vs.\ other encoders across all datasets combined.}
    \label{tab:mw_overall}
    \setlength{\tabcolsep}{6pt}
    \begin{tabular}{l|cc}
        \toprule
        Comparison & $p$-value & Significant \\
        \midrule
        Identity vs.\ Transformer & \textbf{$<$0.0001} & Yes \\
        Identity vs.\ MLP        & \textbf{0.0115}    & Yes \\
        \bottomrule
    \end{tabular}
\end{table}

As shown in \cref{tab:mw_per_dataset}, Identity is significantly better than Transformer in 3/6 datasets and MLP in 2/6 datasets ($p < 0.05$). When aggregating across all datasets (\cref{tab:mw_overall}), Identity achieves significantly lower MSE than both encoders, confirming that the Identity Paradox is not an artifact of selective reporting but a statistically defensible finding under our probabilistic protocol.

\subsection{Generalizability to Non-Stationary Domains}
\label{sec:exchange_rate}

To assess whether the Identity Paradox extends beyond the six standard benchmarks (which are dominated by periodic, low-noise signals), we evaluate on the \textbf{Exchange-Rate} dataset, a widely used non-stationary benchmark with volatile, non-periodic dynamics.
We conduct a targeted hyperparameter search for two representative Identity-based models—\textbf{PatchLinear} (Patch-wise Embedding + Identity Encoder) and \textbf{iLinear} (Variate-wise Embedding + Identity Encoder)—and compare against iTransformer and PatchTST under their respective published configurations.

\begin{table}[th]
    \centering
    \caption{MSE results on the Exchange-Rate dataset. PatchLinear and iLinear (Identity Encoder) remain competitive with or outperform full architectures on this non-stationary benchmark.}
    \label{tab:exchange_rate}
    \setlength{\tabcolsep}{5pt}
    \begin{tabular}{c|cccc|c}
        \toprule
        pred\_len & iLinear & PatchLinear & iTransformer & PatchTST & Avg \\
        \midrule
        96  & 0.094 & \textbf{0.083} & 0.086 & 0.088 & 0.088 \\
        192 & 0.187 & 0.176 & 0.177 & \textbf{0.176} & 0.179 \\
        336 & 0.343 & 0.325 & 0.331 & \textbf{0.301} & 0.325 \\
        720 & 0.874 & \textbf{0.844} & 0.847 & 0.901 & 0.867 \\
        \midrule
        Avg & 0.375 & \textbf{0.357} & 0.360 & 0.367 & --- \\
        \bottomrule
    \end{tabular}
\end{table}

As shown in \cref{tab:exchange_rate}, PatchLinear achieves the lowest average MSE (0.357) across all prediction horizons, outperforming both iTransformer (0.360) and PatchTST (0.367).
This confirms that the Identity Paradox is not an artifact of benchmark periodicity: even on a non-stationary, volatile dataset, Identity-based models with well-designed embeddings remain competitive, reinforcing the dominance of data view over Encoder complexity.

\subsection{Hyperparameter Tunability Analysis}
\label{sec:hparam_tunability}

A key concern regarding the Identity Paradox is whether the observed performance advantage stems from favorable default hyperparameters rather than intrinsic architectural simplicity.
To address this, we analyze the \textbf{optimal hyperparameter configurations} identified by our paired EC sampling for representative models on two contrasting datasets.

\paragraph{Robust Case: Electricity.}
On Electricity, both PatchLinear (Identity Encoder) and PatchTST (Transformer Encoder) converge to remarkably consistent optimal configurations across all prediction horizons (Table~\ref{tab:hparam_robust}).
Specifically, both models consistently select $\texttt{seq\_len}=512$ and $\eta=0.001$, indicating that on datasets with strong periodicity and ample training data, the hyperparameter landscape is smooth and model-agnostic.

\begin{table}[th]
    \centering
    \caption{Optimal hyperparameter configurations on Electricity (robust case). Both models exhibit consistent configurations across all horizons.}
    \label{tab:hparam_robust}
    \setlength{\tabcolsep}{5pt}
    \begin{tabular}{c|ccc|ccc}
        \toprule
        & \multicolumn{3}{c|}{PatchLinear (Identity)} & \multicolumn{3}{c}{PatchTST (Transformer)} \\
        \cmidrule(lr){2-4} \cmidrule(lr){5-7}
        pred\_len & seq\_len & $d_{\text{model}}$ & $\eta$ & seq\_len & $d_{\text{model}}$ & $\eta$ \\
        \midrule
        96  & 512 & 512 & 1e-3 & 512 & 128 & 1e-3 \\
        192 & 512 & 512 & 1e-3 & 512 & 128 & 1e-3 \\
        336 & 512 & 512 & 1e-3 & 512 & 128 & 1e-3 \\
        720 & 512 & 512 & 1e-3 & 512 & 128 & 1e-3 \\
        \bottomrule
    \end{tabular}
\end{table}

\paragraph{Brittle Case: ETTm1.}
On ETTm1, the contrast is stark (Table~\ref{tab:hparam_brittle}).
While PatchLinear maintains a stable configuration ($\texttt{seq\_len}=512$, $\eta=0.001$), PatchTST's optimal $d_{\text{model}}$ fluctuates substantially (64$\to$256$\to$128) and the learning rate oscillates between $0.001$ and $0.0001$.
This brittleness confirms that the Transformer's performance is highly sensitive to hyperparameter choices, while the Identity Encoder's simplicity inherently avoids such tuning instability.

\begin{table}[th]
    \centering
    \caption{Optimal hyperparameter configurations on ETTm1 (brittle case). PatchTST's optimal configuration varies considerably across horizons, while PatchLinear remains stable.}
    \label{tab:hparam_brittle}
    \setlength{\tabcolsep}{5pt}
    \begin{tabular}{c|ccc|ccc}
        \toprule
        & \multicolumn{3}{c|}{PatchLinear (Identity)} & \multicolumn{3}{c}{PatchTST (Transformer)} \\
        \cmidrule(lr){2-4} \cmidrule(lr){5-7}
        pred\_len & seq\_len & $d_{\text{model}}$ & $\eta$ & seq\_len & $d_{\text{model}}$ & $\eta$ \\
        \midrule
        96  & 512 & 512 & 1e-3  & 512 & 64  & 1e-3   \\
        192 & 512 & 512 & 1e-3  & 512 & 256 & 1e-4   \\
        336 & 512 & 512 & 1e-3  & 512 & 128 & 1e-4   \\
        720 & 512 & 256 & 1e-3  & 512 & 256 & 1e-3   \\
        \bottomrule
    \end{tabular}
\end{table}

\subsection{Multi-Seed Robustness Verification}
\label{sec:multiseed}

To verify that the Identity Paradox is not an artifact of a specific random seed, we repeat the full paired EC evaluation on ETTh1 using three independent seeds (333, 2025, 2026) for both PatchLinear (Patch-wise Embedding + Identity Encoder) and PatchTST (Patch-wise Embedding + Transformer Encoder).

\begin{table}[th]
    \centering
    \caption{Multi-seed robustness verification on ETTh1. Both $\hat{\mu}$ and $\hat{\sigma}$ remain highly stable across seeds for both models, confirming that our conclusions are not seed-sensitive.}
    \label{tab:multiseed}
    \setlength{\tabcolsep}{6pt}
    \begin{tabular}{c|c|ccc}
        \toprule
        Model & Metric & seed 333 & seed 2025 & seed 2026 \\
        \midrule
        \multirow{2}{*}{PatchLinear} & $\hat{\mu}$ & 0.4258 & 0.4259 & 0.4255 \\
                                     & $\hat{\sigma}$ & 0.0378 & 0.0374 & 0.0372 \\
        \midrule
        \multirow{2}{*}{PatchTST}    & $\hat{\mu}$ & 0.4484 & 0.4485 & 0.4455 \\
                                     & $\hat{\sigma}$ & 0.0580 & 0.0563 & 0.0509 \\
        \bottomrule
    \end{tabular}
\end{table}

As shown in \cref{tab:multiseed}, both models exhibit negligible variation in $\hat{\mu}$ (within 0.003) and moderate variation in $\hat{\sigma}$ across the three seeds.
Crucially, the relative ordering—PatchLinear outperforming PatchTST in both effectiveness and stability—is preserved across all seeds, confirming that the Identity Paradox is a robust finding rather than a seed-dependent artifact.

\end{document}